\documentclass[10pt,twocolumn,letterpaper]{article}

\usepackage{wacv}
\usepackage{times}
\usepackage{epsfig}
\usepackage{graphicx}
\usepackage{amsmath}
\usepackage{amssymb}
\usepackage{booktabs}

\usepackage{algorithm}
\usepackage{algorithmic}

\usepackage{enumitem}
\setitemize{noitemsep,topsep=0pt,parsep=0pt,partopsep=0pt}

\usepackage{tikz}

\newcommand{\argmax}[1]{\underset{#1}{\operatorname{arg}\!\operatorname{max}}\;}

\def\wacvPaperID{1342} 

\wacvalgorithmstrack   

\wacvfinalcopy 

\ifwacvfinal
\usepackage[breaklinks=true,bookmarks=false]{hyperref}
\else
\usepackage[pagebackref=true,breaklinks=true,colorlinks,bookmarks=false]{hyperref}
\fi

\begin{document}

\title{Interpolated SelectionConv for Spherical Images and Surfaces}

\author{David Hart, Michael Whitney, Bryan Morse\\
Brigham Young University\\ 
{\{davidmhart, mikeswhitney, morse\}@byu.edu}
}

\maketitle
\thispagestyle{empty}

\begin{abstract}
    We present a new and general framework for convolutional neural network operations on spherical (or omnidirectional) images. Our approach represents the surface as a graph of connected points that doesn't rely on a particular sampling strategy. Additionally, by using an interpolated version of SelectionConv, we can operate on the sphere while using existing 2D CNNs and their weights. Since our method leverages existing graph implementations, it is also fast and can be fine-tuned efficiently. Our method is also general enough to be applied to any surface type, even those that are topologically non-simple. We demonstrate the effectiveness of our technique on the tasks of style transfer and segmentation for spheres as well as stylization for 3D meshes. We provide a thorough ablation study of the performance of various spherical sampling strategies.
\end{abstract}

\section{Introduction}

Omnidirectional (or spherical) images are ever present in modern computer vision, from 360$^\circ$ videos for VR to HDRI images as lighting references in 3D graphics. Much research has gone into trying to bring the benefits of deep learning to this useful domain. A naive approach is to simply apply 2D Convolutional Neural Networks (CNNs) to a projection of the sphere, such as on the equirectangular image or a cubemap unfolding. However, all planar projections of a sphere lead to distortion of the 3D content. Thus, traditional 2D CNNs perform poorly in such cases.

To resolve this, two approaches have been well explored previously. The first  is to design the convolution operator uniquely for the sphere, looking at important properties such as rotation equivariance of the convolution operation~\cite{Cohen2019,Cohen2018,Chiyu2019}. 
The second is to try to transfer what has been learned from a 2D CNN to a specialized network that can operate on spherical images, either through distorting the convolution operator itself \cite{Coors2018,Tateno2018} or through special projections of the spherical image into the plane \cite{Eder2020,Zhang2019}. Both of these approaches have potential shortcomings. In the first, by redefining a convolution specific to the spherical domain, all training data must also be in the spherical domain, for which available datasets tend to be much smaller than their 2D counterparts. In the second approach, techniques are often restricted to particular samplings of the sphere (most often the icosphere \cite{Eder2020,Zhang2019}), which can only exist at specific resolutions, and transitions to these representations may require fine-tuning to properly adjust the CNN weights \cite{Zhang2019}. 

\begin{figure}[t]
\begin{center}
\begin{tabular}{cc}
\includegraphics[width=0.30\linewidth]{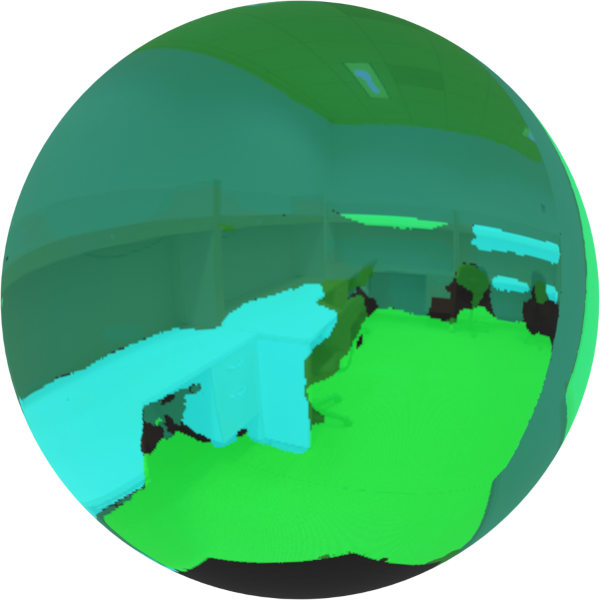} & 
\includegraphics[width=0.35\linewidth,trim={0 5cm 0 0},clip]{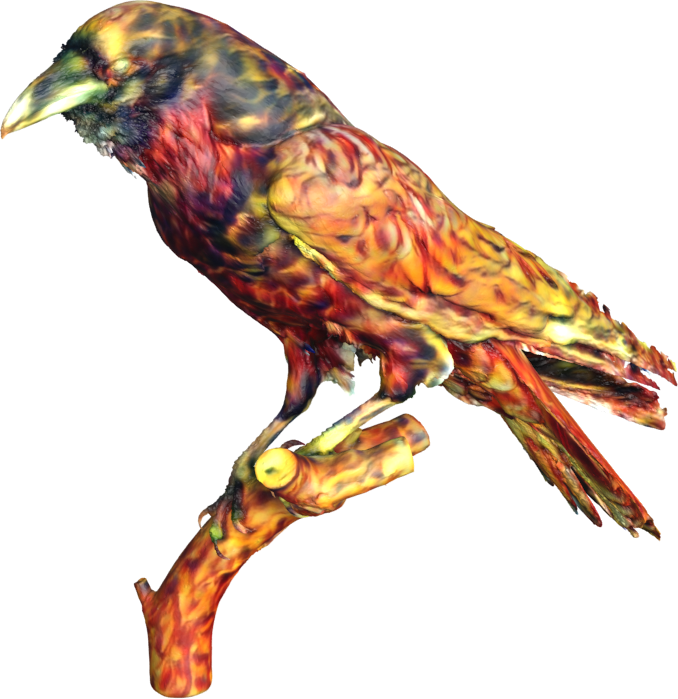}
\end{tabular}
\end{center}
   \caption{Interpolated SelectionConv can perform tasks such as segmentation on spheres (left) and 3D mesh stylization (right).
   }
\label{fig:intro}
\end{figure}

This work aims to begin to bridge the gap between these two approaches to operating on omnidirectional images. We propose a framework that builds on
our previous work of SelectionConv~\cite{SelectionConv}, a
graph-based approach for transferring weights from a 2D CNN. By using an edge-interpolated version of this method, we can create a graph of points on the sphere using any sampling scheme that we desire and still maintain the anisotropic behavior of convolution that is needed for transferring from 2D CNNs. Our 
new 
approach is general while remaining fast and efficient since it builds on previous graph network infrastructure.

We show that our method is a simple and effective way to bring 2D techniques into the spherical domain, while avoiding many of the drawbacks of regular SelectionConv and previous approaches. We also show that customizable sampling techniques may be more effective than the icosphere approach that has generally been used. Lastly, while we primarily focus on spherical images for this work, our method is general enough to operate on any surface or 3D mesh. 
Examples of tasks that can be performed with our method are shown in Fig.~\ref{fig:intro}.

In summary, our contributions are as follows:

\begin{itemize}
    \item We introduce an interpolation-based scheme for using SelectionConv in non grid-like spaces.
    \item We adapt Interpolated SelectionConv to work for surface point-cloud representations by modifying the selection function and clustering method.
    \item We demonstrate the effectiveness of our graph representation with customizable sampling techniques on the tasks of spherical stylization and segmentation. We also demonstrate stylization of 3D meshes. We achieve results comparable to state-of-the-art methods.
    \item We present a thorough ablation study of various spherical sampling and clustering techniques. 
\end{itemize}

\section{Related Works}

\subsection{SelectionConv} \label{sec:SelectionRelated}

This work builds heavily on the idea of selection-based convolution, which was first proposed in SelectionConv~\cite{SelectionConv}, 
a method for using CNNs on irregular image types such as cubemaps and masked images. 
In selection-based convolution, graph networks can be manipulated to be commensurate with 2D CNNs. This is done by preprocessing a graph's adjacency matrix into multiple adjacency matrices, where each matrix represents a cardinal or ordinal direction. By doing this, spatial structure is built back into the graph and weights can be transferred from a trained 2D CNN to a modified graph network.
Thus, graph structures can be designed for specific irregular image types and the same 2D networks can operate in these new domains.

In \cite{SelectionConv}, all experiments were done with graph representations that assume a single selection per target node. 
This allowed graph connections to go across seams for irregular images, but a single selection is not sufficient when building the graph in more general spaces such as arbitrary surfaces or point clouds.
We extend this approach by demonstrating how to effectively allow multiple selections per node to give interpolated values when selections lie between discrete spatial locations.
This is similar to techniques such as \cite{InterpConv}, but with the added benefit that we can orient the interpolation for each individual node on a surface. 

\subsection{Spherical CNNs} \label{sec:RelatedWork_Spherical}

Neural networks designed for spheres fall roughly into two categories. The first is projection-based CNNs that aim to learn on traditional 2D images and then utilize that information in some way on the sphere. The second is surface-based CNNs that have special convolution operations designed specifically for spheres.\\

\noindent \textbf{Projection-based CNNs}

The first approaches to applying CNNs to the sphere operated on projections of the sphere such as the equirectangular projection \cite{Lai2018}, cubemap projection \cite{Cheng2018,Monroy2018}, or even a combination of multiple projections \cite{Jiang2021,Su2017,Wang2020}. After the proposal of deformable convolution by Dai \etal \cite{Dai2017}, many researchers focused on deforming the convolution operator specifically for sphere projections \cite{Coors2018,Su2019,Tateno2018}. 

Lee \etal~\cite{Lee2019} later demonstrated the effectiveness of operating on an icosahedral approximation of the sphere. Zhang \etal~\cite{Zhang2019} adapted a similar representation, but allow for a more direct CNN operation by breaking the icosahedron into 5 planar projections. Eder \etal~\cite{Eder2020} increased performance on many spherical tasks by representing the sphere as a set of tangent images at the icosphere vertices. \\

\noindent \textbf{Surface-based CNNs}

Cohen \etal \cite{Cohen2019,Cohen2018} have proposed a convolution structure designed specifically for spheres and surfaces that is rotation and gauge equivariant. Representations have also been designed that utilize spherical harmonics~\cite{Esteves2018} or define the convolution in terms of a linear combination of differential operators \cite{Chiyu2019}. This line of work continues to expand to be used for more complex and general spherical networks, such as LSTMs~\cite{Xu2021}.

Our method aligns closely with projection-based approaches since it transfers weights from a 2D CNN and operates with oriented convolutions on the sphere. Our approach, however, builds on a framework that is not limited to spheres, and, like surface-based CNNs, is not constrained to a specific sampling technique.

\subsection{Sphere and Mesh Style Transfer}

Many spherical approaches demonstrate performance on tasks such object detection, semantic segmentation, and depth prediction. While we compare our method on the task of segmentation, we also demonstrate that our representation generates high-quality stylizations of the sphere. Ruder \etal~\cite{Ruder2018} were the first to demonstrate style transfer on a spherical image, but their method requires a slow optimization approach or fine-tuning a style network for spheres. Like \cite{SelectionConv}, our approach can stylize the sphere in a single feed-forward pass with any content and style image, but we do so with fewer artifacts than previous approaches.

Since our representation can easily be applied to general surfaces, we demonstrate style transfer for meshes as well. There have been many approaches to applying convolutions to surfaces and 3D meshes~\cite{Haim2019,Hanocka2019,Lahav2020,Sinha2016,Wiersma2020}, but these convolutions are trained on 3D data for 3D-specific tasks. Some methods have performed mesh style transfer~\cite{StyLit,SelectionConv,StyBlit,yin2021}, but these approaches are optimization-based, require professional reference renderings, use aligned texture maps, or have artifacts at seams. Our approach is simple, fast, and has no model or style prerequisites.

\section{Interpolated SelectionConv} \label{sec:Interpolation}

As discussed in Sec.~\ref{sec:SelectionRelated}, our method builds on the original SelectionConv \cite{SelectionConv}, which proposes a modified graph network that can distinguish between the edges of a graph during a convolution operation. 
Thus, if a graph is designed and processed for a specific type of irregular image, selection-based graph convolution will preserve the orientation and allow a traditional CNN to operate even in the irregular space. 

In order to distinguish the edges during a graph convolution, the graph must be preprocessed into multiple adjacency matrices. Selection-based convolution extends the common formulation of graph convolution to reflect these different spatially-relative adjacency matrices. Specifically,
\begin{equation}
    \label{eq:selection-conv}
    \mathbf{X}^{(k+1)} = \sum_m{\tilde{\textbf{S}}_m \mathbf{X}^{(k)} \textbf{W}_m}
\end{equation}
\noindent where $m$ represents a given direction/selection, 
$\tilde{\textbf{S}}_m$ is the adjacency matrix for that selection, $\mathbf{X}^{(k)}$ is the current node activations, and $\textbf{W}_m$ is the learned weights for that selection. Rather than a binary adjacency matrix, note that $\tilde{\textbf{S}}_m$ is normalized so that multiple incoming edges can have the same selection, allowing for more general point-cloud-like structures.  This work further utilizes this normalization to break the restriction that each edge have a single selection.

In
our
original SelectionConv work, edges were given the selection of the cardinal or ordinal direction most closely aligned with the spatial relationship between the respective pixels or nodes. With general sampling patterns, however, node relationships do not lie so cleanly along specific axes. Thus, it is beneficial to be able to interpolate between multiple selections. To do this, we first assign multiple edges to a single node with each edge having a different selection (i.e., we assign the same source and target node to multiple adjacency matrices). Then, we assign each of those edges an edge weight that determines how much the spatial relationship between the nodes matches the given direction.

With this, we have two different senses of normalization. The first accounts for our introduced interpolation by normalizing all the assigned edges for a given source and target node, making the total of the interpolation weights across selections equal to one, or mathematically:
\begin{equation}
\hat{S}_m(i,j) = \frac{S_m(i,j)}{\sum_k{S_k(i,j)}}
\end{equation}
\noindent This guarantees that a single target node can not have an unbalanced amount of influence during the aggregation step. The second normalizes in the manner discussed in \cite{SelectionConv}, where each of the adjacency matrices is normalized individually so that for a given source node, the sum of all the target node edge weights equals one, or mathematically:
\begin{equation}
\tilde{S}_m(i,j) = \frac{\hat{S}_m(i,j)}{\sum_k{\hat{S}_m(i,k)}}
\end{equation}

With this formulation, we are no longer constrained to discrete selections but can interpolate selection values within the convolution operation. There are various ways to interpolate a location's value given a set of points. We demonstrate two simple and effective ways to interpolate selection weights given source and target nodes.

\subsection{Angle-based Interpolation}\label{sec:Angle}

\begin{figure}
\begin{center}
\begin{tabular}{cc}
\includegraphics[width=0.35\linewidth]{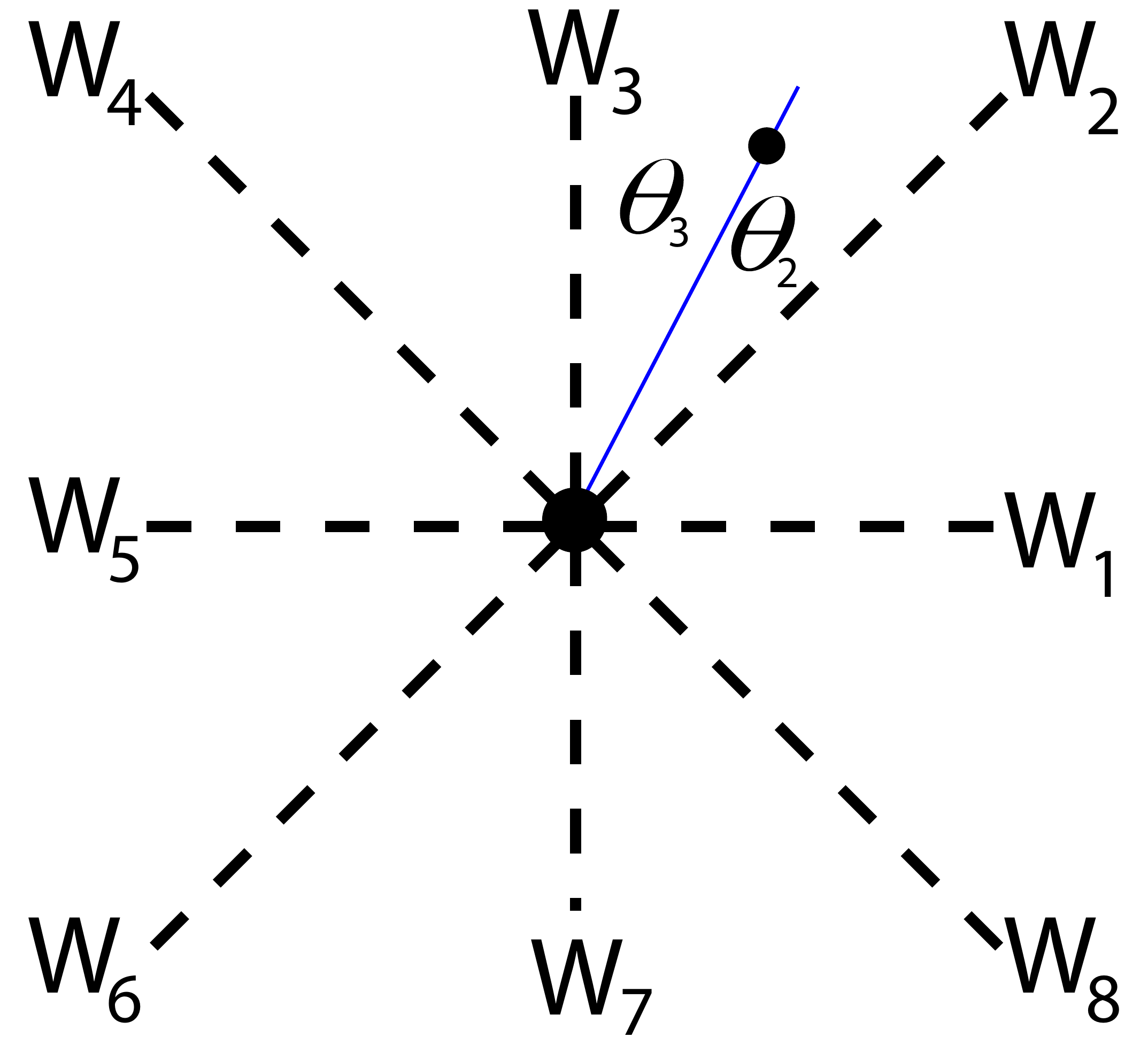}
& \includegraphics[width=0.35\linewidth]{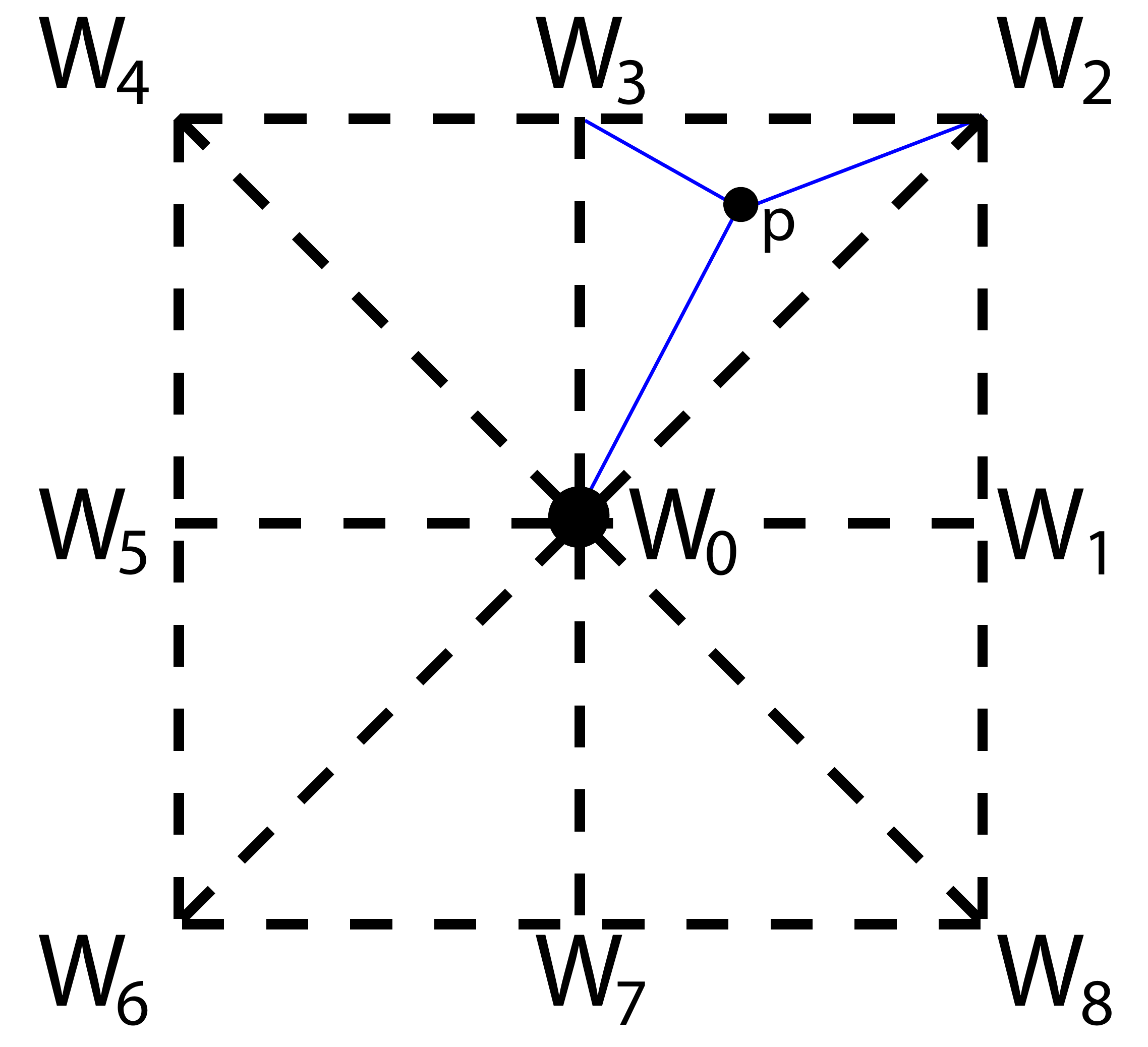}
\\ 
a) Angular Interpolation &
b) Barycentric Interpolation
\end{tabular}
\end{center}
   \caption{Illustration of a) angular interpolation versus b) barycentric interpolation. In angular interpolation, the edge to each node is added to the adjacency matrix of the two selections it lies between. If a local distance/radius can be determined, barycentric interpolation can be used which also includes the central selection. 
   }
\label{fig:interpolation}
\end{figure}

One way to practically determine the interpolation of a node between two selections is to use angles. Instead of simply choosing the selection in the direction that most closely matches a given edge, we use the angles between the given edge and the unit vectors for each selection direction (excluding the central selection). The two selections, $a$ and $b$, with the smallest angles are added to the respective adjacency matrices and the edge weights are equal to the convex combination of the two angles, or mathematically:
\begin{equation}
w_a = \frac{\theta_b}{\theta_a+\theta_b}
\end{equation}
\begin{equation}
w_b = 1 - w_a
\end{equation}
\noindent This process is illustrated in Fig.~\ref{fig:interpolation}.a.

\subsection{Barycentric Interpolation}

The angle-based interpolation is intuitive and would be expected to perform well for graph structures where nodes are spaced approximately equidistant from each other. However, another interpolation scheme can be used that accounts properly for distance from the source node. If an average or expected distance between nodes can be calculated, then the interpolation can be triangulated between three selections using barycentric interpolation. This process is illustrated in Fig.~\ref{fig:interpolation}.b.

Though the general form of barycentric interpolation requires inverse matrices or other time-consuming calculations, it can be dramatically simplified since the three points of interpolation will always lie along a perfect right triangle with two bases of length $d$. In these circumstances, the three edge weights become
\begin{equation}
w_0 = 1 - \frac{\text{max}(|p|)}{d}
\end{equation}
\begin{equation}
w_a = \frac{\text{max}(|p|) - \text{min}(|p|)}{d}
\end{equation}
\begin{equation}
w_b = \frac{\text{min}(|p|)}{d}
\end{equation}
\noindent where $p$ is the location of the target node relative to the source node. Further details and a proof of this formulation are shown in Appendix A of the supplemental material.

\section{The Spherical Graph}\label{sec:Spheres}

As noted in Sec.~\ref{sec:RelatedWork_Spherical}, many networks that operate on spherical images must first project the spherical data to some form. 
Ideally, rather than working on projections, it would be best to work on the spherical data directly without relying on projection. Thus, we generate our graph structure in 3D space using the spherical surface and adjust our selection function appropriately. We will now give the details of how we generate the exact graph structure.

\subsection{The Spherical Selection Function} \label{sec:SphericalFunction}

The original selection function used in \cite{SelectionConv} is
\begin{equation}
\label{eq:selection}
    s(v_i,v_j) =
        \begin{cases}
            0 \text{ if } \|\mathbf{x}_j - \mathbf{x}_i\| < \epsilon \\
            \argmax{k} \mathbf{D}_k \cdot \left( \mathbf{x}_j - \mathbf{x}_i \right) \ \mbox{otherwise}
        \end{cases}
\end{equation}
\noindent where $v_i$ and $v_j$ are two nodes of the graph, $\mathbf{x}_i \in \mathbb{R}^2$ and $\mathbf{x}_j \in \mathbb{R}^2$ are their respective positions, and $\{ \mathbf{D}_k \in \mathbb{R}^2, 1 \leq k \leq 8 \}$  represents the set of unit vectors for each of the 8 cardinal/ordinal directions. In order to work on spheres, we modify this selection function by making selections relative to the surface of the sphere, using a local planar approximation to determine orientation.

To make a local planar approximation, we first need the normal $\mathbf{\hat{z}}$ to each node position (which is conveniently the normalized version of the 3D location $\mathbf{x}_i \in \mathbb{R}^3$ for a sphere). Next, we generate a rotation matrix using the approximate up-vector of $\mathbf{\tilde{y}} = \left<0,1,0\right>$ and Graham-Schmidt orthogonalization. Specifically,
\begin{equation}
    \mathbf{\hat{x}} =  \mathbf{\tilde{y}} \times \mathbf{\hat{z}}
\end{equation}
\begin{equation}
    \mathbf{\hat{y}} =  \mathbf{\hat{z}} \times \mathbf{\hat{x}}
\end{equation}
\begin{equation}
    \mathbf{R} = 
    \begin{bmatrix}
        \mathbf{\hat{x}}\\
        \mathbf{\hat{y}}\\
        \mathbf{\hat{z}}
    \end{bmatrix}
\end{equation}
\noindent With this rotation matrix, edges from the node can be rotated relative to the surface. Then, the new z-coordinate is dropped, giving a local planar approximation of the coordinate relative to the surface. This dropping of z-coordinate can be represented by multiplying by a truncated identity matrix:
\begin{equation}
    \mathbf{I}_t = 
    \begin{bmatrix}
        1 & 0 & 0\\
        0 & 1 & 0\\
    \end{bmatrix}
\end{equation}

\noindent Thus, the final spherical selection function (for a single selection) becomes
\begin{equation}
    s(v_i,v_j) =
        \begin{cases}
            0 \text{ if } \|\mathbf{x}_j - \mathbf{x}_i\| < \epsilon \\
            \argmax{k} \mathbf{D}_k \cdot \mathbf{I}_t\mathbf{R}\left( \mathbf{x}_j - \mathbf{x}_i \right) \ \mbox{otherwise}
        \end{cases}
\end{equation}

\begin{figure}
\begin{center}
\begin{tabular}{cc}
\includegraphics[width=0.40\linewidth]{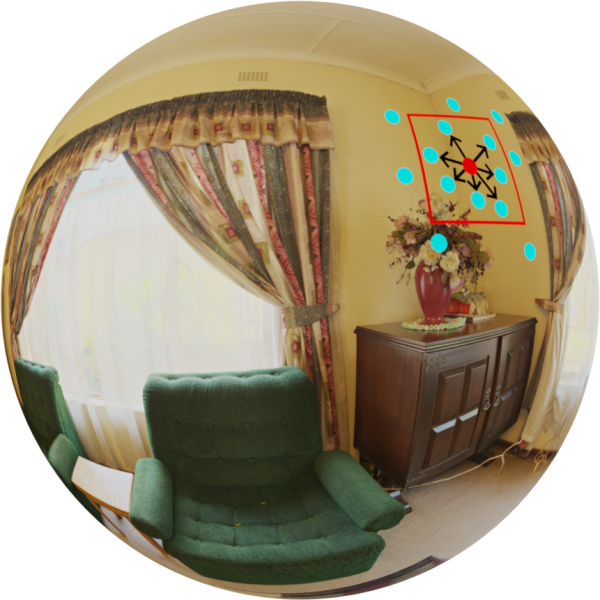}
& \includegraphics[width=0.43\linewidth]{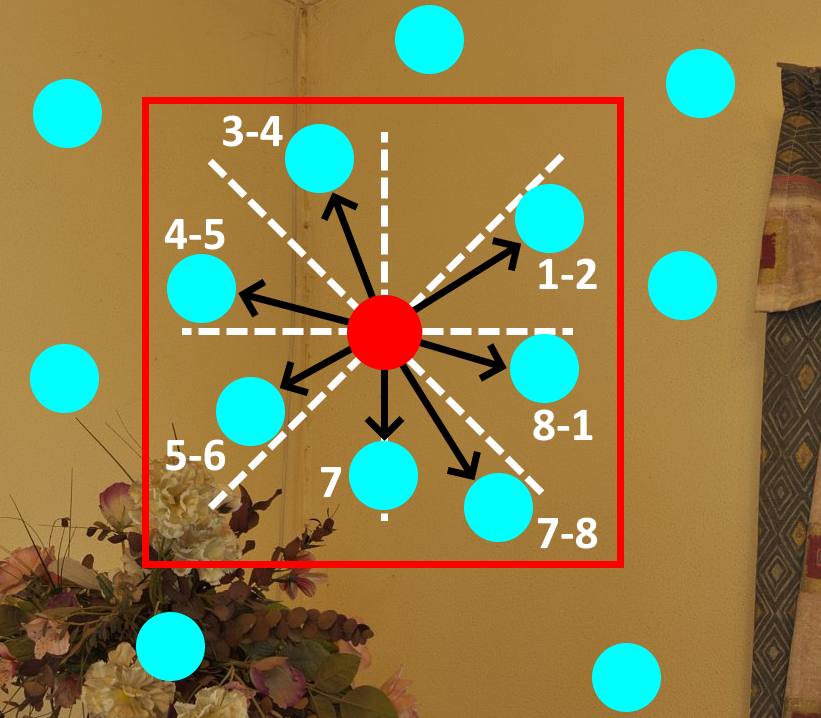}
\\
a) Spherical Graph &
b) Local Planar Projection
\end{tabular}
\end{center}
   \caption{Visualization of the graph connections being made in 3D space (a), and selections being made based on the local planar projection at each node (b).
   }
\label{fig:sphericalgraph}
\end{figure}

\noindent Once the main selection is made, other selections and interpolation values can be determined according to Sec.~\ref{sec:Interpolation}. An illustration of our final graph structure is shown in Fig.~\ref{fig:sphericalgraph}.

\subsection{Sampling}
Most spherical data is stored in the form of an equirectangular image, where the spherical data has been projected to a cylinder and then unfolded to the plane. This representation is easy to store, but the pixels do not represent locations that are equidistantly sampled across the sphere. Preferably, the algorithm should sample points from the image in such a way that their 3D locations are more consistently spaced across the surface.

For this work, we consider the following sampling strategies when comparing to the equirectangular baseline:
\begin{itemize}
    \item \textbf{Fibonacci Spiral}: A common and fast approach for approximating equidistant points.
    \item \textbf{Icosphere}: Points generated by taking the icosahedron and subdividing iteratively.
    \item \textbf{Layering}: A sampling method described in \cite{sphereSampling}, where points are appropriately distanced in $\theta$ based on equidistant $\phi$ values.
\end{itemize}

\begin{figure*}
\begin{center}
\begin{tabular}{ccccc}

\includegraphics[width=.14\linewidth,trim={75 60 70 60},clip]{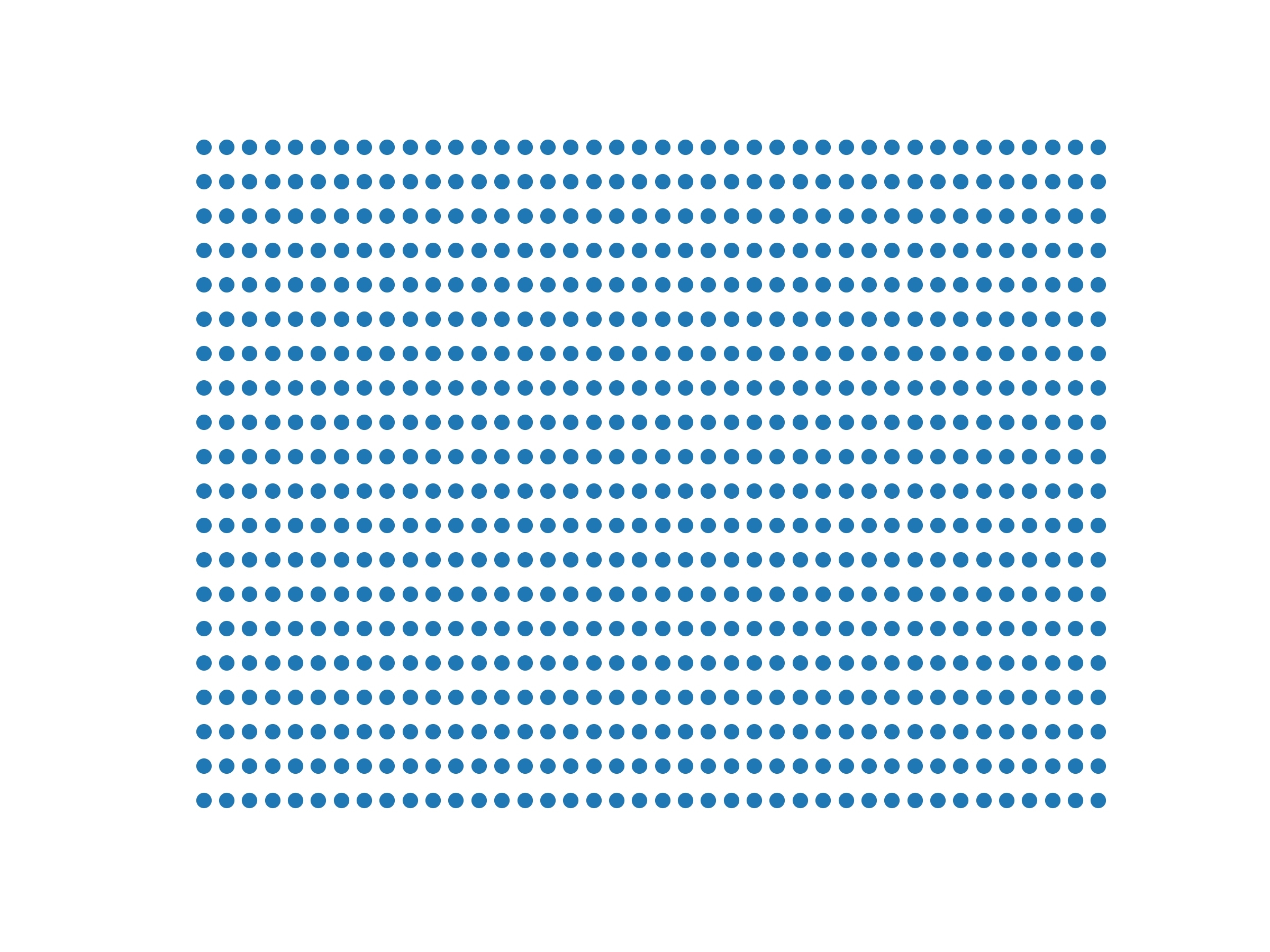} &
\includegraphics[width=.14\linewidth,trim={75 60 70 60},clip]{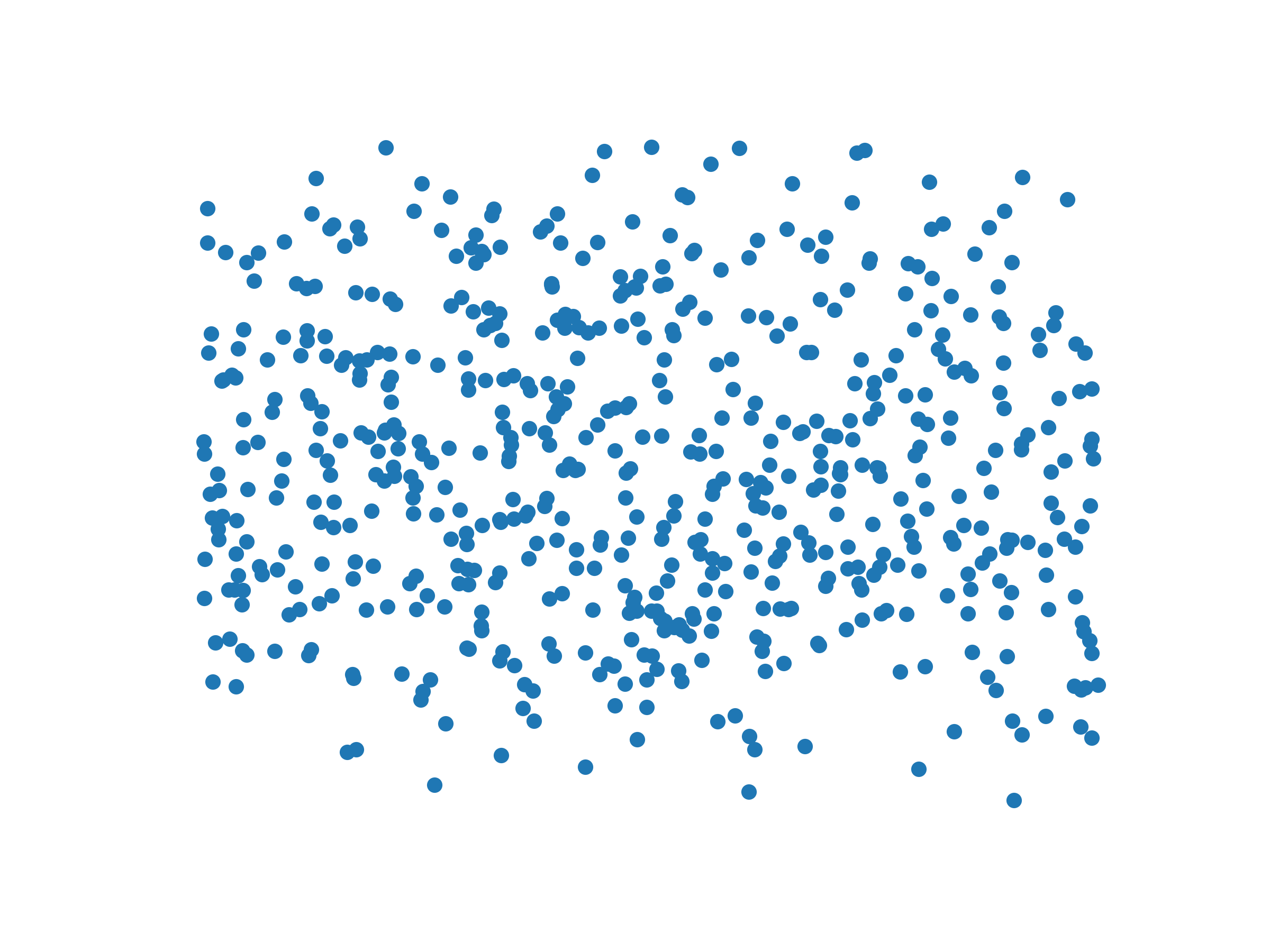} &
\includegraphics[width=.14\linewidth,trim={75 60 70 60},clip]{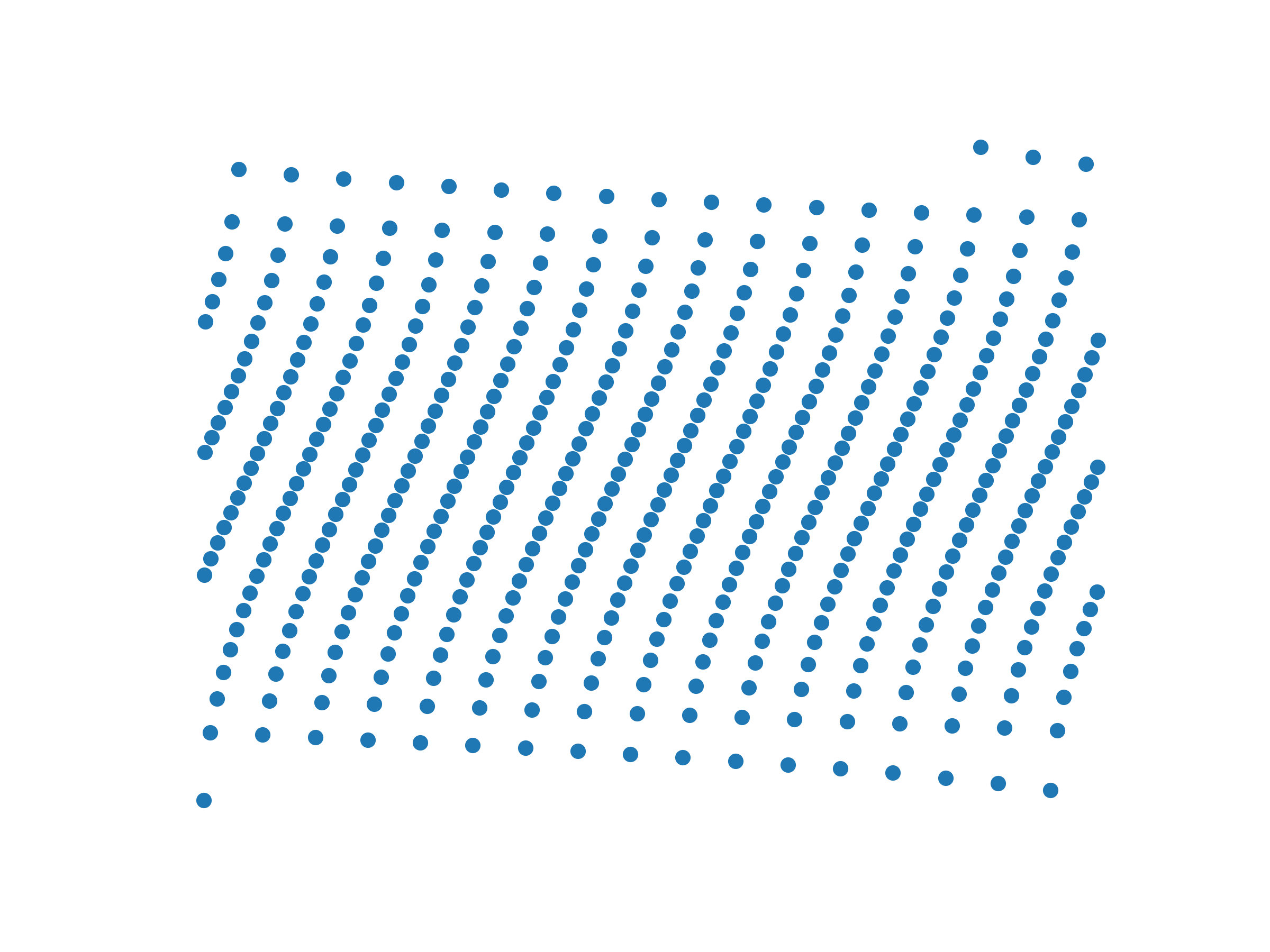} &
\includegraphics[width=.14\linewidth,trim={75 60 70 60},clip]{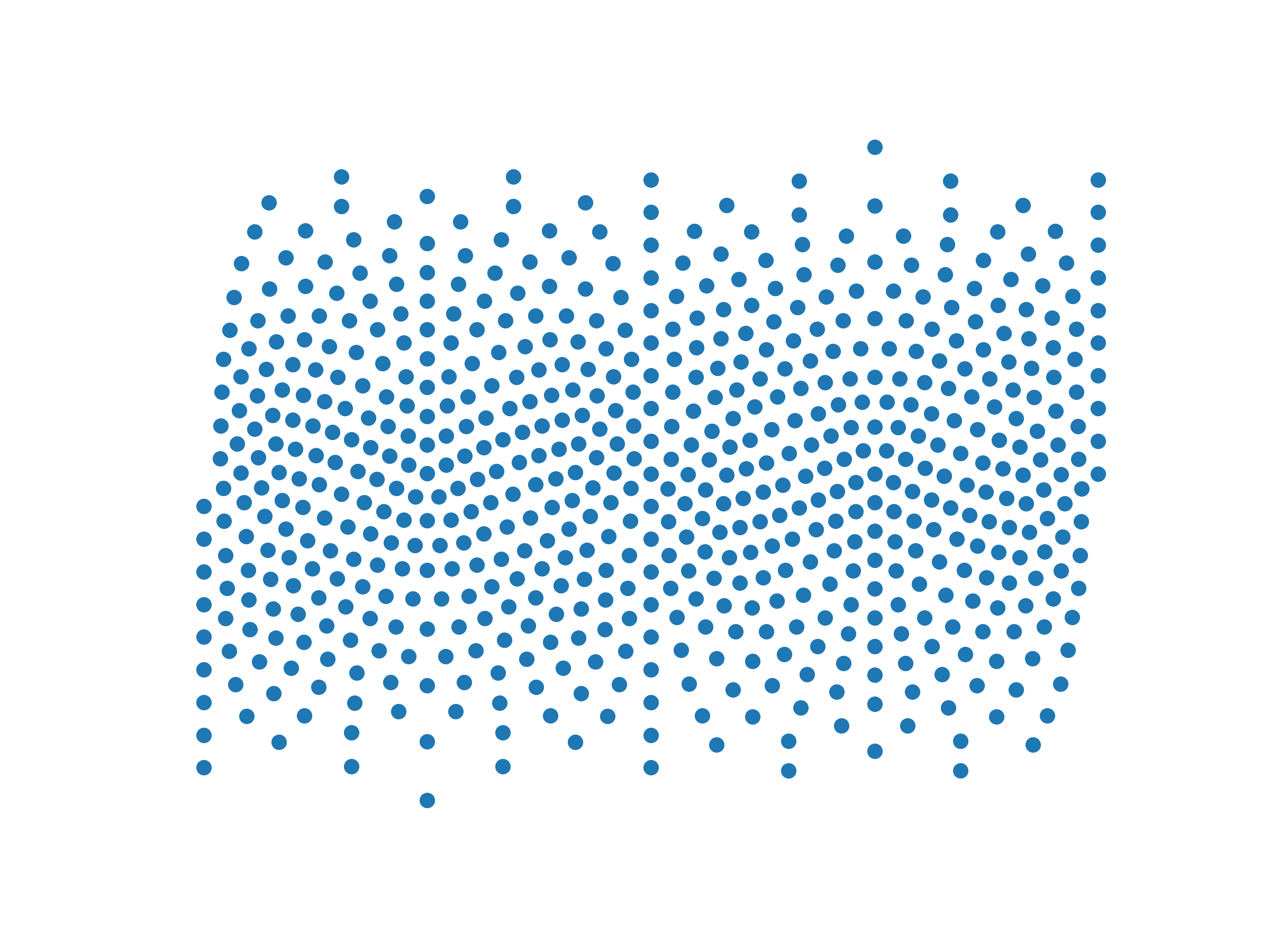} &
\includegraphics[width=.14\linewidth,trim={75 60 70 60},clip]{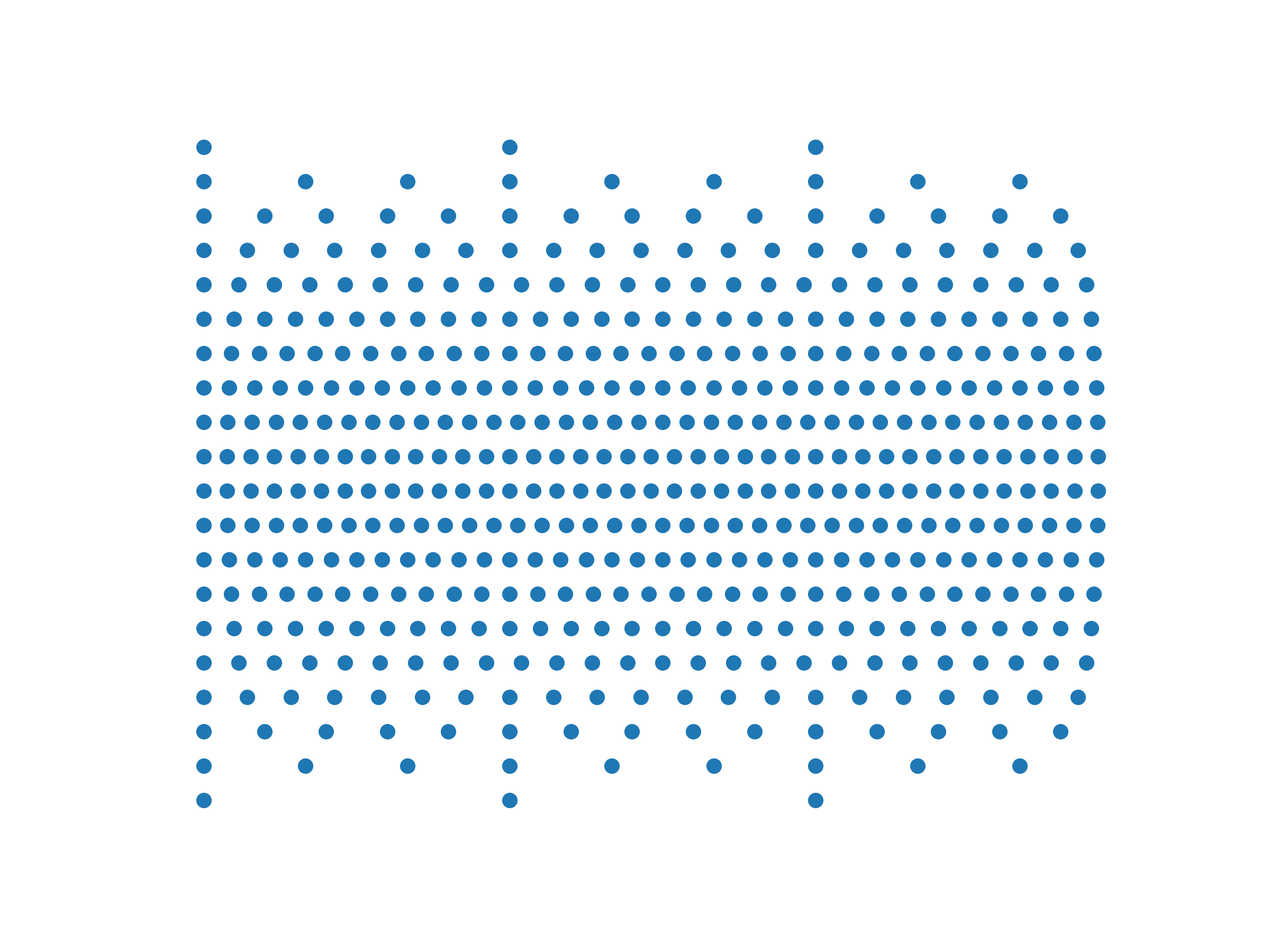}
\\
\includegraphics[width=.16\linewidth,trim={120 70 100 100},clip]{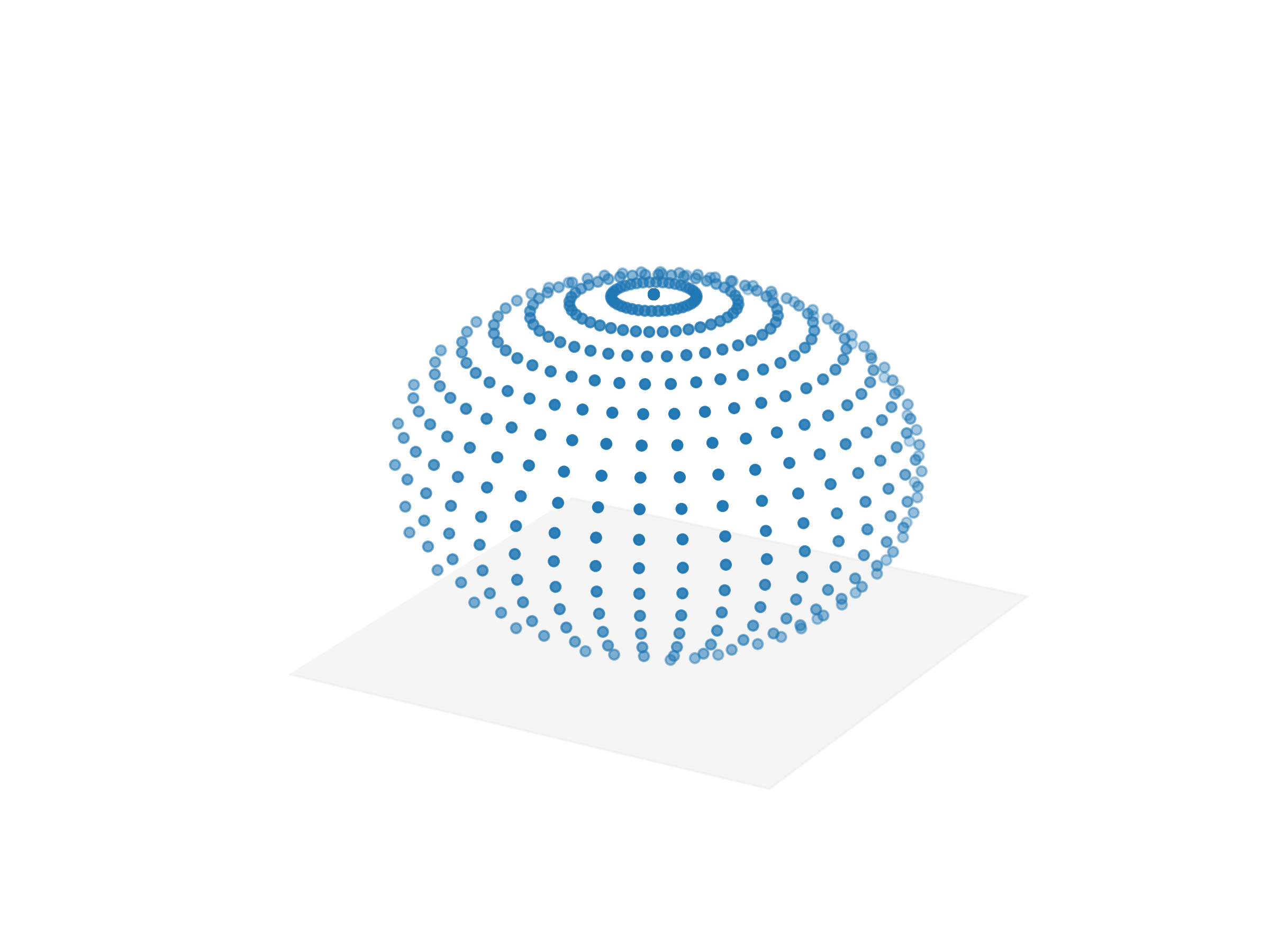} &
\includegraphics[width=.16\linewidth,trim={120 70 100 100},clip]{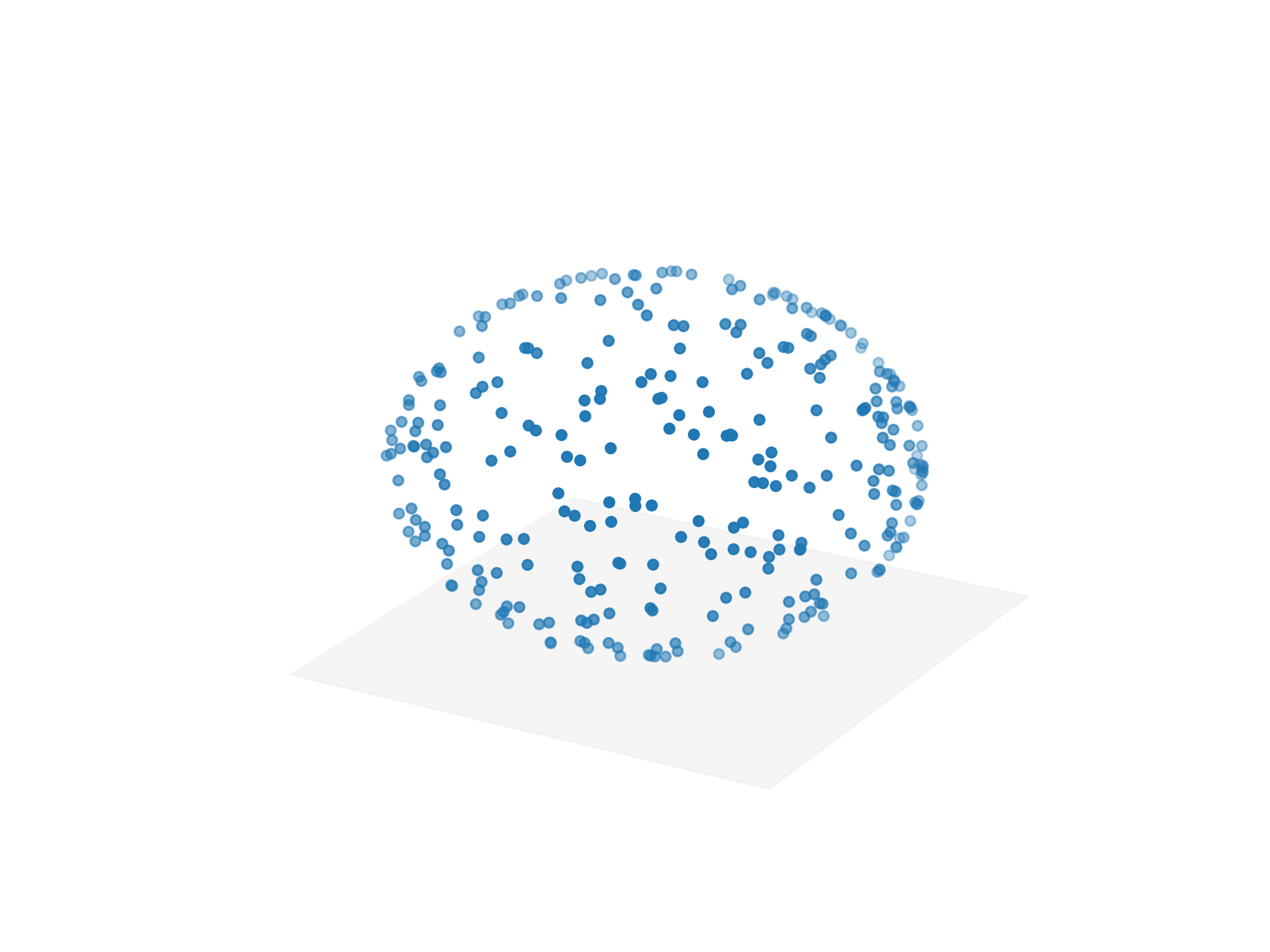} &
\includegraphics[width=.16\linewidth,trim={120 70 100 100},clip]{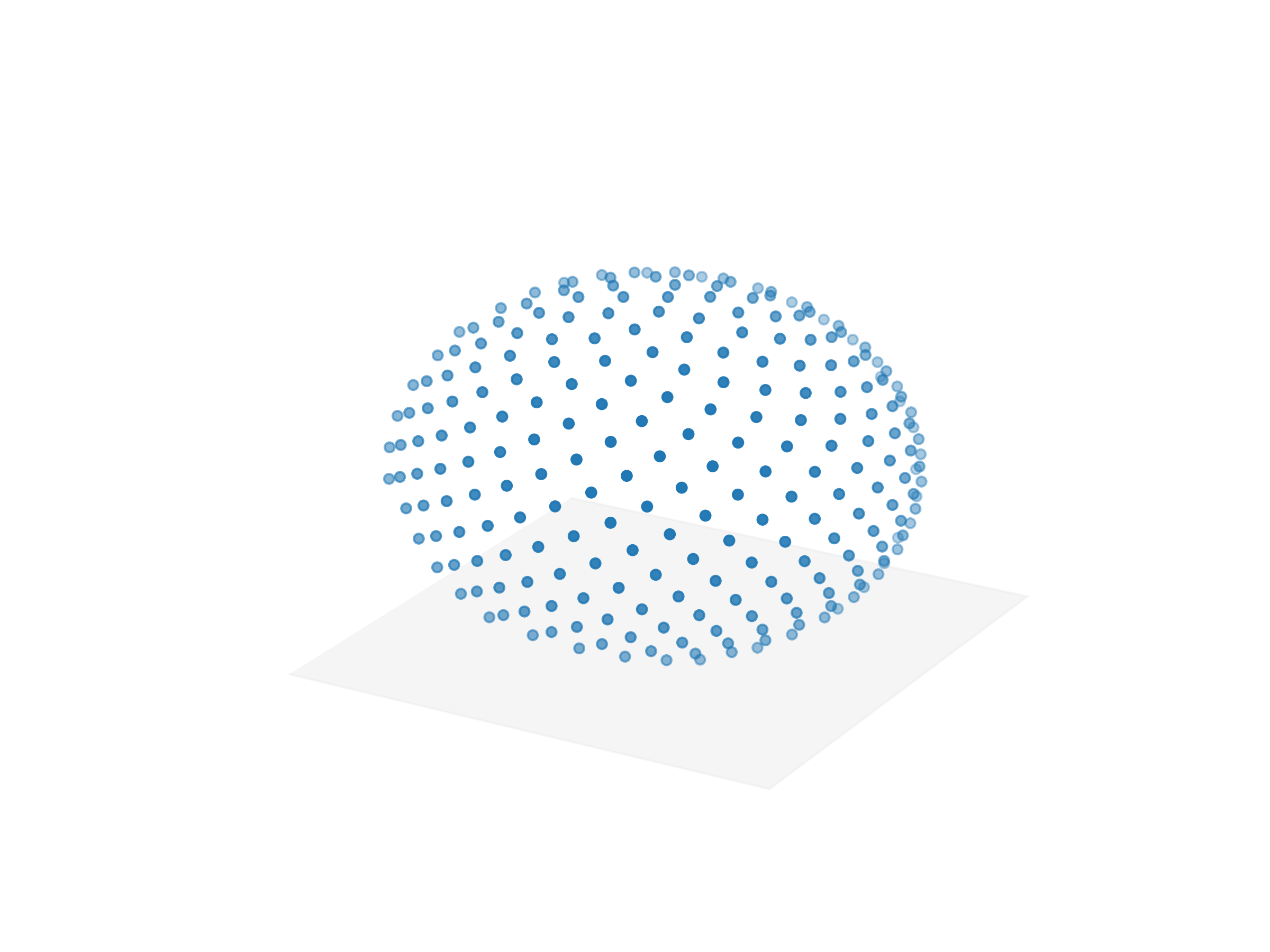} &
\includegraphics[width=.16\linewidth,trim={120 70 100 100},clip]{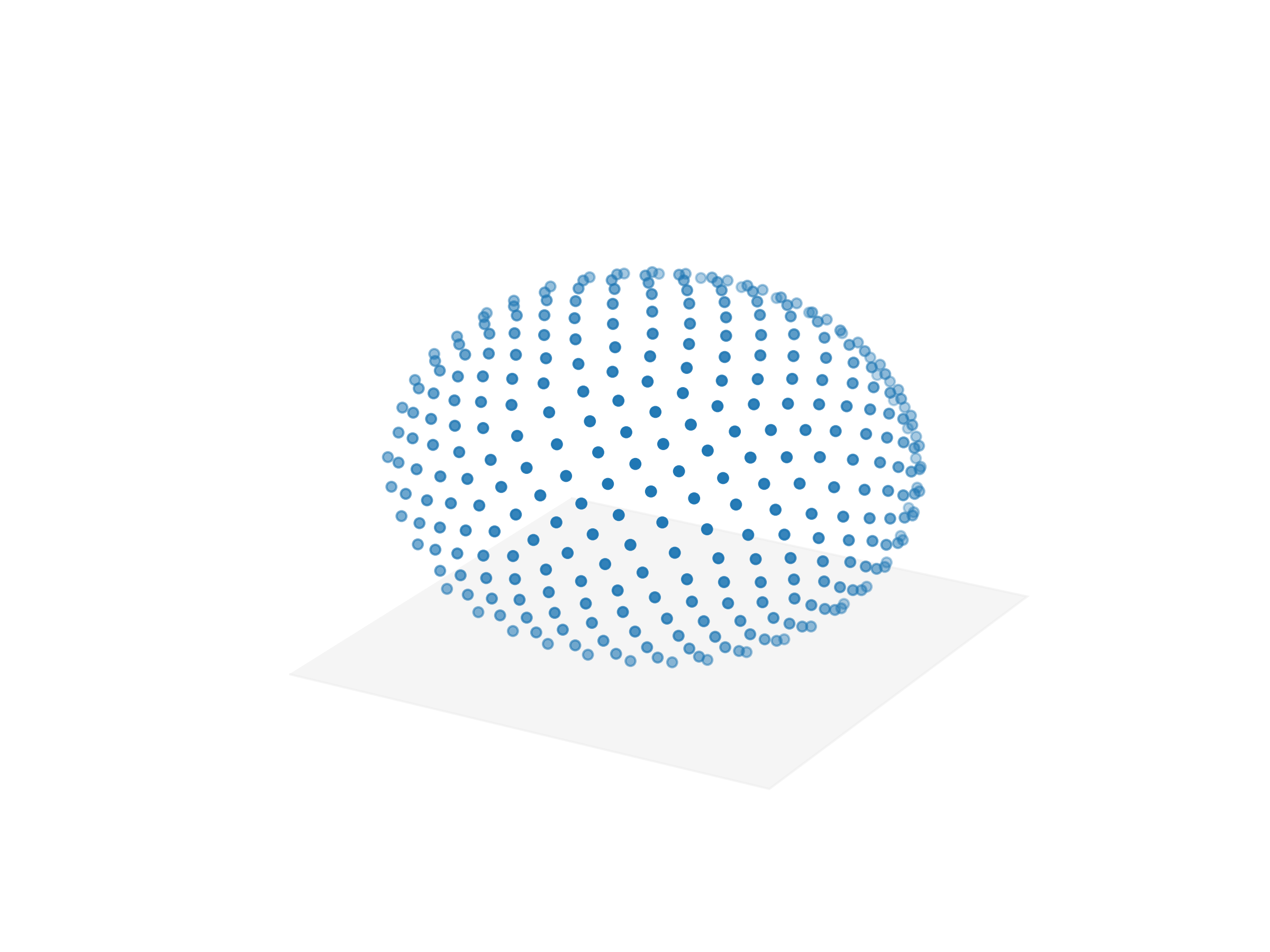} &
\includegraphics[width=.16\linewidth,trim={120 70 100 100},clip]{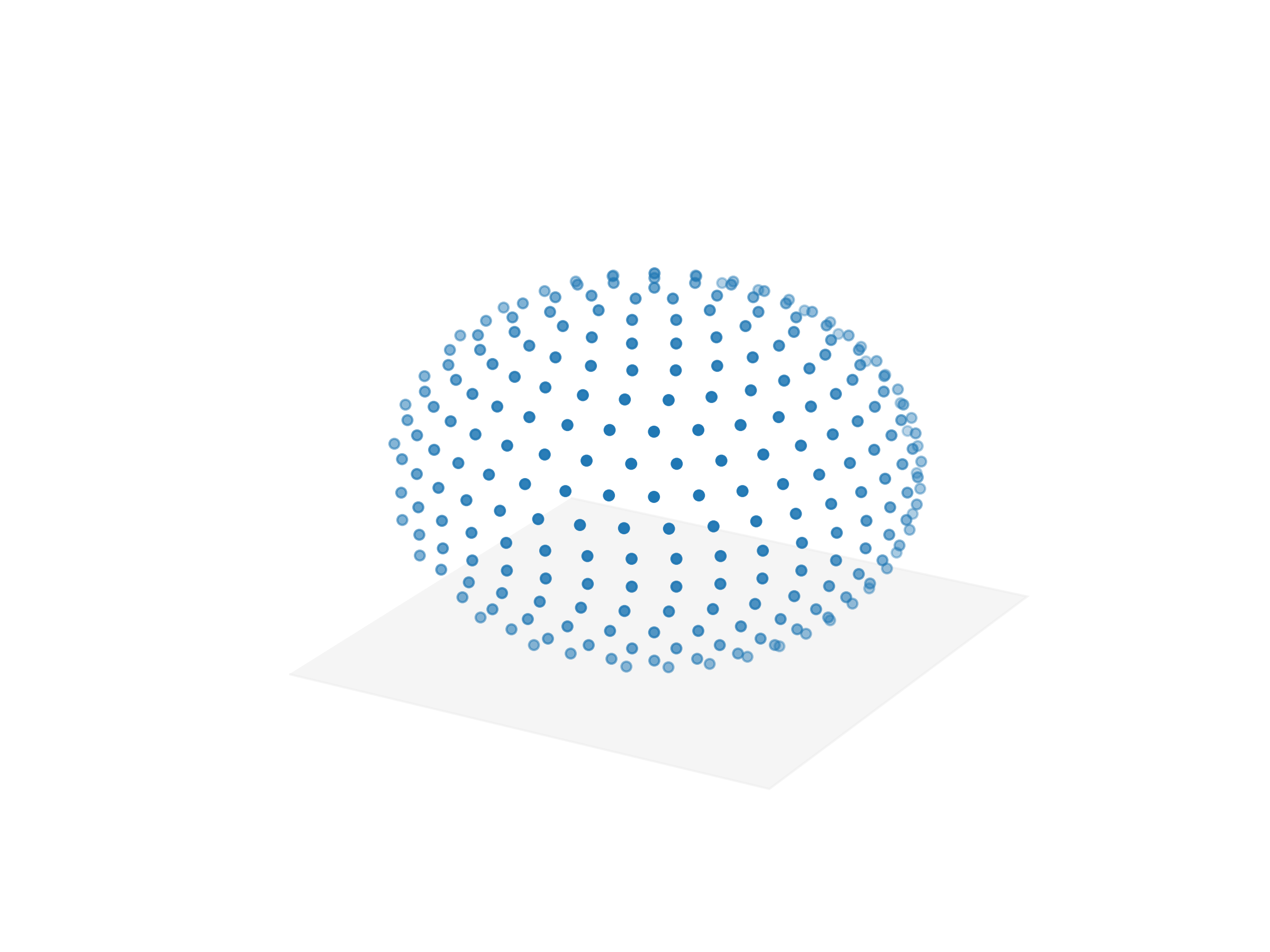}
\\
Equirectangular &
Random &
Fibonacci Spiral &
Icosphere &
Layering

\end{tabular}
\end{center}
   \caption{ 
        The various sampling methods we test for building our graph on the spherical surface. The sampling points on the equirectangular image (top) correspond to particular points on the spherical surface (bottom).
    }
\label{fig:sampling}
\end{figure*}

\noindent Examples of each of these sampling schemes are shown in Fig.~\ref{fig:sampling}. 
We include the layering sampling algorithm here (Algorithm 1) for convenience, which we adapt from \cite{sphereSampling}.

\begin{algorithm}
\caption{Layering Sampling}
\begin{algorithmic}[1]
\REQUIRE $N_\phi$, (i.e. number of sampling rows)
\STATE $a \gets \pi^2/N_\phi^2$
\STATE $d \gets \sqrt{a}$
\STATE $M_\phi \gets \text{round}[\pi/d]$
\STATE $d_\phi \gets \pi/M_\phi$
\STATE $d_\theta \gets a/d_\phi$

\FOR{each $m$ in $0...M_\phi -1$}
    \STATE $\phi \gets \pi(m+0.5)/M_\phi$
    \STATE $M_\theta \gets \text{round}[2\pi\text{sin}(\phi/d_\theta)]$
    
    \FOR{each $n$ in $0...M_\theta - 1$}
        \STATE $\theta = 2\pi n/M_\theta$
        \STATE Append point ($\theta,\phi$)
    \ENDFOR
\ENDFOR
\end{algorithmic}
\end{algorithm}


\subsection{Clustering}

Nodes in the graph need to be clustered together for the downsampling stages of the CNNs. When operating on irregular images as in \cite{SelectionConv}, these could be easily determined by grouping pixels together on a 2D grid. However, determining clusters for a sphere based on the 2D equirectangular image removes the benefits of working in a 3D space. Thus, to cluster in 3D, we must resample the sphere at a lower resolution. Each current node is clustered to the resampled node of closest proximity. 

All of the previous sampling techniques (as well as random sampling) can be used, with an aim for a desired cluster size. For example, a stride of 2 in each dimension is commonly used in downsampling layers of CNNs. This would give a desired cluster size of 4, and each of the sampling methods would be used in the following ways:

\begin{itemize}
    \item \textbf{Random}: Randomly generate 1/4 as many points.
    \item \textbf{Fibonacci Spiral}: Resample with 1/4 as many points.
    \item \textbf{Icosphere}: Complete one less subdivision of the icosahedron (since 1/4 as many points will be generated).
    \item \textbf{Layering}: Increase the distance between $\phi$ values by 2 and resample.
\end{itemize}


\subsection{Customizable Resolution}

When 2D CNNs are trained, there is generally a fixed resolution that the input data is set too. Additionally, there is usually a small range of field-of-views in the cameras that were used to generate the training data. In a 2D sense, this intrinsically sets an expected size and scale that the network is expecting to see certain objects. While data augmentation and other techniques can make the network more flexible to some degree, it is important that our graph structures lie at relatively the same resolution and field-of-view that the network is expecting.

For example, the Stanford2D-3D-S dataset \cite{Armeni2017} that we evaluate on in Sec.~\ref{sec:Results} uses training images that have a field-of-view (FOV) between 45$^\circ$ and 75$^\circ$.
If a 2D CNN was trained on N $\times$ N images of that camera, the relative angle between points on our 3D surface in any direction should be approximately:
\begin{equation}
  \Delta \theta = \frac{\text{FOV}}{N}  
\end{equation}

\noindent This spacing is handled by the sampling method we use for the sphere. As mentioned previously, most methods have relied on icosphere sampling for their approaches, but the icosphere is constructed through subdivisions and thus is limited to discrete resolutions that are quadruplings of the initial resolution. With a sampling approach such as layering, however, we can directly control the sampling density on the sphere and thus match the desired $\Delta \theta$ more precisely. This gives it improved performance over the icosphere, which we demonstrate in Sec.~\ref{sec:Ablation}.


\section{Surface Graphs}

The approach described in Sec.~\ref{sec:Spheres} was specific to spheres, but with a few simple modifications, it can be applied to any general surface, specifically 3D meshes. To generate the graph on a surface, we start by randomly sampling the surface. For each node, the normal of the face it was sampled from and its respective UV coordinate are also stored. Next, edges connecting nearby points are made using a KNN approach (while also accounting for folds by culling points with disparate normals).

Then, to determine selections, the same selection process and selection function as Sec.~\ref{sec:SphericalFunction} can be used with two key differences. First, the approximate up-vector for the surface is either manually assigned (if a reasonable volume orientation exists) or is randomly generated. Second, the normal $\mathbf{\hat{z}}$ used for the Graham-Schmidt orthogonalization is the stored face normal for each node. All other steps in the process remain the same for the graph generation. The surface is then resampled at a lower resolution to determine clustering nodes, and the whole process is repeated for the needed number of downsamplings.

As is shown in Sec.~\ref{sec:MeshResults}, one benefit of this approach is that, just like the sphere, we can customize the approximate resolution of the point cloud on the mesh by controlling the number of points in the initial sampling. Increasing the number of initial points increases the level of detail to which we operate on the mesh.


\section{Results} \label{sec:Results}
For the results presented, we use the layering method for both sampling and clustering discussed in Sec.~\ref{sec:Spheres} and the angular interpolation  presented in Sec.~\ref{sec:Angle}. 
We also use replicate padding in the networks to handle missing selections. 
We use this representation for its ability to customize the resolution on the sphere and since we found it to perform the best overall on all tasks. 
A more thorough analysis of the different sampling and clustering techniques, as well as a comparison between the interpolation techniques, is presented in Sec.~\ref{sec:Ablation}.



\subsection{Spherical Stylization}

\begin{figure*}
\begin{center}
\begin{tabular}{rcccc}
\raisebox{0.075\linewidth}{\rotatebox[origin=c]{90}{Spherical Image}} &
\includegraphics[width=.32\linewidth]{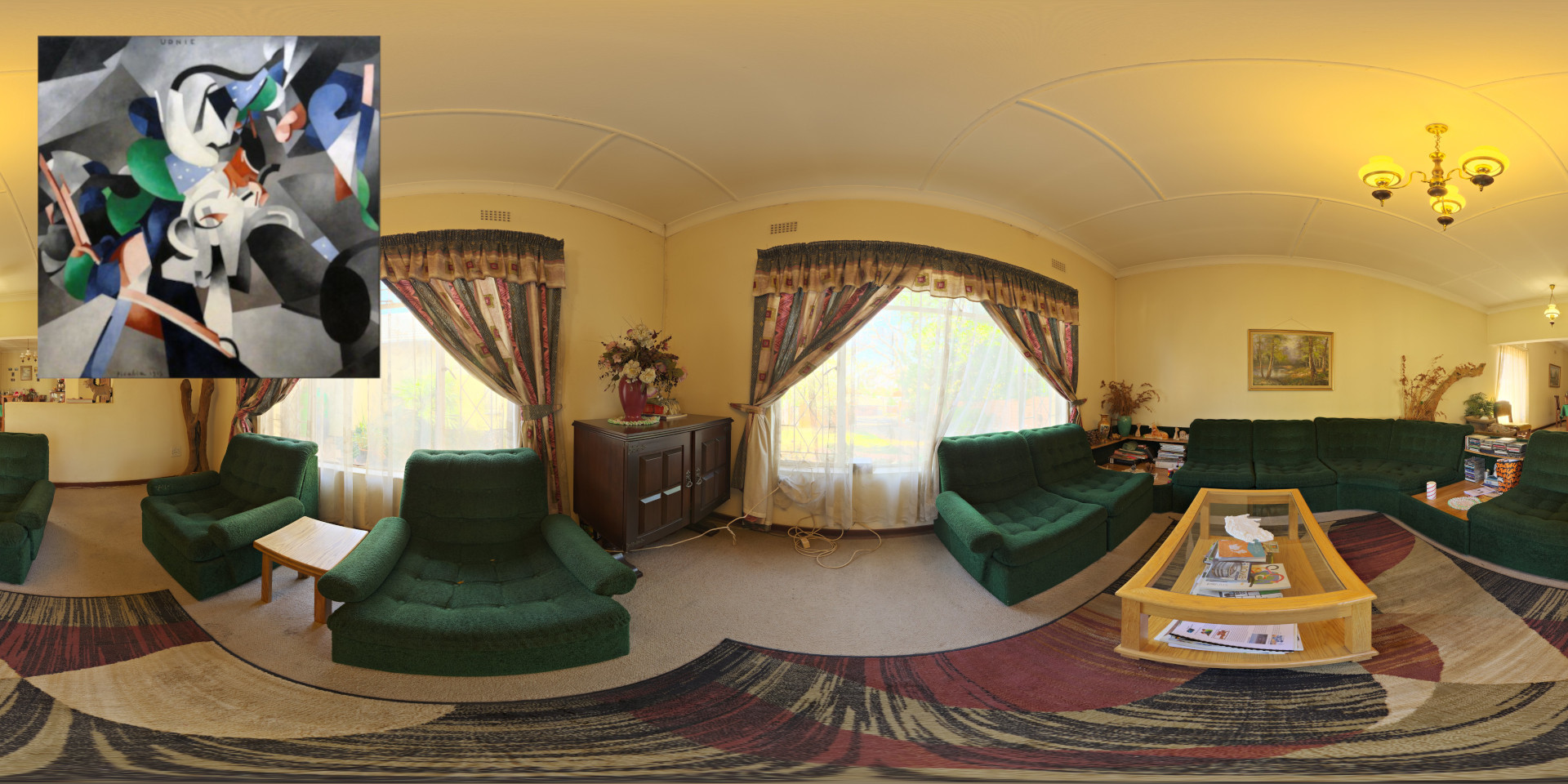} &
\includegraphics[width=.16\linewidth]{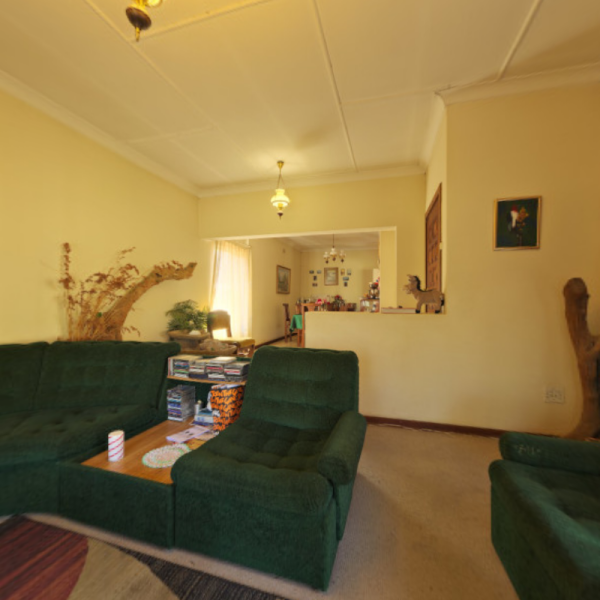} &
\includegraphics[width=.16\linewidth]{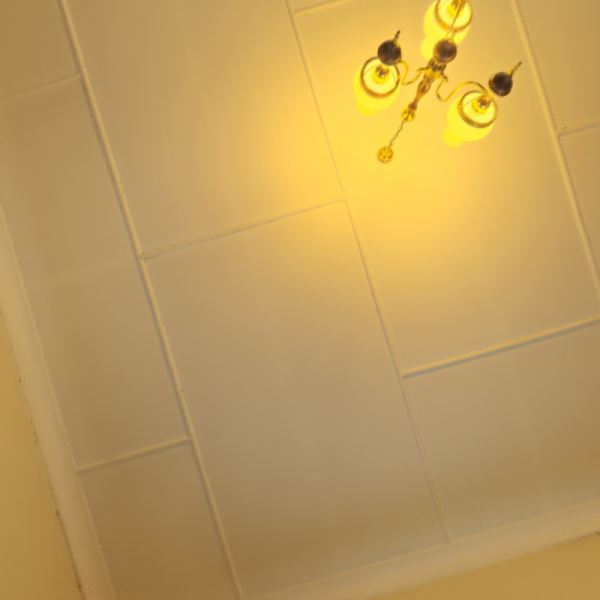} &
\includegraphics[width=.16\linewidth]{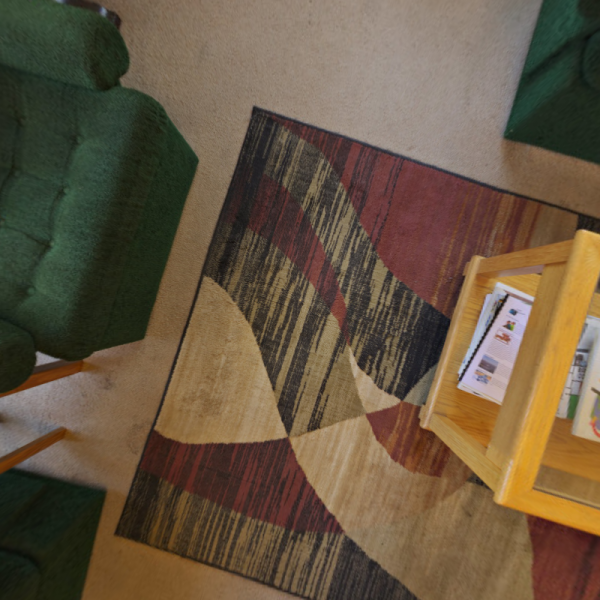}
\\
\raisebox{0.075\linewidth}{\rotatebox[origin=c]{90}{Naive}} &
\includegraphics[width=.32\linewidth]{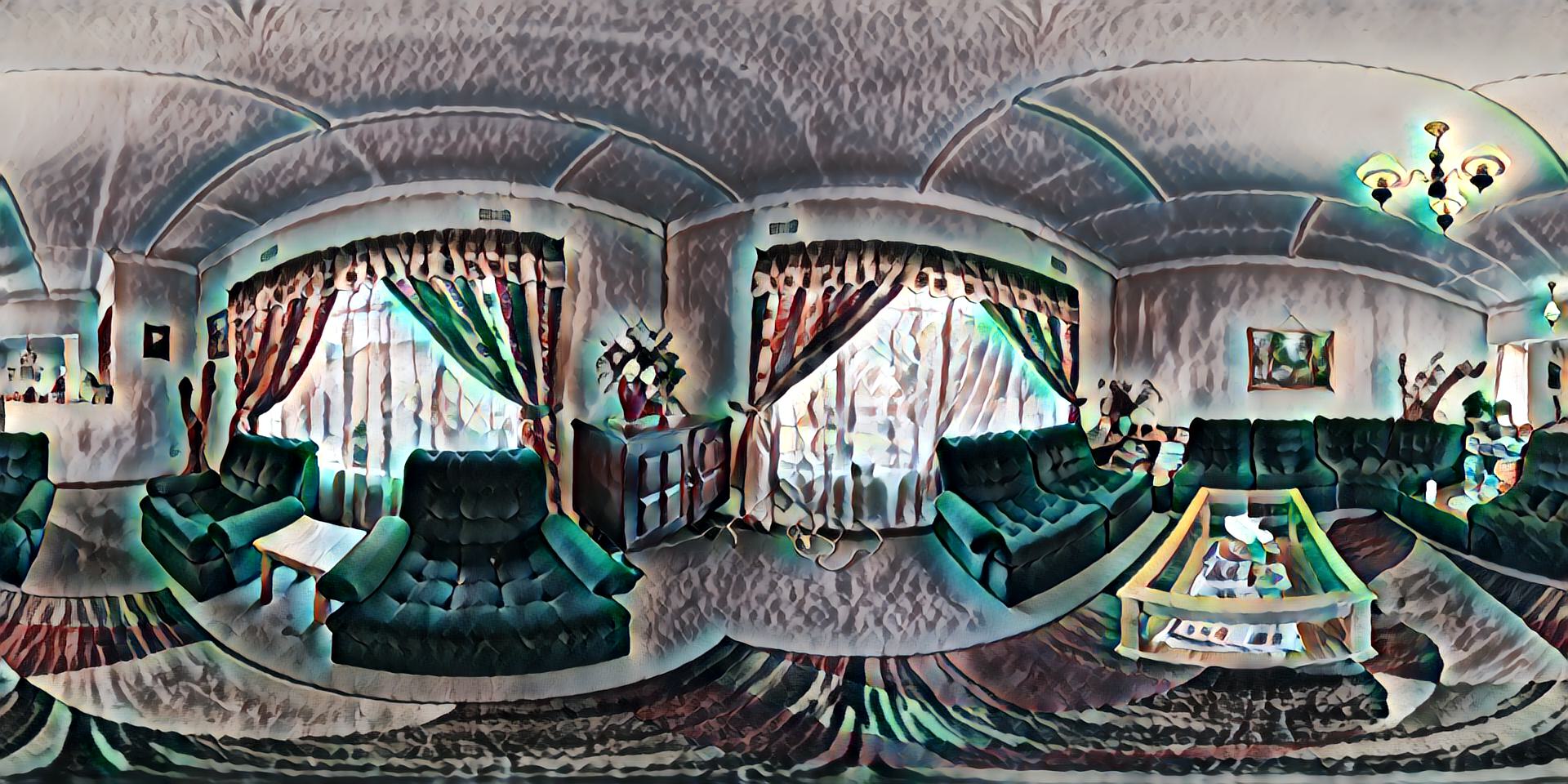} &
\includegraphics[width=.16\linewidth]{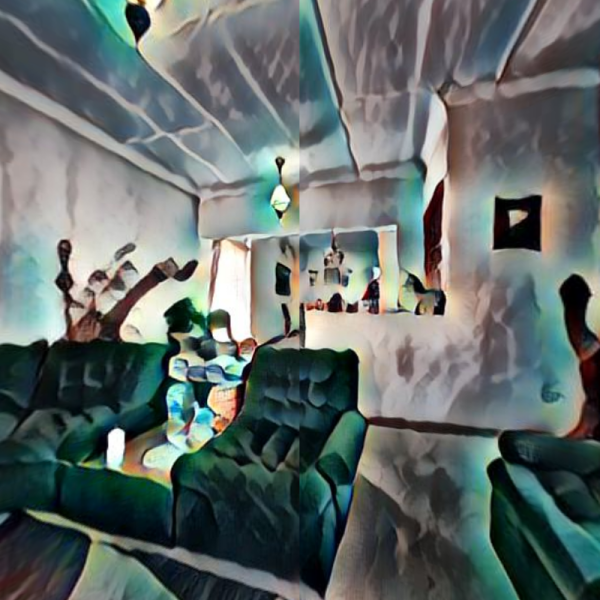} &
\includegraphics[width=.16\linewidth]{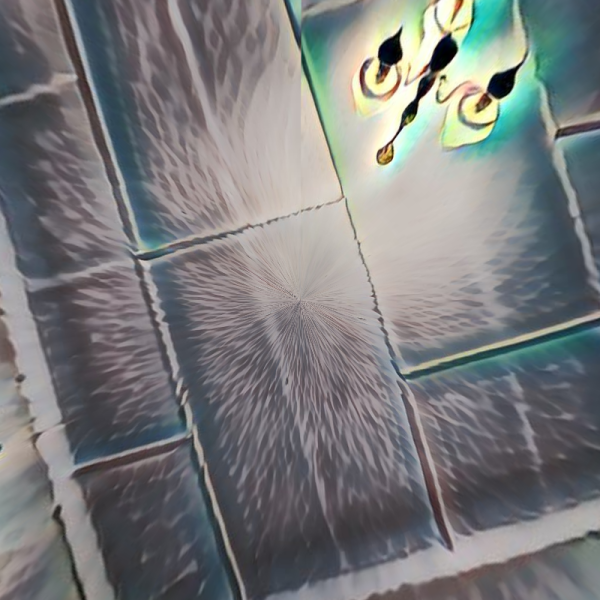} &
\includegraphics[width=.16\linewidth]{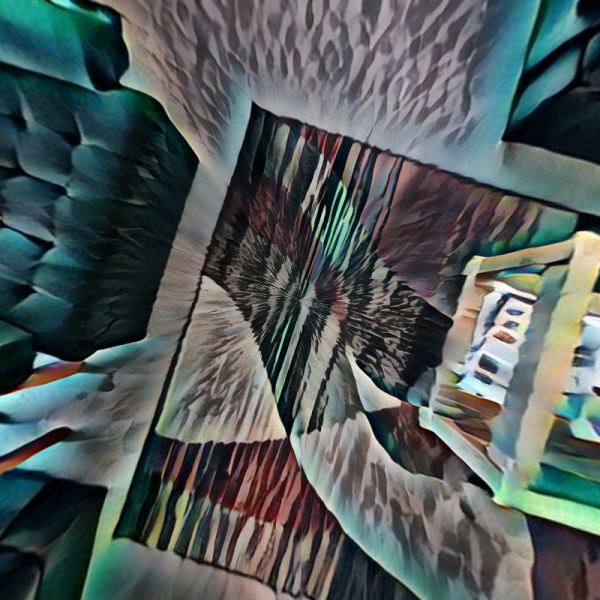}
\\ 
\raisebox{0.075\linewidth}{\rotatebox[origin=c]{90}{SelectionConv~\cite{SelectionConv}}} &
\includegraphics[width=.32\linewidth]{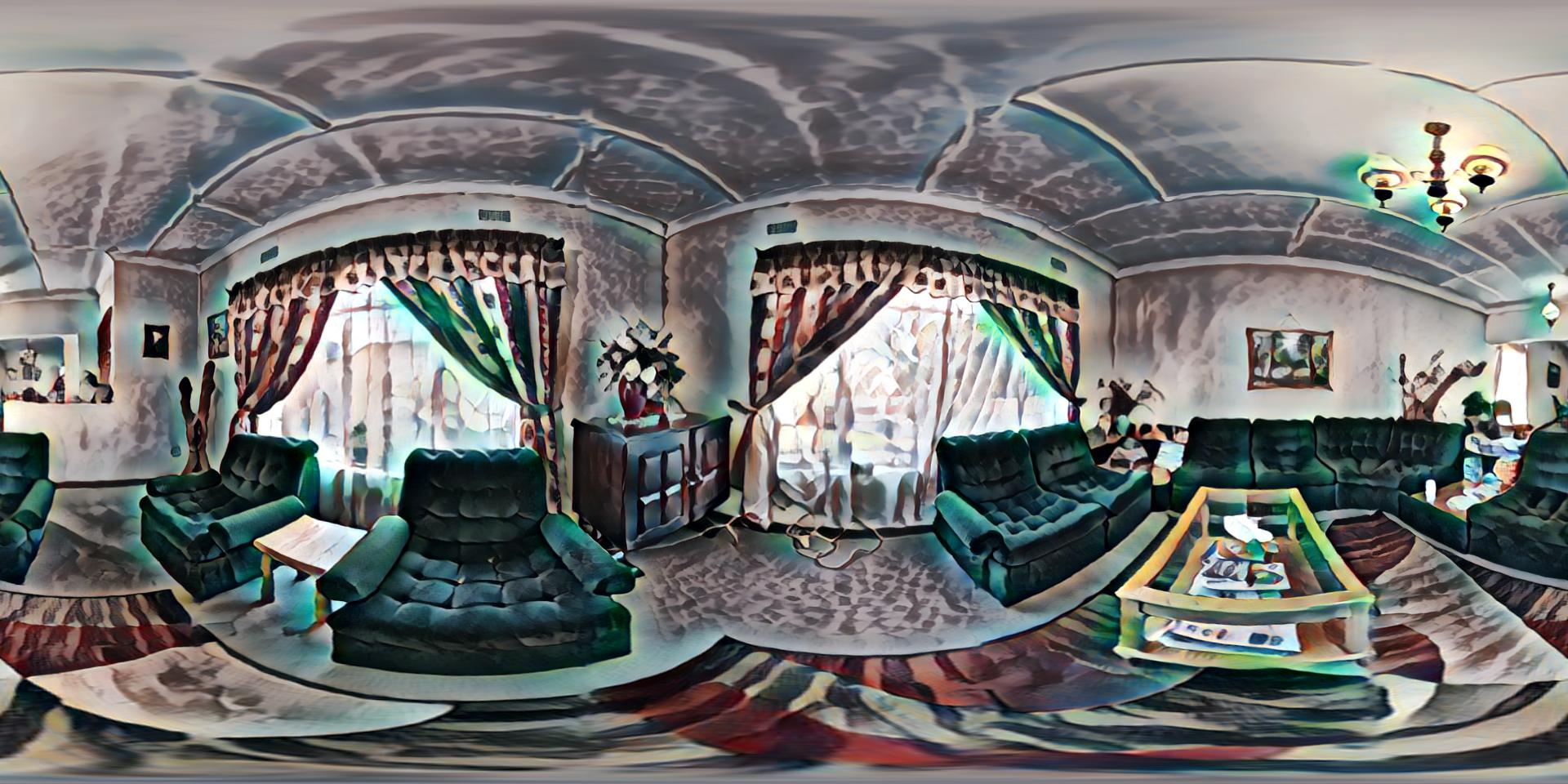} &
\includegraphics[width=.16\linewidth]{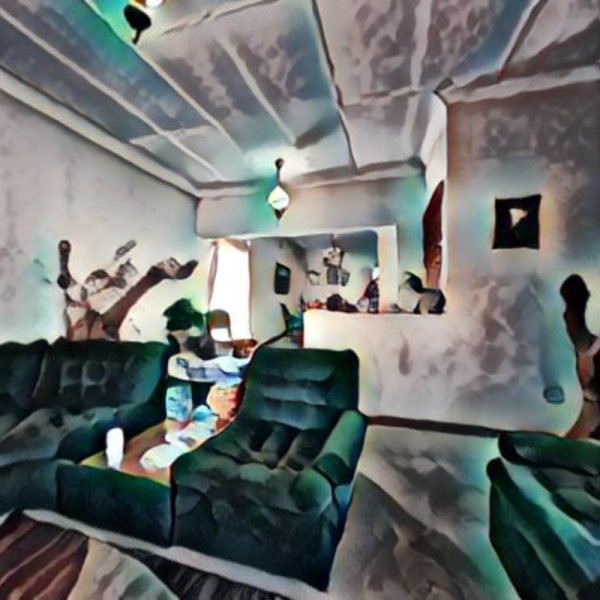} &
\includegraphics[width=.16\linewidth]{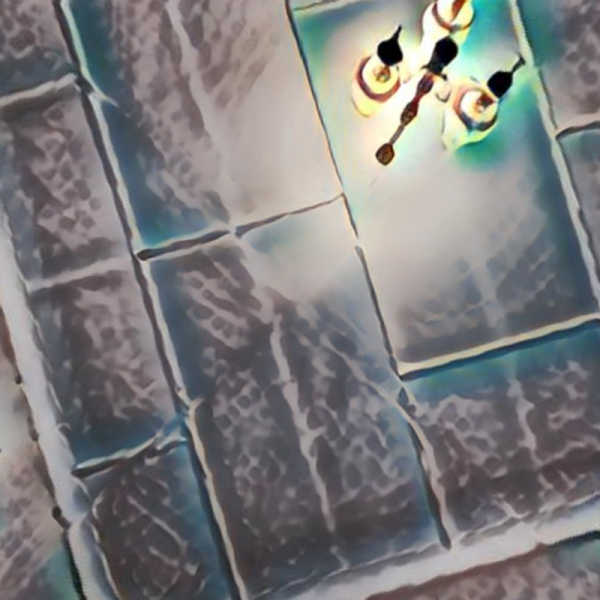} &
\includegraphics[width=.16\linewidth]{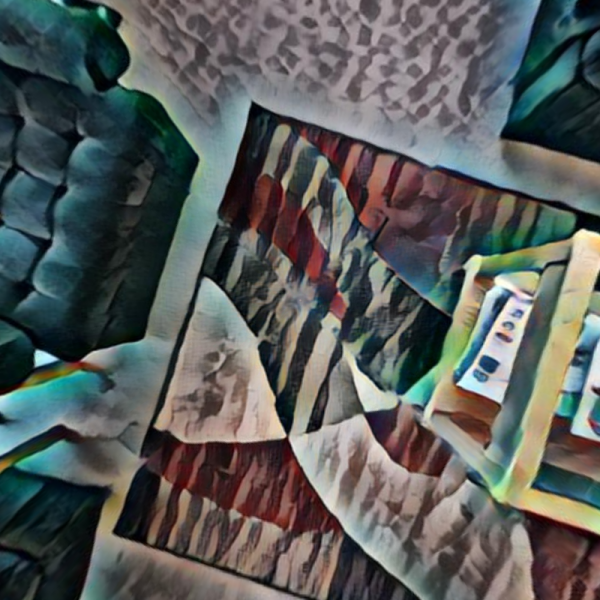}
\\ 
\raisebox{0.075\linewidth}{\rotatebox[origin=c]{90}{Ours}} &
\includegraphics[width=.32\linewidth]{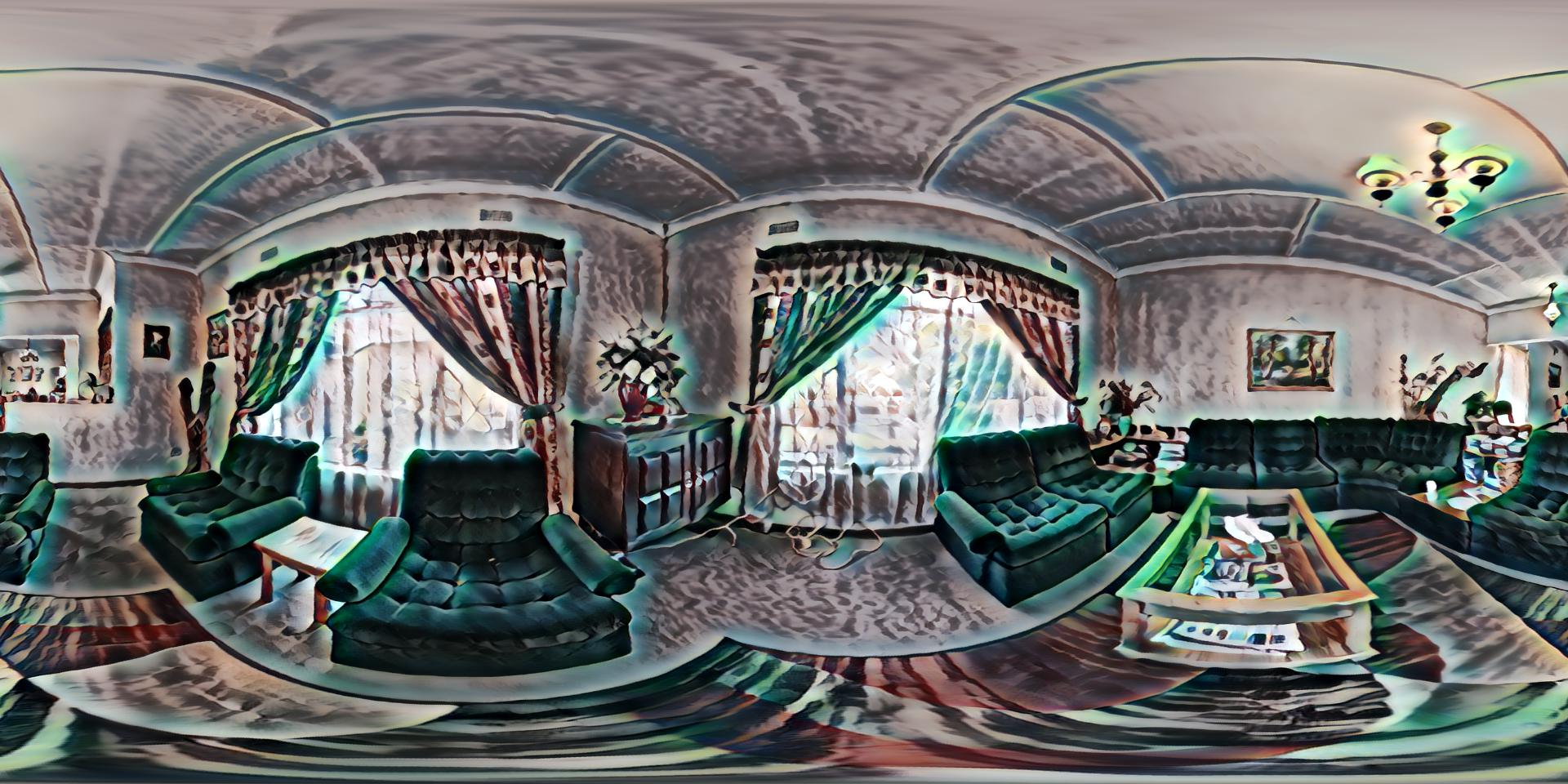} &
\includegraphics[width=.16\linewidth]{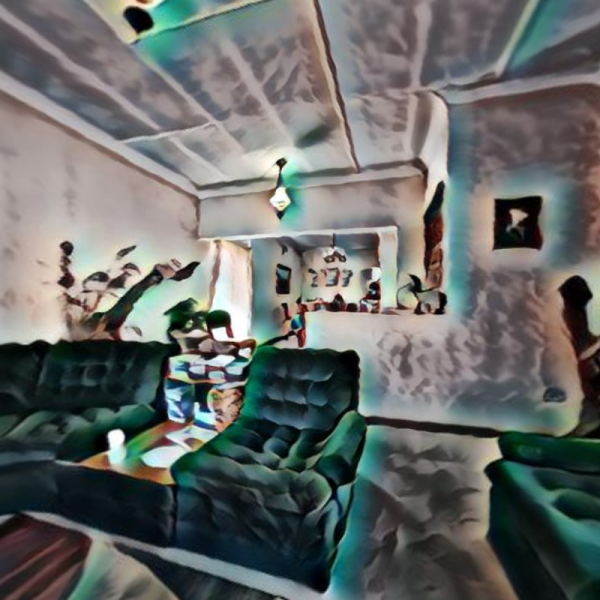} &
\includegraphics[width=.16\linewidth]{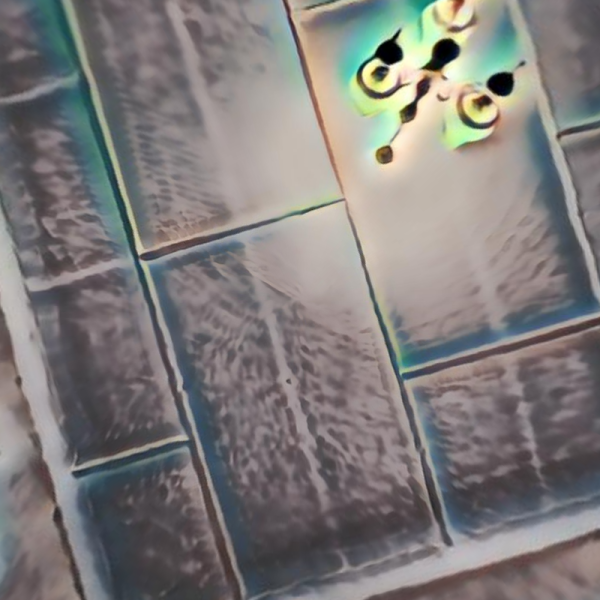} &
\includegraphics[width=.16\linewidth]{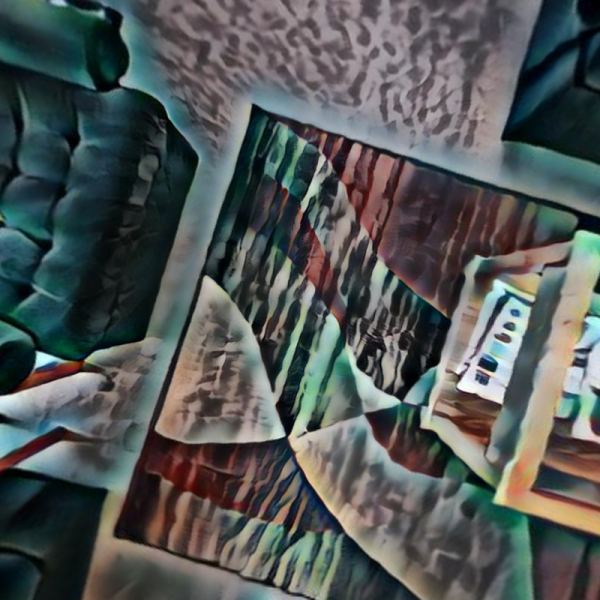} 
\\
&
Equirectangular &
Back &
Top &
Bottom
\end{tabular}
\end{center}
    \caption{A 
    360$^\circ$ image
    (1st row), its stylization when naively stylizing the equirectangular image (2nd row), using the cube-map graph setup in \cite{SelectionConv} (3rd row), and compared to our interpolated spherical representation (4th row). The equirectangular projection along with various views of the scene are presented. In the naive approach, note the vertical seam in the middle of the back view as well as the distortion in the top and bottom views. In the original SelectionConv results, note the artifacts in the top and bottom views along the seam connections (making an x shape). Those artifacts are removed with our new method. Public domain image courtesy of polyhaven.com.
    }
\label{fig:spherestyle}
\end{figure*}

We start with a clear visual example of the effectiveness of transferring weights from a 2D CNN to our graph structures without the need of any additional fine-tunning. We demonstrate style transfer with the approach proposed by Li \etal \cite{Li2019} on our spherical representations. 
SelectionConv~\cite{SelectionConv} achieved spherical style transfer by building a fully connected graph structure based on a cube map. While this approach improved the stylization over the naive method
of stylizing the equirectangular image directly, 
the uninterpolated nature of the graph can give artifacts
along pixels where a direction transition occurs. 
Our 
new
approach removes such artifacts, as shown in Fig.~\ref{fig:spherestyle}. This is possible because 
we 
can smoothly vary the selection between nodes in 3D space. 
Additionally, when reconstructing the equirectangular image, the relevant points are interpolated in 3D space, which gives a smooth and plausible stylization throughout the whole sphere.
Our new method is also similar to \cite{SelectionConv} in run time and memory usage.



\begin{figure}
\vspace{0.2in}
\begin{center}
\begin{tabular}{cccc}
\hspace{-0.12in}
\includegraphics[width=.21\linewidth]{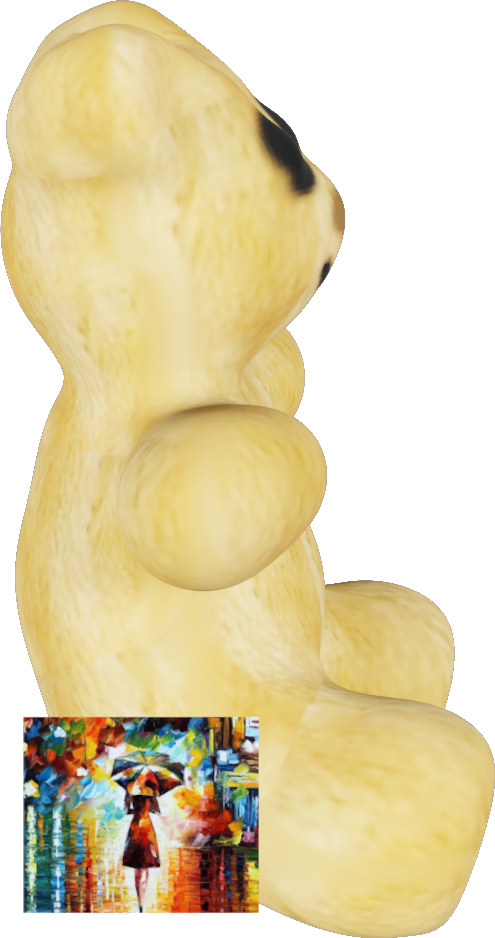} &
\includegraphics[width=.21\linewidth]{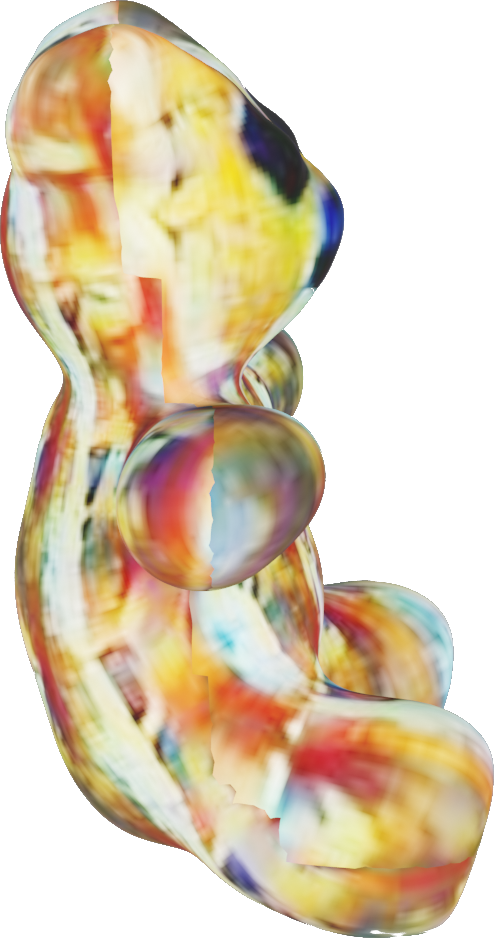} &
\includegraphics[width=.21\linewidth]{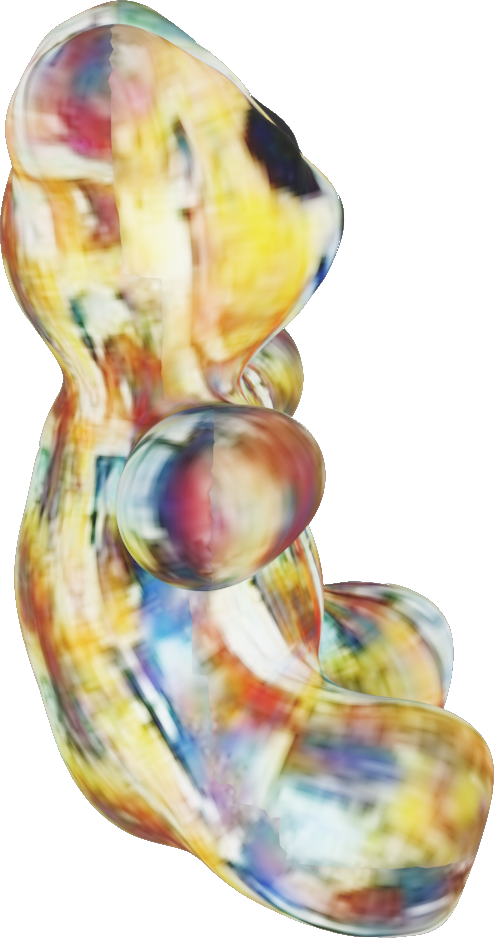} &
\includegraphics[width=.21\linewidth]{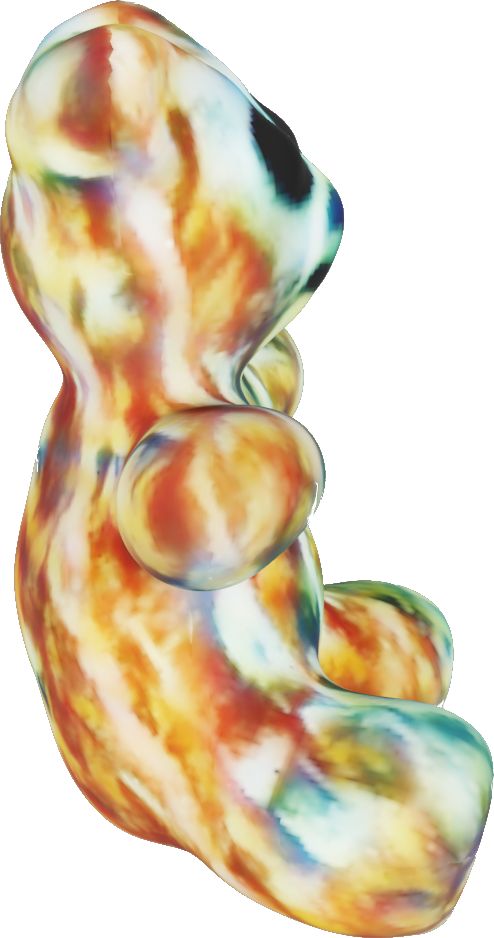} 
\\
a) Mesh &
b) Naive &
c) \cite{SelectionConv} &
d) Ours
\end{tabular}
\end{center}
   \caption{ 
   When the texture map of the original mesh (a) is stylized naively (b), many artifacts are present along the UV seams. Stylizing with SelectionConv (c) removed some of those artifacts, but inconsistencies remain. 
   Interpolated SelectionConv 
   (d) retains far greater consistency along seams.  
   }
\label{fig:teddystyle}
\end{figure}

\begin{figure}
\begin{center}
\begin{tabular}{ccc}
\includegraphics[width=.30\linewidth]{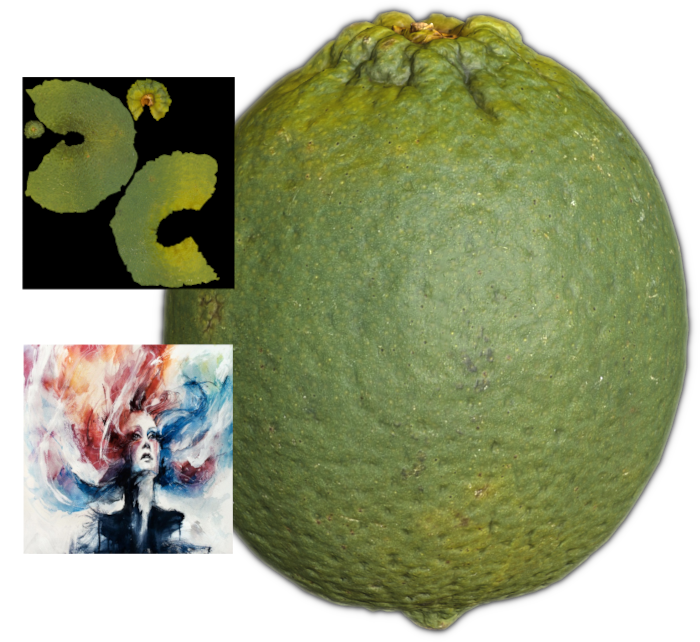} &
\includegraphics[width=.24\linewidth]{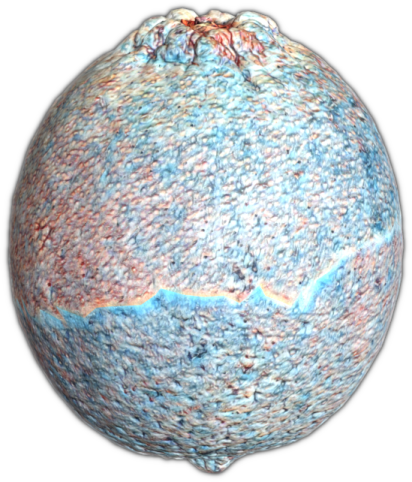} &
\includegraphics[width=.24\linewidth]{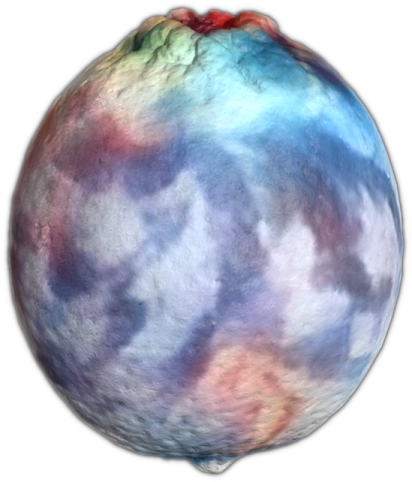} 
\\
\includegraphics[width=.32\linewidth]{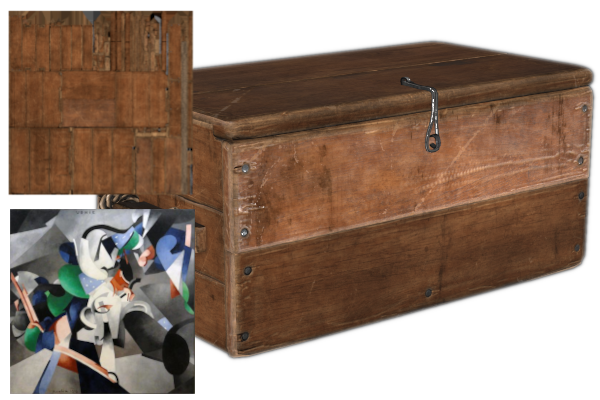} &
\includegraphics[width=.26\linewidth]{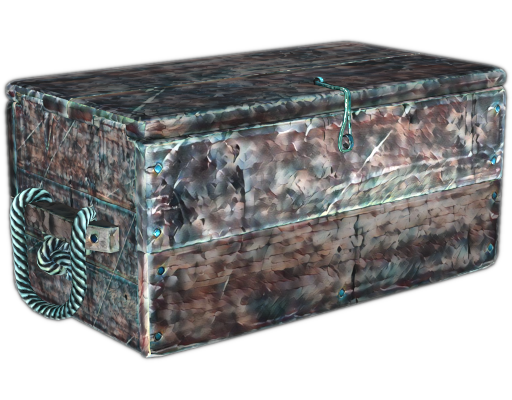} &
\includegraphics[width=.26\linewidth]{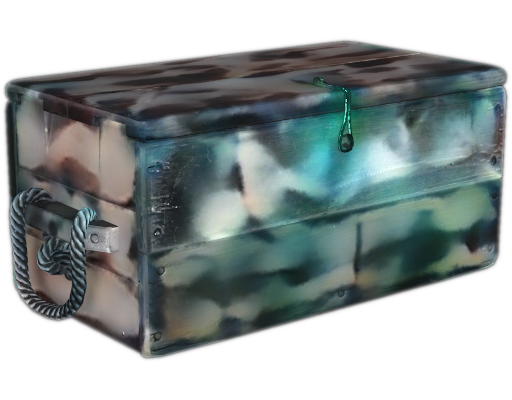} 
\\
a) Mesh &
b) Naive &
c) Ours
\end{tabular}
\end{center}
   \caption{ Example stylizations of high quality meshes with 4K textures (a). Naively stylizing the texture map (b) is slow, leaves artifacts on seams, and provides little control for the level of detail for the stylization. Our approach (c) removes seam artifacts while allowing the user to control the number of sampling points, which in turn determines the speed and detail of the stylization. Public domain meshes courtesy of polyhaven.com. }
\label{fig:meshstylization}
\end{figure}

\subsection{Spherical Segmentation}

A common spherical image task is omnidirectional semantic segmentation. We evaluate our method on such a task using the Stanford 2D-3D-S dataset \cite{Armeni2017}. It consists of equirectangular images of various indoor scenes with 13 different classes. 

We first train a simple 6-layer deep U-Net architecture~\cite{UNet}, similar to the architecture used in \cite{Chiyu2019,Zhang2019}, 
and include depth information with the input to be consistent with state-of-the-art methods. 
We train on the provided precomputed views of the spherical image. Once the network has been trained in 2D space, we transfer over the weights to our graph representation with interpolated selections. We use the layering approach and scale the resolution to match the average $\Delta \theta$ from the training data.

We can also fine-tune the graph representation on the full 3D data. These fine-tuning steps can be completed quickly since our graph structure only needs to be calculated once, then it can be reused in every successive iteration. Only the node feature values need to be updated.

The mean IOU for our method, both before and after fine-tuning in spherical space, is shown in Table~\ref{tab:segmentation}, along with comparisons to other previous works. We evaluate on the full scale equirectangular images ($4096 \times 2048$) and follow the standard practice of weighting the metrics for each pixel by cosine of their latitude location, similar to \cite{Chiyu2019,Zhang2019}. 

As is shown, our method has performance comparable to state-of-the-art approaches, even after a direct transfer without fine-tuning. While \cite{Zhang2019} and \cite{Eder2020} have higher mean IOU scores,
we note that \cite{Zhang2019} must use a fine-tuned icosphere representation 
and that \cite{Eder2020} requires processing 80 different $512 \times 512$ planar projections to make a single spherical image, whereas our approach can evaluate in a single pass with a reusable graph structure between spherical images.


\begin{table}[]
    \centering
    \begin{tabular}{|cc|cc|}
        \hline
        Method & Input & mIOU & fine-tuned \\
        \hline
        Cubemap \cite{SelectionConv} & RGB & 36.3\% & - \\
        UGSCNN \cite{Chiyu2019} & RGB-D & - & 38.3\%   \\
        GaugeNet \cite{Cohen2019} & RGB-D & - & 39.4\%   \\
        Sphere (Ours) & RGB-D & 39.9\%  & \textbf{41.4\%}  \\
        HexRUNet \cite{Zhang2019} & RGB-D & - & 43.3\%  \\
        TangentIms \cite{Eder2020} & RGB-D & 38.9\% & 51.9\% \\
        \hline
    \end{tabular}
    \caption{Mean intersection-over-union (mIOU) on the Stanford 2D-3DS dataset segmentation task \cite{Armeni2017}. The mIOU is shown for both direct transfer (left) and after fine-tuning (right). Our method performs comparably to state-of-the-art approaches.}
    \label{tab:segmentation}
\end{table}

%
%



\subsection{Mesh Stylization} \label{sec:MeshResults}

Lastly, we demonstrate the ability of our method to generalize to general surfaces by stylizing 3D meshes.
In \cite{SelectionConv}, a UV edge-pairing process was required to determine edges across seams in the texture map. This process failed to remove all artifacts along seam boundaries and was time consuming for high-polygon-count meshes. 
Our 
new 
approach avoids this computationally expensive step while achieving far better consistency along UV seams
since the edges and selections are made in 3D space, rather than on the topologically complex 2D texture map. A comparison of 
this
approach to the previous work is shown in Fig.~\ref{fig:teddystyle}. Additionally, our method allows users to control the number of initial sampling points, making it time invariant to the complexity of the mesh. This also provides a simple way for controlling the level of detail in the stylization on the mesh
while operating on high quality meshes and texture maps.
Examples of these benefits are shown in Fig.~\ref{fig:meshstylization} 
and in the supplemental material. 


\section{Ablation Study}\label{sec:Ablation}

\begin{table*}[]
    \centering
    \begin{tabular}{|c|ccccc|}
        \hline
        Sampling\textbackslash Clustering & Random & Equirectangular & Spiral & Icosphere & Layering \\
        \hline
        Equirectangular & 35.1\% / 35.0\%  & 38.1\% / 37.0\% & 38.9\% / 38.8\% & 38.7\% / 38.5\% & 40.7\% / 39.8\%  \\
        Spiral & 32.8\% / 32.5\%  & 32.8\% / 32.3\% & 35.7\% / 35.9\% & 36.3\% / 36.2\% & 37.9\% / 37.2\% \\
        Icosphere &  31.9\% / 30.8\% & 33.3\% / 31.9\% & 35.9\% / 35.4\%  & 36.5\% / 35.6\% & 38.1\% / 36.7\% \\
        Layering &  34.3\% / 33.6\% & 36.2\% / 34.3\% & 38.1\% / 37.5\% & 38.8\% / 38.2\%  & \textbf{39.9\%} / 38.7\%  \\
        \hline
    \end{tabular}
    \caption{Ablation study comparing mIOU results for the various sampling, clustering, and interpolation approaches on the Stanford2D-3D-S segmentation task. Angle-based interpolation results are shown on the left. Barycentric results are shown on the right. Results for random clustering were averaged over multiple configurations.}
    \label{tab:ablation}
\end{table*}


We presented the layering sampling and clustering technique (with angular interpolation) as our best representation of the sphere. However, one of the benefits of our method is that we are not constrained to any specific graph structure. Thus, we give the performance of the other possible representations of our sphere. 

We test the various sampling methods and clustering methods with the previously presented segmentation task without fine-tuning. Additionally, we compare barycentric interpolation versus angle-based interpolation for these various representations. The results are shown in Table~\ref{tab:ablation}.

The results show some interesting patterns. First, all methods perform worst when using random and equirectangular clustering. This is to be expected since the points will not be pooled together based on equidistant locations. Second, we note that spiral and icosphere sampling perform poorest overall since the spiral approach generates neighborhoods that tend to be more irregular and neither allows for customizing the resolution in a precise way. (We also note that icosphere performs more poorly than demonstrated elsewhere, but we attribute this to the fact that we do not explicitly modify the convolution operator to be hexagonal like many icosphere-specific methods~\cite{Lee2019,Zhang2019}.) Also,  barycentric interpolation performs slightly below  angle-based interpolation for most sampling and clustering representations. We attribute this to the blurring effect that comes from the larger number of edges, especially when compared to the angle-based representation when most points are reasonably spaced. Lastly, we note that layering clustering did the best for all methods. Equirectangular sampling with layering clustering even outperformed the layering/layering representation on the segmentation task, but we note that it used 25\% more nodes and edges, which makes it slower to run and harder to fine-tune.



\section{Conclusion}

We have presented a general framework for transferring weights from 2D convolutional networks to graph networks that can operate on spherical images and surfaces. These networks can be fine-tuned even further in their specific domains as needed. This approach allows for a simple and effective way to improve performance on spherical tasks without requiring large datasets that are specific to the spherical domain. We have also demonstrated possible applications for 3D meshes and have provided a thorough ablation study exploring sampling of spherical surfaces. \\

\noindent \textbf{Limitations}

The main limitation of our approach comes from the extra memory needed to store the graph edges while performing the convolution. Though we describe our method mathematically using adjacency matrices, the implementation actually uses memory-efficient edge indexes. However, this means that allowing multiple selections for the same edge increases the size of the edge index. Additionally, since the interpolation values also need to be stored, the memory needed for a graph with many nodes can become quite large. Since our focus was on direct transfer, this is usually not an issue and most graphs can still fit in GPU memory, but fine-tuning any task must be done with a very small batch size to be feasible.

Like the original SelectionConv, we can't use larger convolution kernel sizes because the edge hopping process is not differentiable. Thus the networks we used contained only $3 \times 3$ and $1 \times 1$ kernels. We note, however, that this limitation is shared with other previous methods~\cite{Chiyu2019, Zhang2019}. \\

\noindent \textbf{Future Work}

Though we demonstrated style transfer on meshes, the field of meshes and general surfaces is extensive and has many applications. More exploration and experiments using our interpolated SelectionConv in this area could be the study of future work. Looking into additional interpolation schemes and techniques for determining expected radial distances is also worth investigation.


\end{document}


\title{Interpolated SelectionConv for Spherical Images and Surfaces \\ Appendix}

\maketitle

\appendix

\section{Barycentric Interpolation Derivation}

For a point $p$ lying inside a square kernel within the neighborhood of 9 weights, there are 8 possible triangles that it may be contained within. Each triangle, however, is a right triangle with two bases with length $d$. Thus, using simple substitution, we can solve for a single triangle and apply the results to any of the 8. We represent our triangle with the three vertices $(0,0)$, $(d,0)$ and $(d,d)$, while representing the transformed point as the coordinate ($u$,$v$). To convert any point $p$ (relative to center of the kernel) to this coordinate system, we first take the $|p|$ so that it lies within the first quadrant, then given that it must be the case that $u \geq v$, we get the following substitution:

\begin{equation}
    u = \text{max}(|p|)
\end{equation}

\begin{equation}
    v = \text{min}(|p|)
\end{equation}

\begin{figure}
\begin{center}
\includegraphics[width=0.80\linewidth]{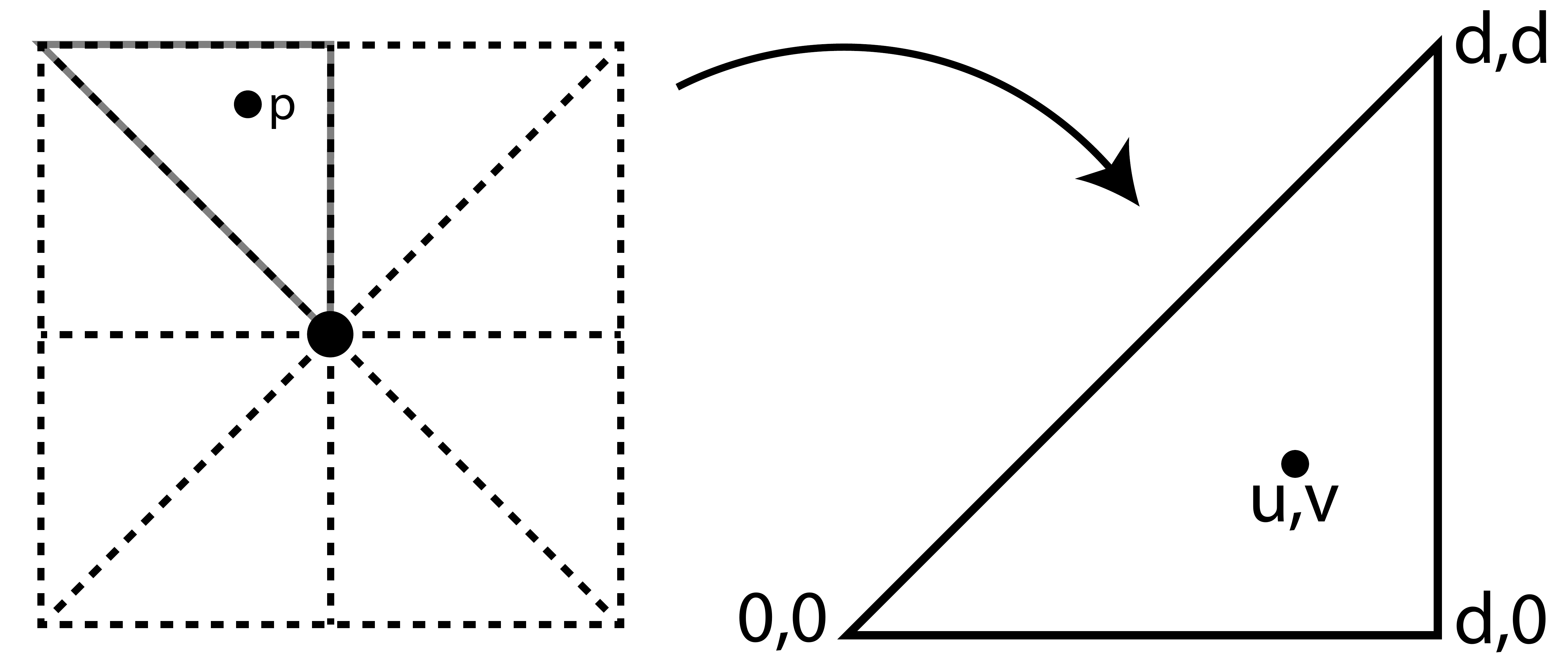}
\end{center}
   \caption{Regardless of which triangle a point $p$ lands within kernel space, it can be substituted for a simple x-axis aligned isosceles right triangle.}
\label{fig:triangle}
\end{figure}

\noindent This simple substitutions works for points in any of the 8 original triangles, as shown in Fig.~\ref{fig:triangle}.

Finding the weights in barycentric interpolation can be expressed as the solving of a system of linear equations. Specifically,

\begin{equation}
    \begin{bmatrix}
        x_0 & x_a & x_b\\
        y_0 & y_a & y_b\\
        1 & 1 & 1
    \end{bmatrix}
    \begin{bmatrix}
        w_0 \\ w_a \\ w_b
    \end{bmatrix}
    =
    \begin{bmatrix}
        x_p \\ y_p \\ 1
    \end{bmatrix}
\end{equation}

\noindent where $w_n$ are the desired interpolation weights we are solving for. This system becomes drastically simplified when we substitute the fixed triangle and our substituted coordinates. Our system becomes

\begin{equation}
    \begin{bmatrix}
        0 & d & d\\
        0 & 0 & d\\
        1 & 1 & 1
    \end{bmatrix}
    \begin{bmatrix}
        w_0 \\ w_a \\ w_b
    \end{bmatrix}
    =
    \begin{bmatrix}
        u \\ v \\ 1
    \end{bmatrix}
\end{equation}

\noindent The inverse of this matrix is:

\begin{equation}
    \begin{bmatrix}
        -1/d & 0 & 1\\
        1/d & -1/d & 0\\
        0 & 1/d & 0
    \end{bmatrix}
\end{equation}

\noindent Multiplying this out to solve the system gives us the following:

\begin{equation}
    \begin{matrix}
    w_0 & = & -u/d  &  & +1 \\ 
    w_a & = & u/d & -v/d & \\
    w_b & = & & v/d &
    \end{matrix}
\end{equation}

\noindent Substituting the original point back in, we get our final result:

\begin{equation}
w_0 = 1 - \frac{\text{max}(|p|)}{d}
\end{equation}
\begin{equation}
w_a = \frac{\text{max}(|p|) - \text{min}(|p|)}{d}
\end{equation}
\begin{equation}
w_b = \frac{\text{min}(|p|)}{d}
\end{equation}

\section{Spherical Style Transfer Results}

\begin{figure*}
\begin{center}

\includegraphics[width=.20\linewidth]{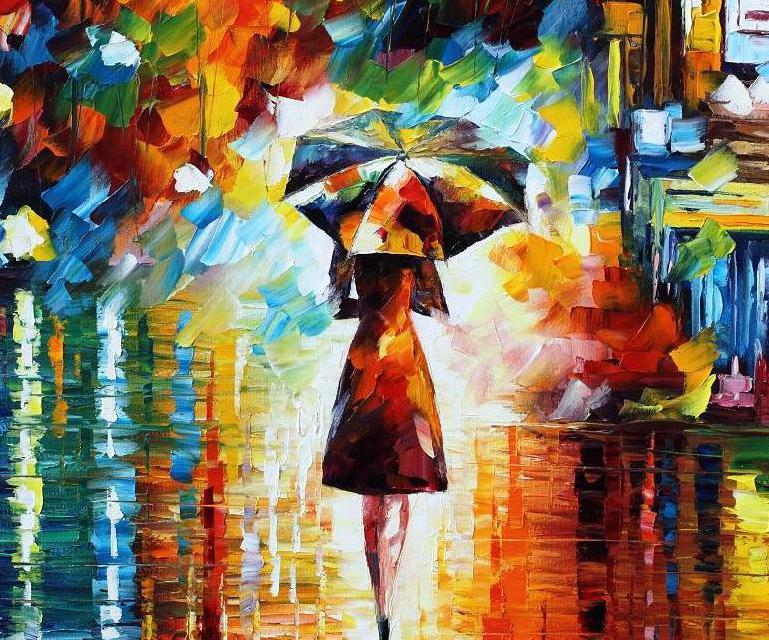}
\vspace{.5in}

\begin{tabular}{rcccc}
\raisebox{0.075\linewidth}{\rotatebox[origin=c]{90}{Spherical Image}} &
\includegraphics[width=.34\linewidth]{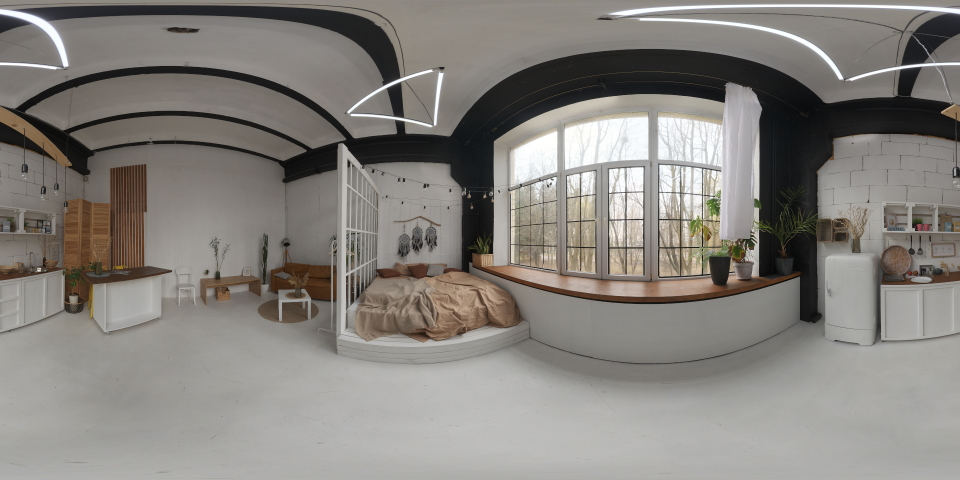} &
\includegraphics[width=.17\linewidth]{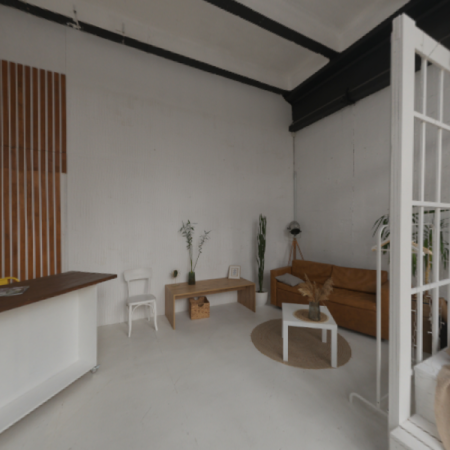} &
\includegraphics[width=.17\linewidth]{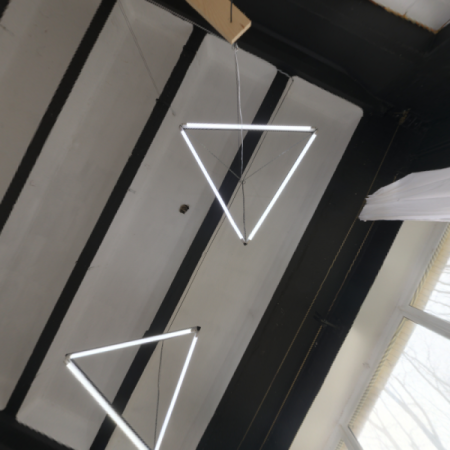} &
\includegraphics[width=.17\linewidth]{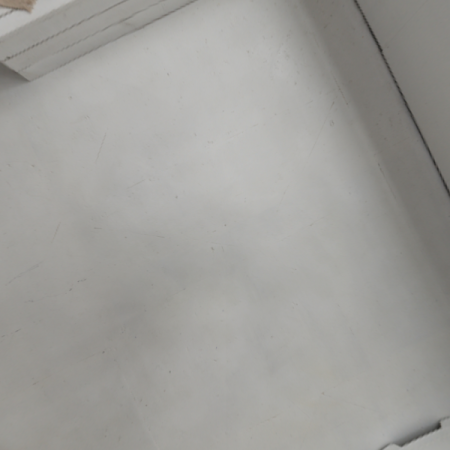}
\\
\raisebox{0.075\linewidth}{\rotatebox[origin=c]{90}{Naive}} &
\includegraphics[width=.34\linewidth]{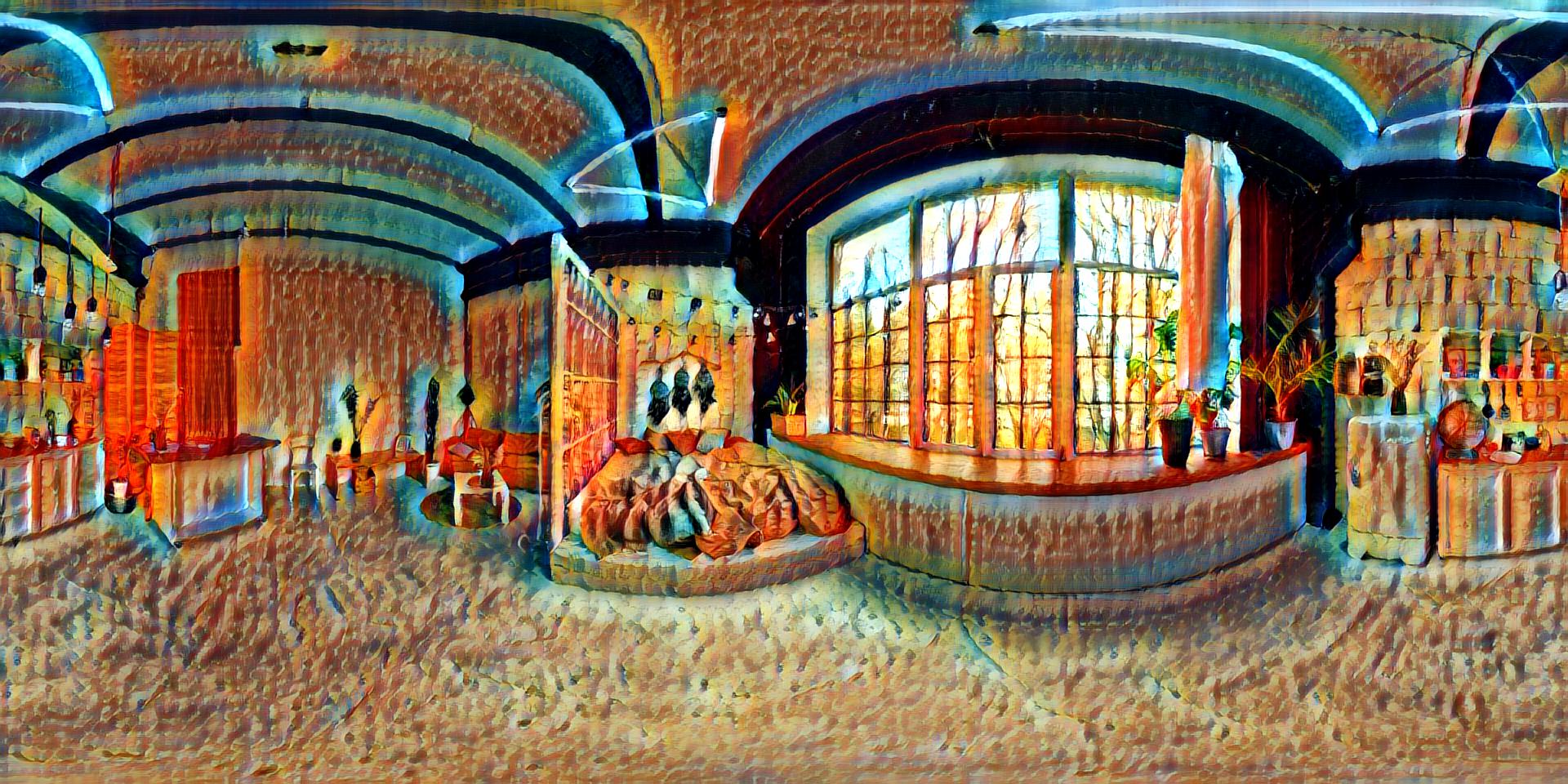} &
\includegraphics[width=.17\linewidth]{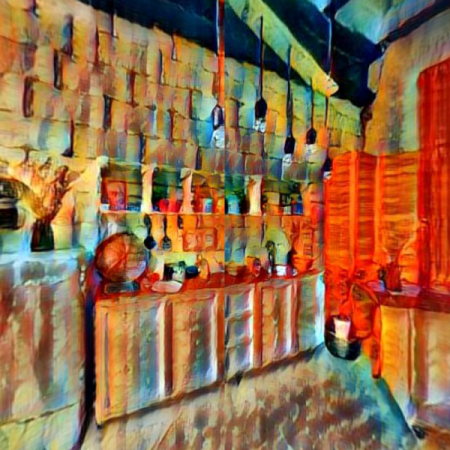} &
\includegraphics[width=.17\linewidth]{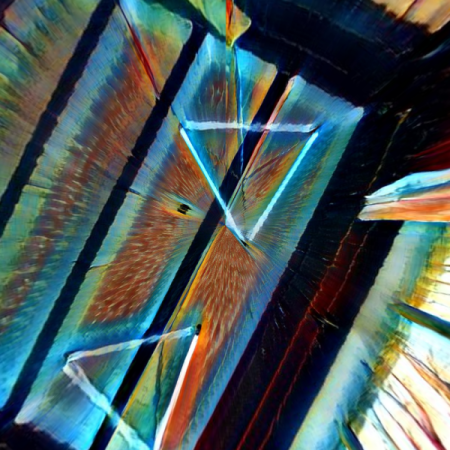} &
\includegraphics[width=.17\linewidth]{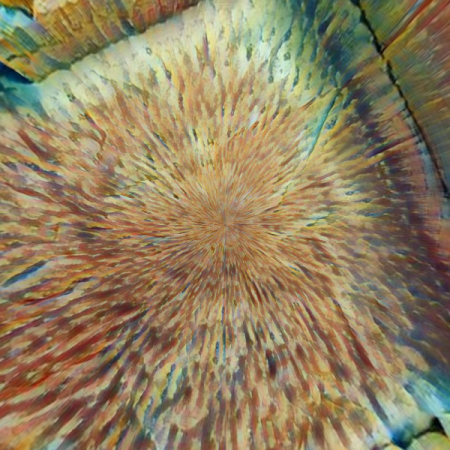} 
\\ 
\raisebox{0.075\linewidth}{\rotatebox[origin=c]{90}{SelectionConv}} &
\includegraphics[width=.34\linewidth]{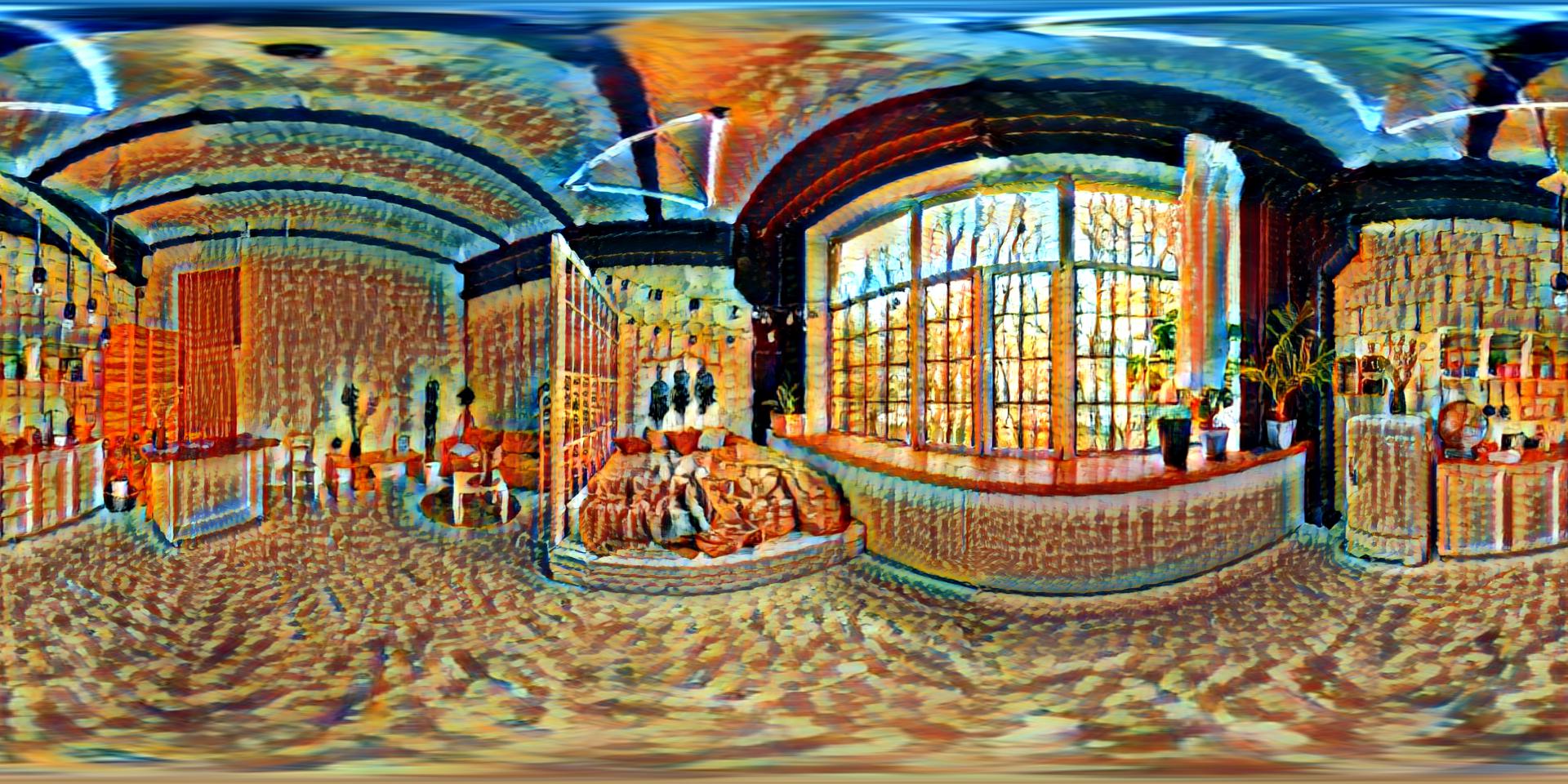} &
\includegraphics[width=.17\linewidth]{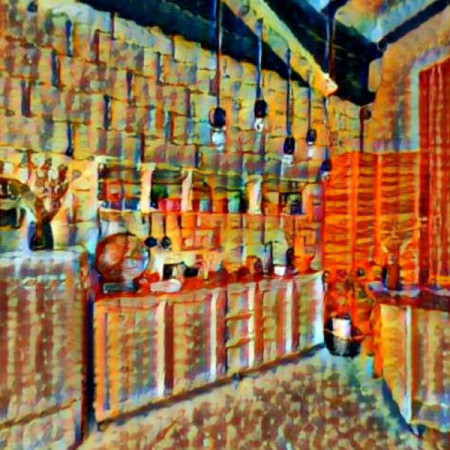} &
\includegraphics[width=.17\linewidth]{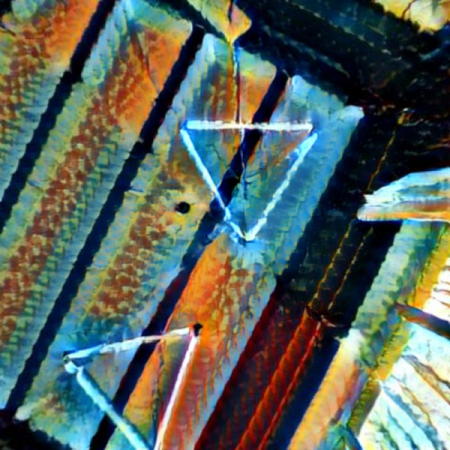} &
\includegraphics[width=.17\linewidth]{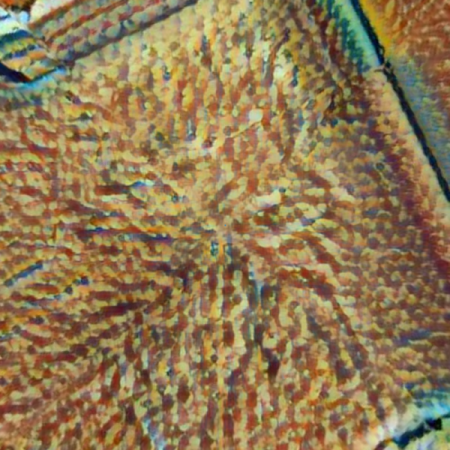} 
\\ 
\raisebox{0.075\linewidth}{\rotatebox[origin=c]{90}{Ours}} &
\includegraphics[width=.34\linewidth]{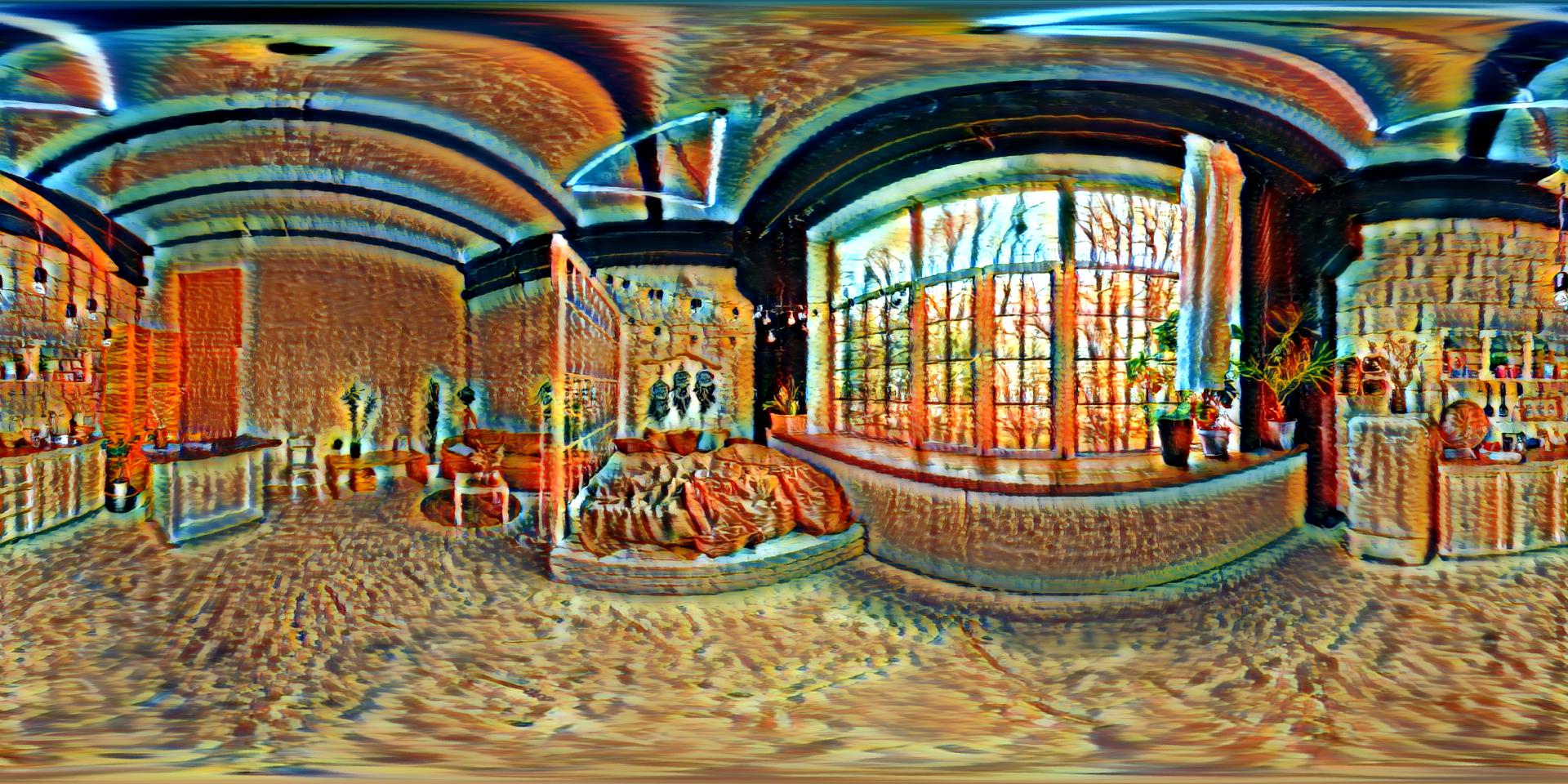} &
\includegraphics[width=.17\linewidth]{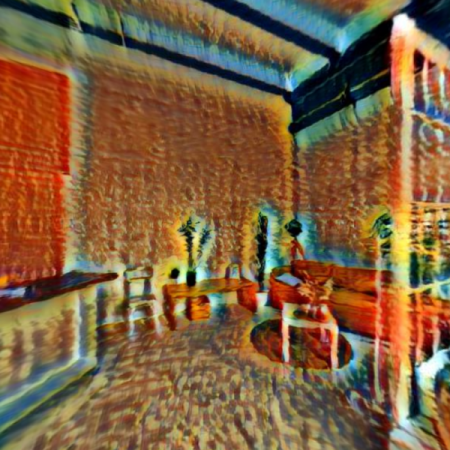} &
\includegraphics[width=.17\linewidth]{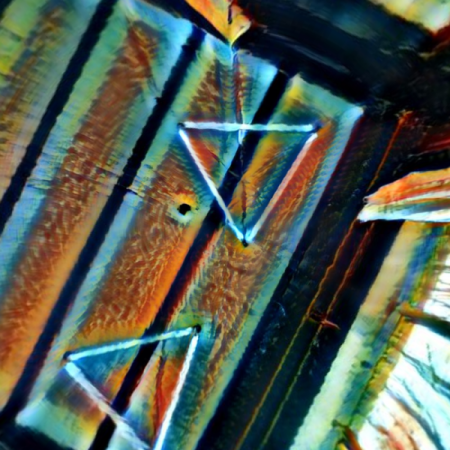} &
\includegraphics[width=.17\linewidth]{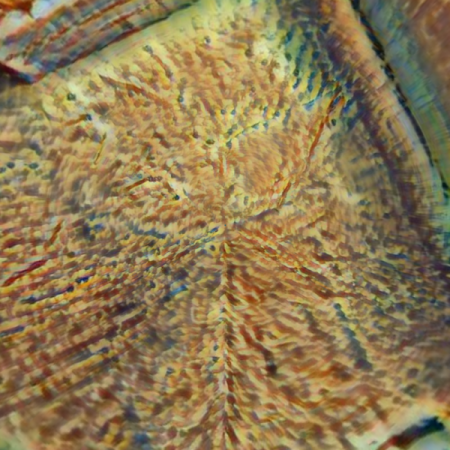} 
\\
&
Equirectangular &
Back &
Top &
Bottom
\end{tabular}
\end{center}
    \caption{A 
    360$^\circ$ image
    (1st row), its stylization when naively stylizing the equirectangular image (2nd row), using the cube-map graph setup from the original SelectionConv paper (3rd row), and compared to our interpolated spherical representation (4th row). The equirectangular projection along with various views of the scene are presented. In the naive approach, note the vertical seam in the middle of the back view as well as the distortion in the top and bottom views. In the original SelectionConv results, note the artifacts in the top and bottom views along the seam connections (making an x shape). Those artifacts are removed with our new method. Public domain image courtesy of polyhaven.com.
    }
\label{fig:spherestyle1}
\end{figure*}

\begin{figure*}
\begin{center}

\includegraphics[width=.20\linewidth]{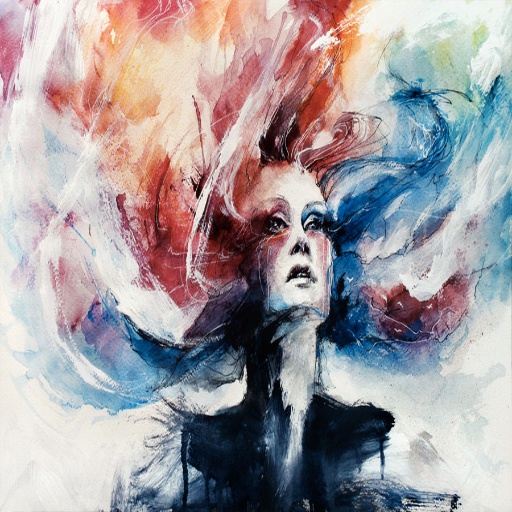}
\vspace{.5in}

\begin{tabular}{rcccc}
\raisebox{0.075\linewidth}{\rotatebox[origin=c]{90}{Spherical Image}} &
\includegraphics[width=.34\linewidth]{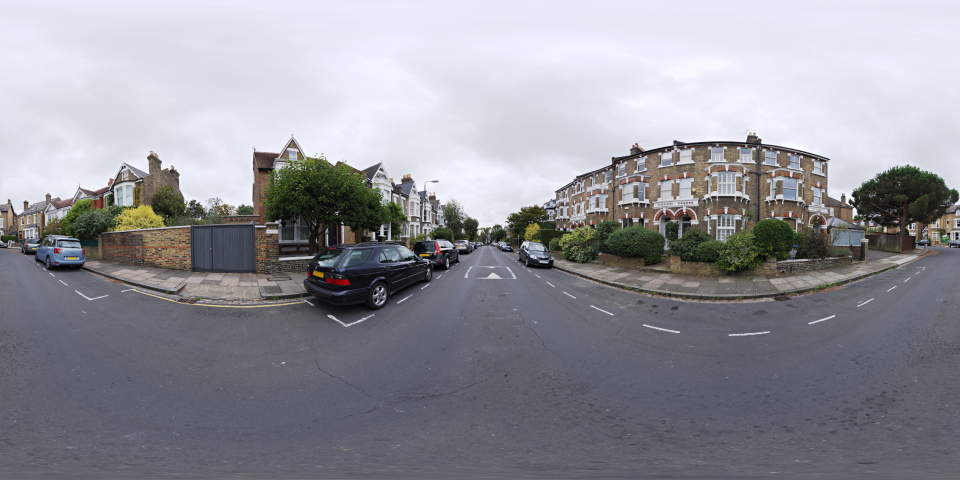} &
\includegraphics[width=.17\linewidth]{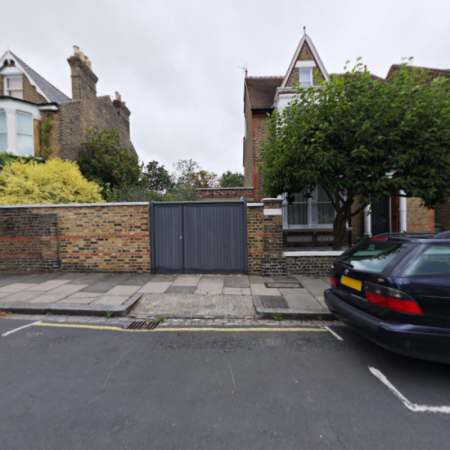} &
\includegraphics[width=.17\linewidth]{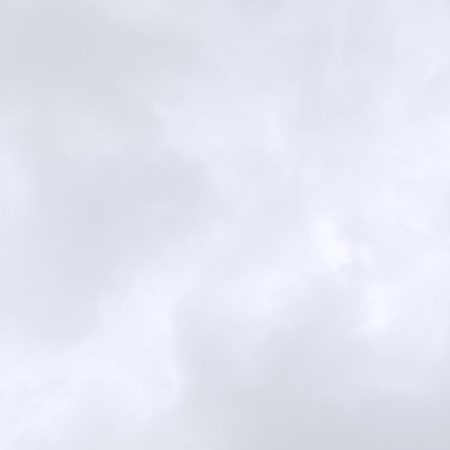} &
\includegraphics[width=.17\linewidth]{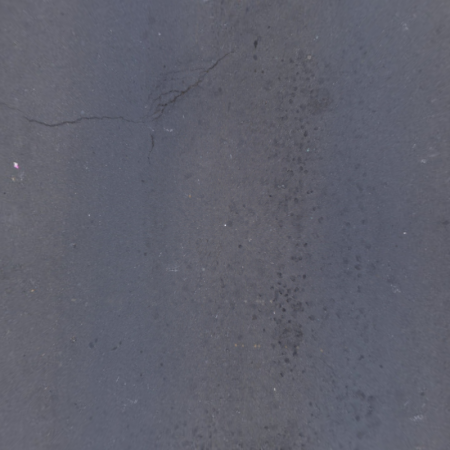}
\\
\raisebox{0.075\linewidth}{\rotatebox[origin=c]{90}{Naive}} &
\includegraphics[width=.34\linewidth]{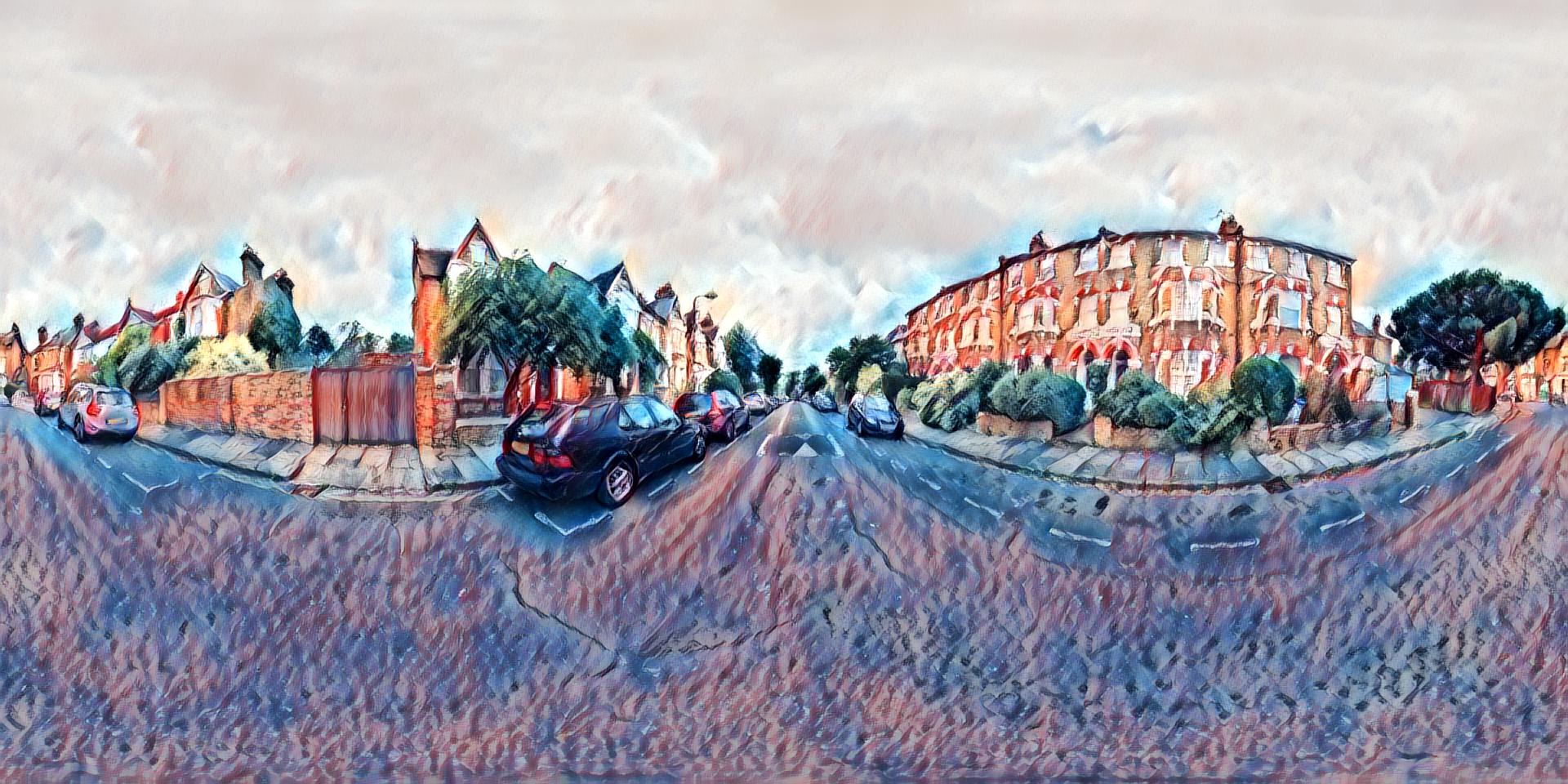} &
\includegraphics[width=.17\linewidth]{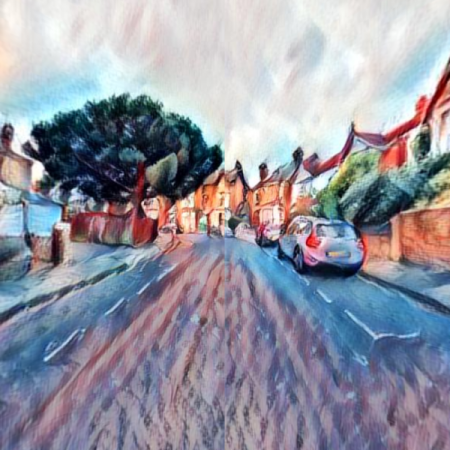} &
\includegraphics[width=.17\linewidth]{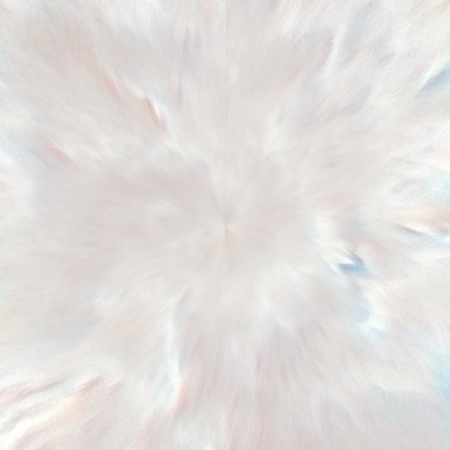} &
\includegraphics[width=.17\linewidth]{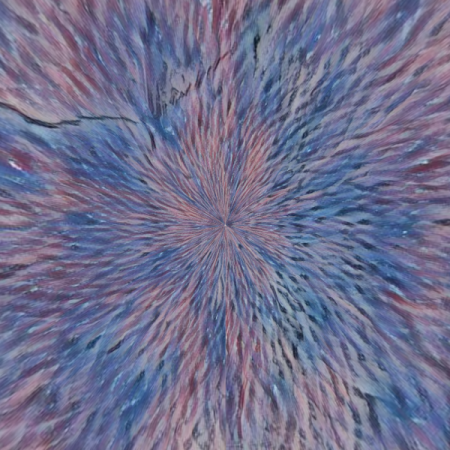} 
\\ 
\raisebox{0.075\linewidth}{\rotatebox[origin=c]{90}{SelectionConv}} &
\includegraphics[width=.34\linewidth]{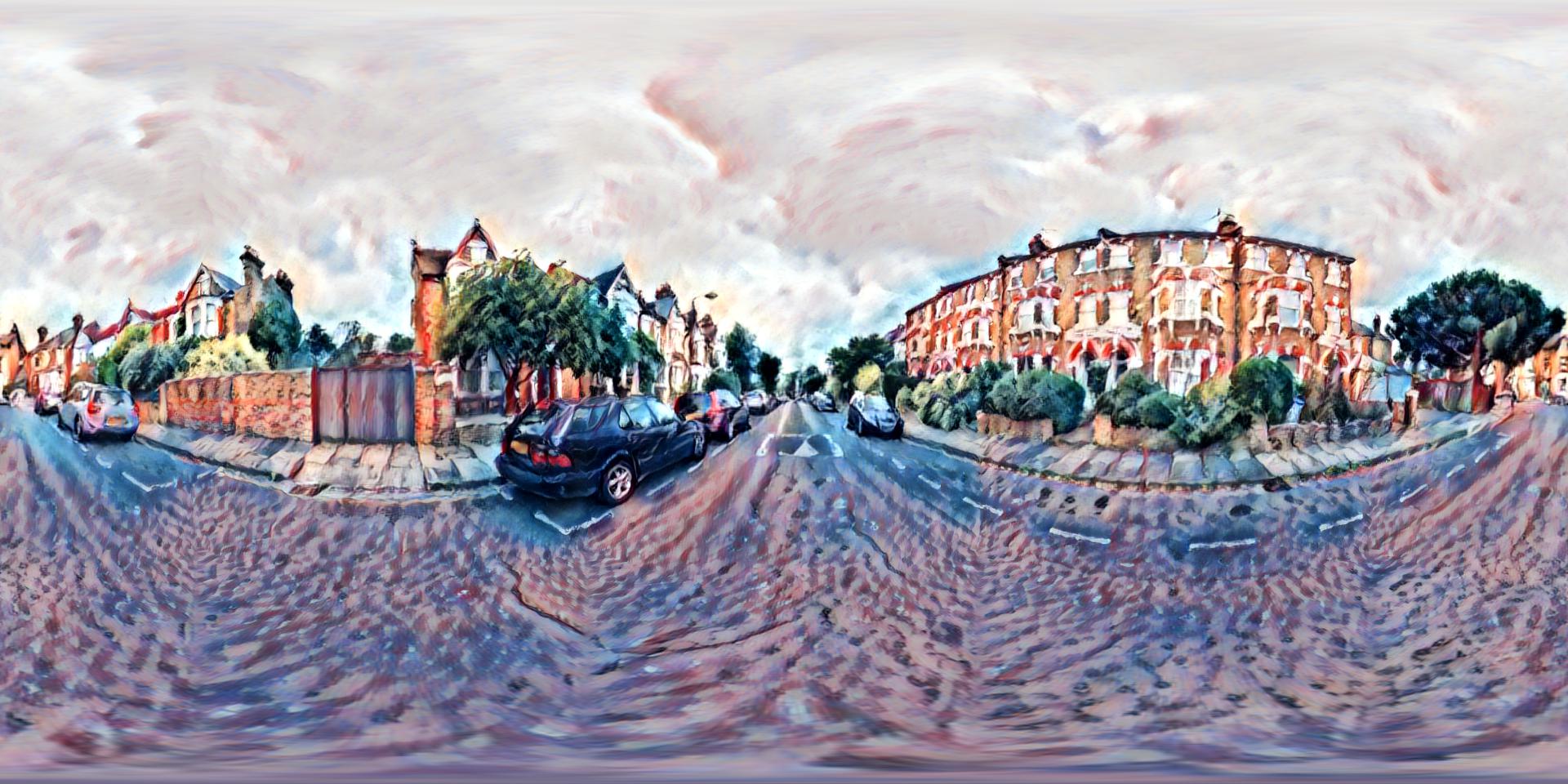} &
\includegraphics[width=.17\linewidth]{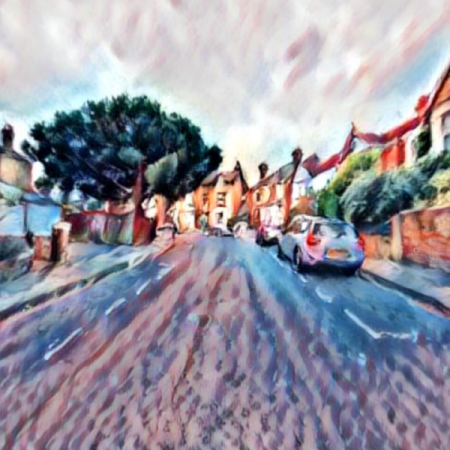} &
\includegraphics[width=.17\linewidth]{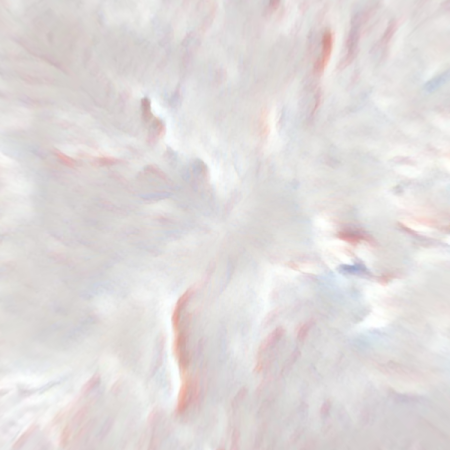} &
\includegraphics[width=.17\linewidth]{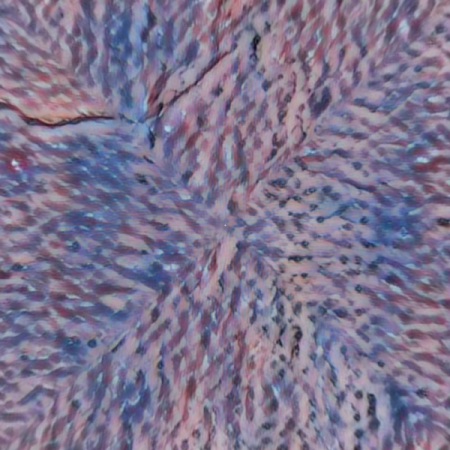} 
\\ 
\raisebox{0.075\linewidth}{\rotatebox[origin=c]{90}{Ours}} &
\includegraphics[width=.34\linewidth]{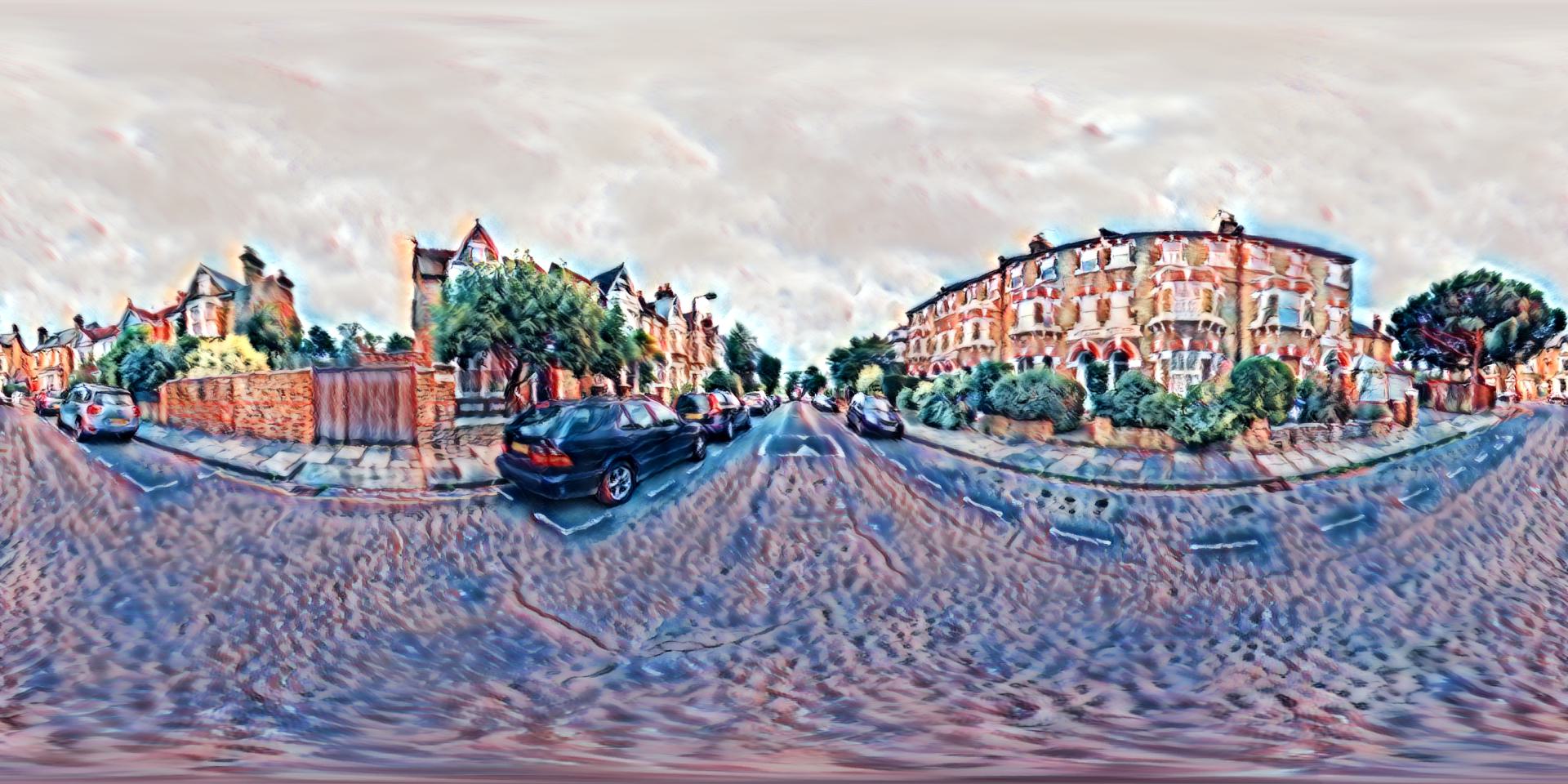} &
\includegraphics[width=.17\linewidth]{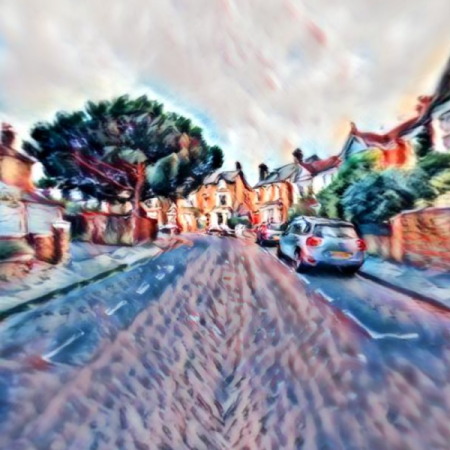} &
\includegraphics[width=.17\linewidth]{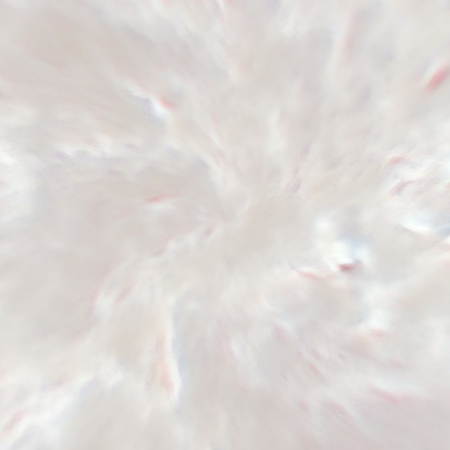} &
\includegraphics[width=.17\linewidth]{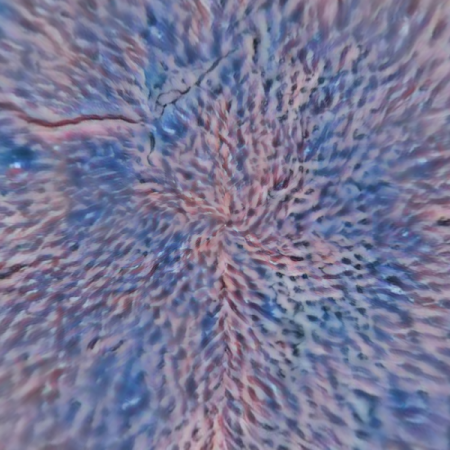} 
\\
&
Equirectangular &
Back &
Top &
Bottom
\end{tabular}
\end{center}
    \caption{A 
    360$^\circ$ image
    (1st row), its stylization when naively stylizing the equirectangular image (2nd row), using the cube-map graph setup from the original SelectionConv paper (3rd row), and compared to our interpolated spherical representation (4th row). The equirectangular projection along with various views of the scene are presented. In the naive approach, note the vertical seam in the middle of the back view as well as the distortion in the top and bottom views. In the original SelectionConv results, note the artifacts in the top and bottom views along the seam connections (making an x shape). Those artifacts are removed with our new method. Public domain image courtesy of polyhaven.com.
    }
\label{fig:spherestyle2}
\end{figure*}

\begin{figure*}
\begin{center}

\includegraphics[width=.20\linewidth]{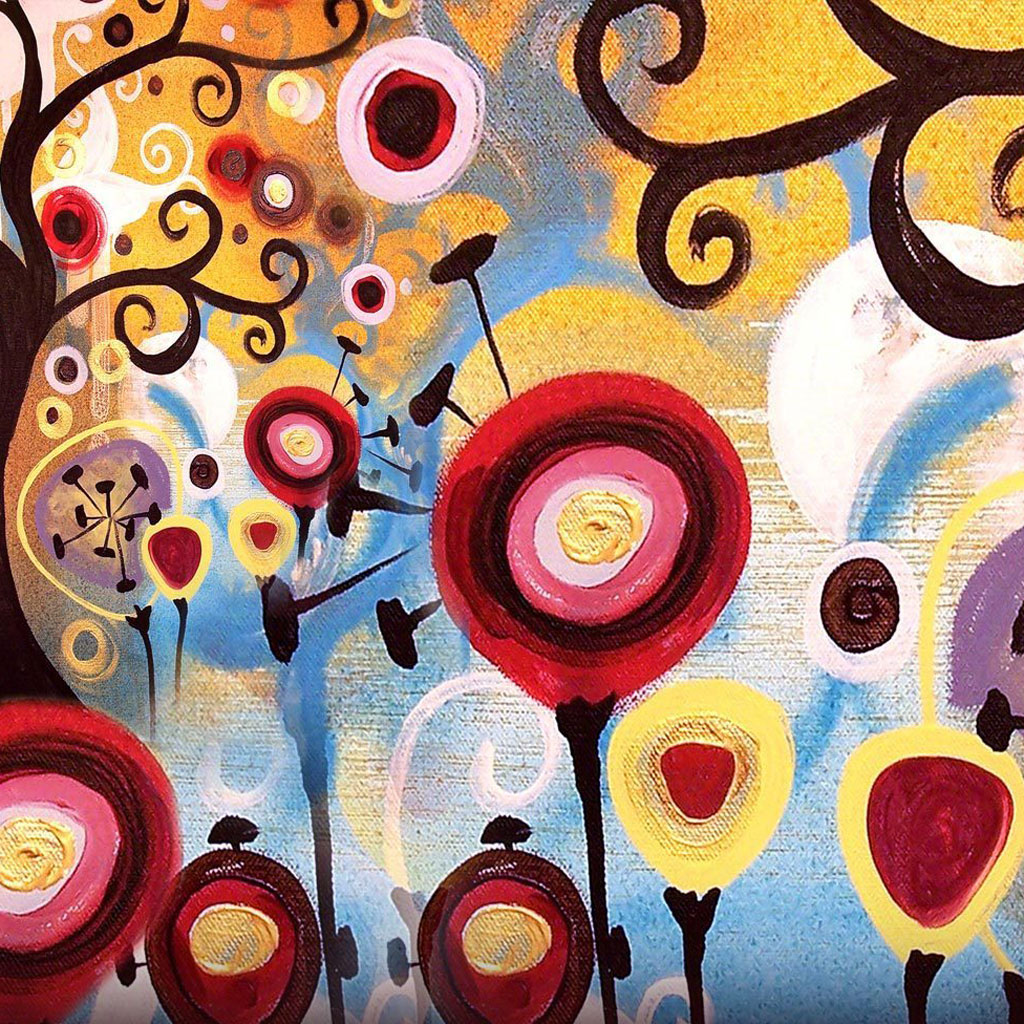}
\vspace{.5in}

\begin{tabular}{rcccc}
\raisebox{0.075\linewidth}{\rotatebox[origin=c]{90}{Spherical Image}} &
\includegraphics[width=.34\linewidth]{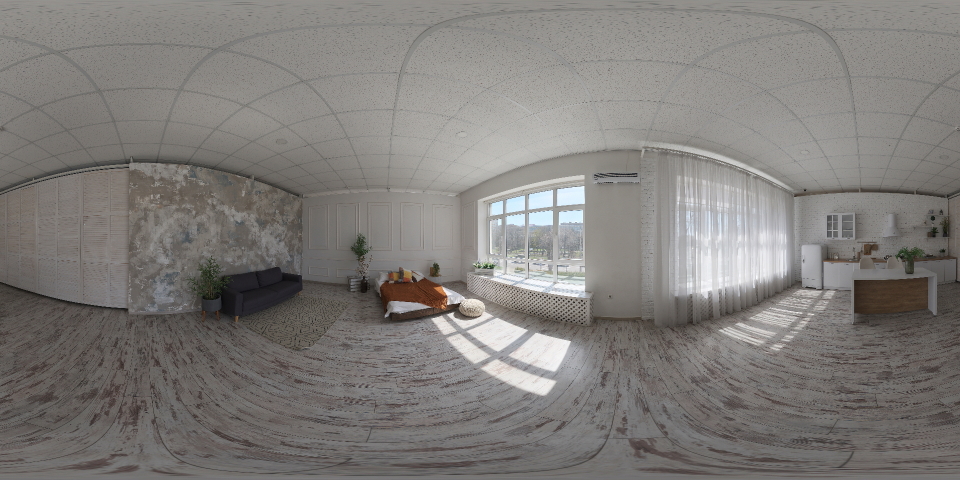} &
\includegraphics[width=.17\linewidth]{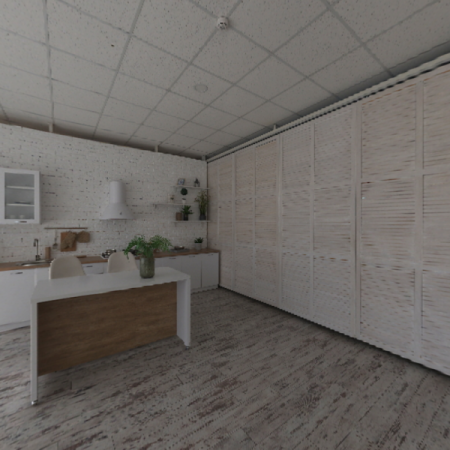} &
\includegraphics[width=.17\linewidth]{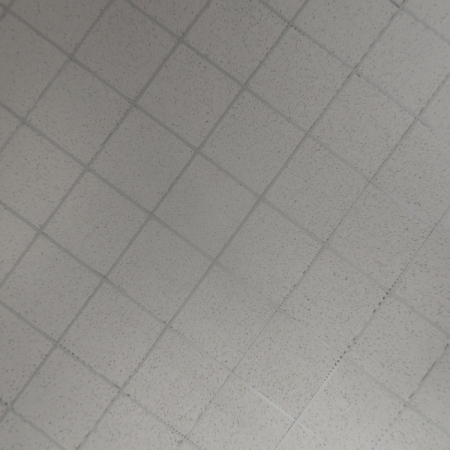} &
\includegraphics[width=.17\linewidth]{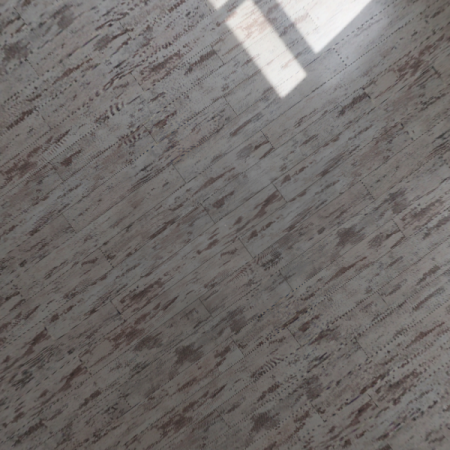}
\\
\raisebox{0.075\linewidth}{\rotatebox[origin=c]{90}{Naive}} &
\includegraphics[width=.34\linewidth]{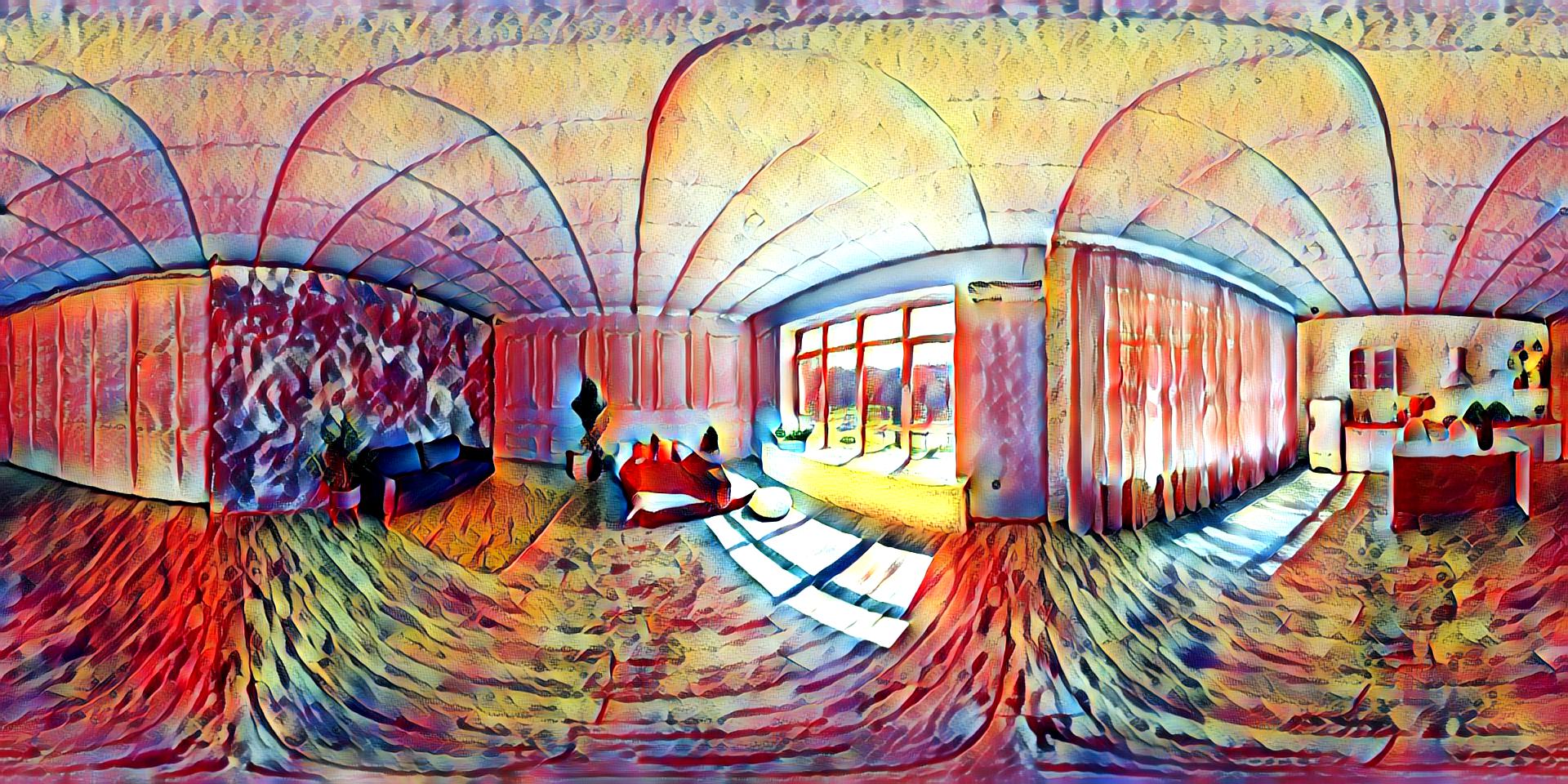} &
\includegraphics[width=.17\linewidth]{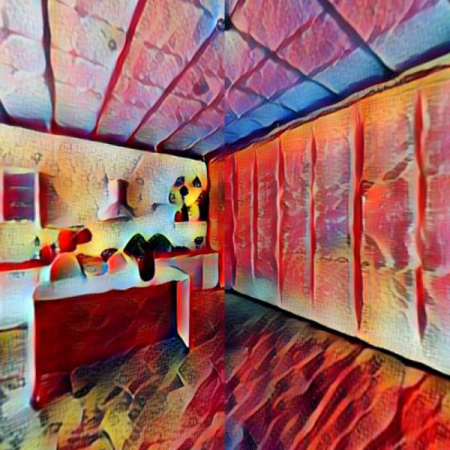} &
\includegraphics[width=.17\linewidth]{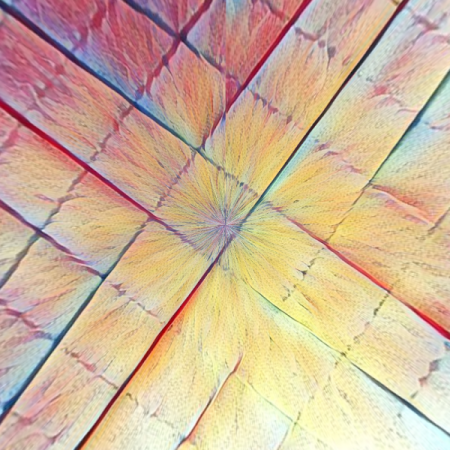} &
\includegraphics[width=.17\linewidth]{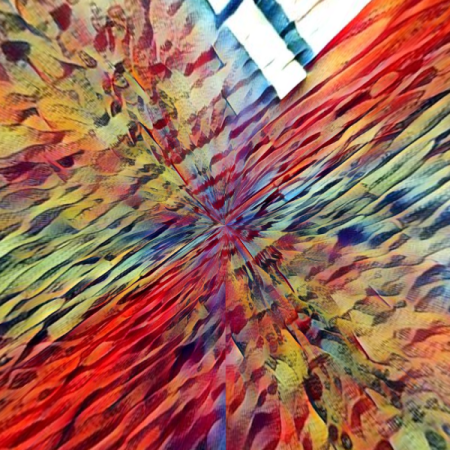} 
\\ 
\raisebox{0.075\linewidth}{\rotatebox[origin=c]{90}{SelectionConv}} &
\includegraphics[width=.34\linewidth]{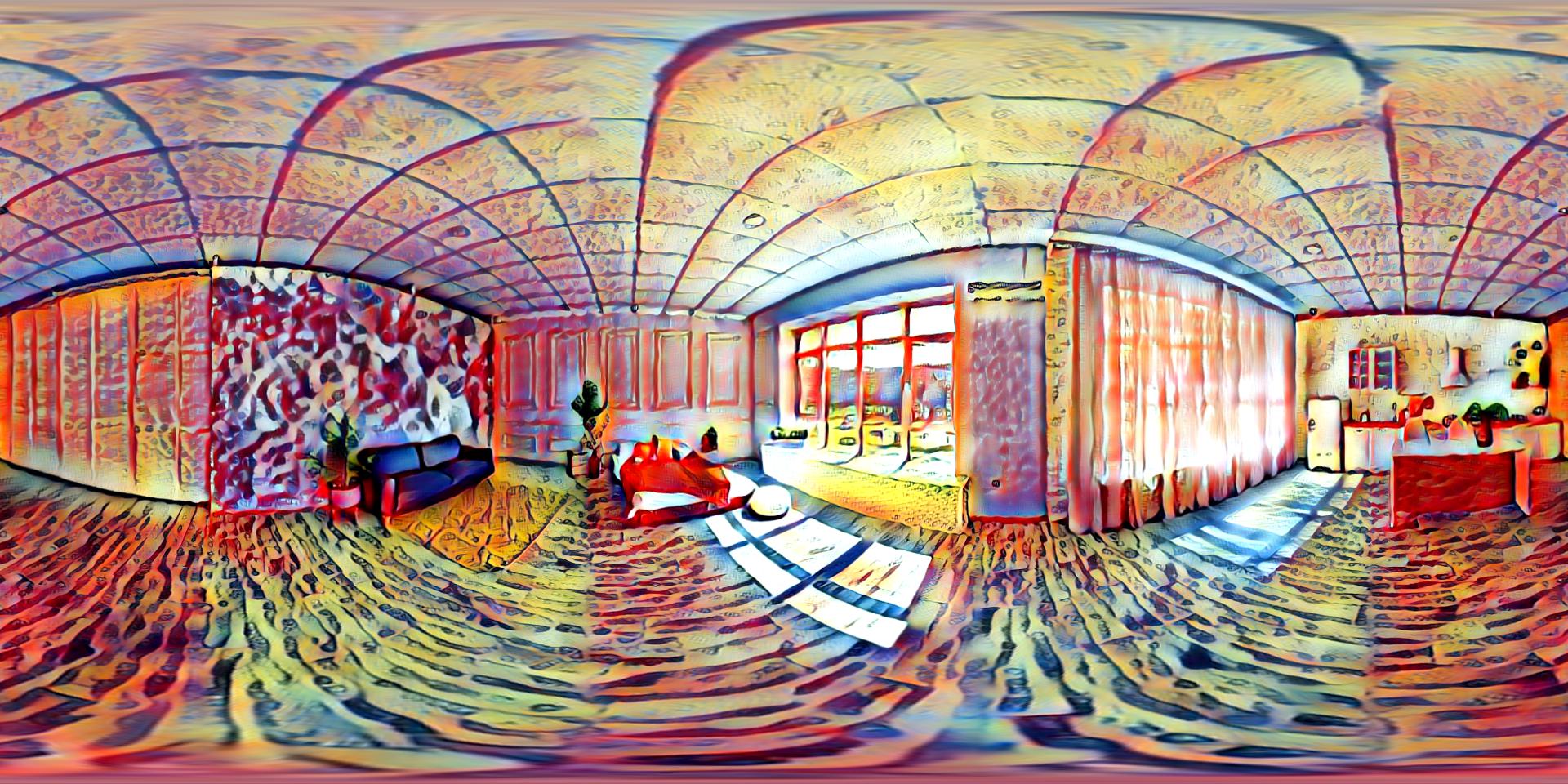} &
\includegraphics[width=.17\linewidth]{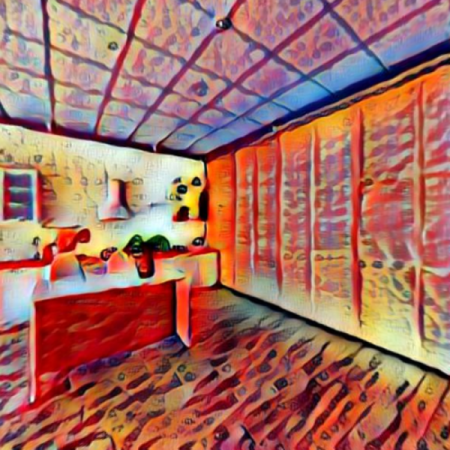} &
\includegraphics[width=.17\linewidth]{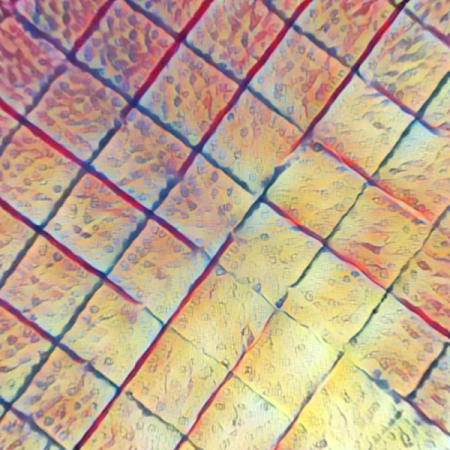} &
\includegraphics[width=.17\linewidth]{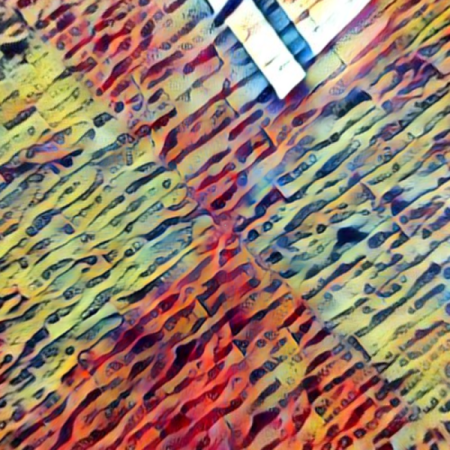} 
\\ 
\raisebox{0.075\linewidth}{\rotatebox[origin=c]{90}{Ours}} &
\includegraphics[width=.34\linewidth]{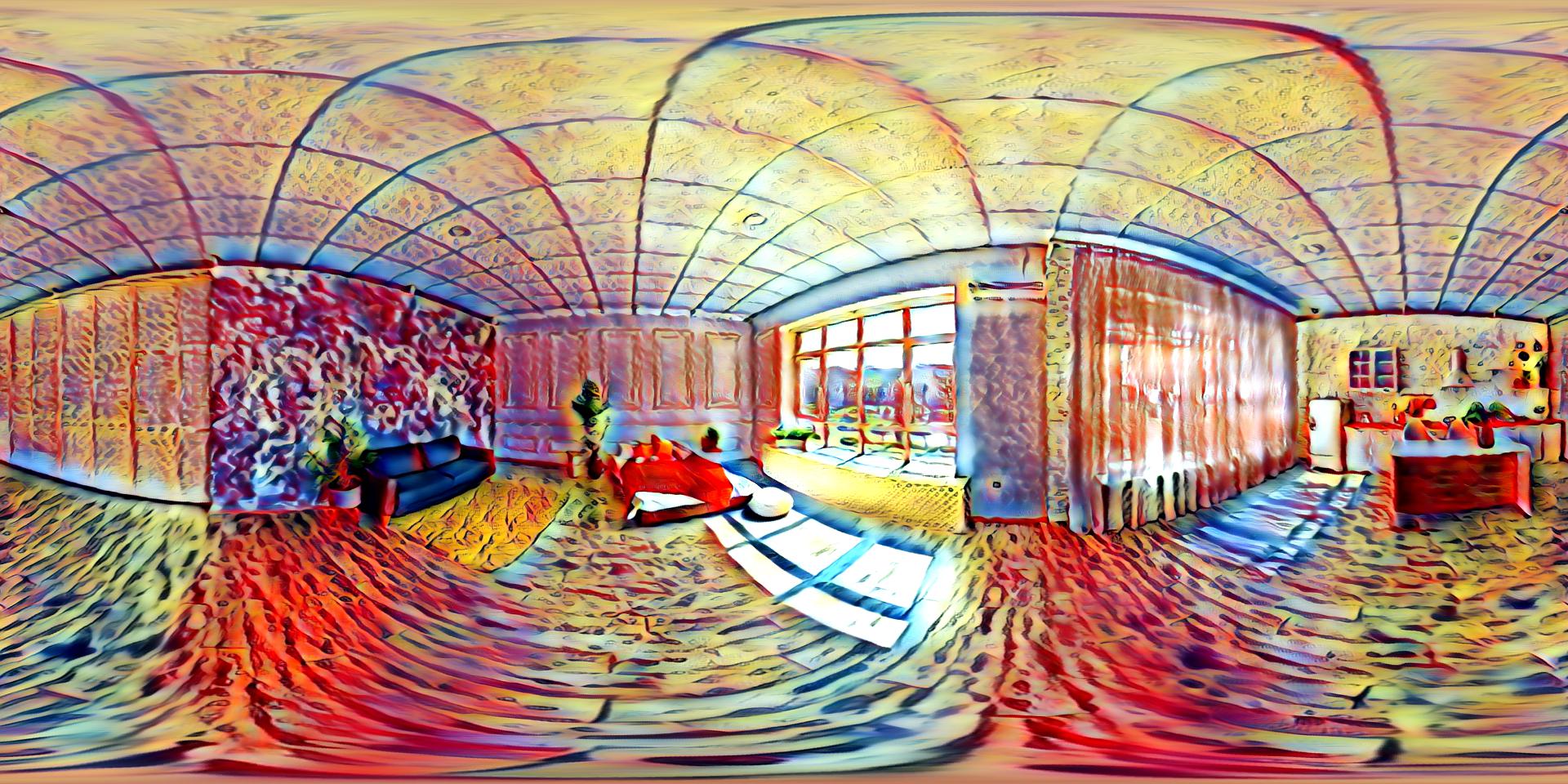} &
\includegraphics[width=.17\linewidth]{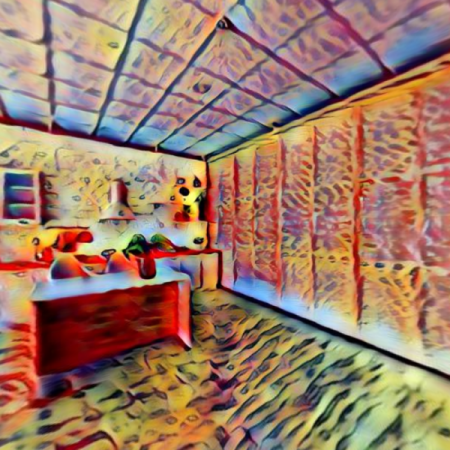} &
\includegraphics[width=.17\linewidth]{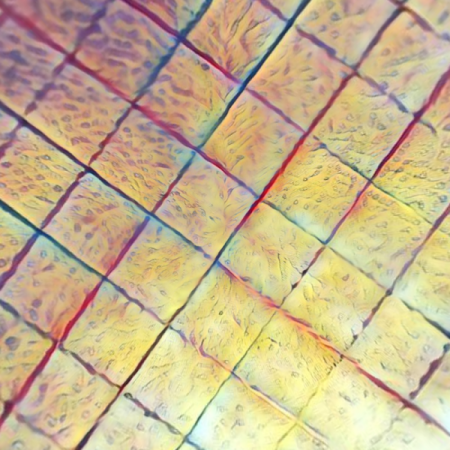} &
\includegraphics[width=.17\linewidth]{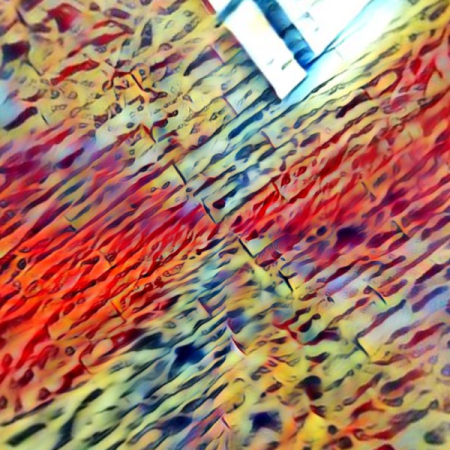} 
\\
&
Equirectangular &
Back &
Top &
Bottom
\end{tabular}
\end{center}
    \caption{A 
    360$^\circ$ image
    (1st row), its stylization when naively stylizing the equirectangular image (2nd row), using the cube-map graph setup from the original SelectionConv paper (3rd row), and compared to our interpolated spherical representation (4th row). The equirectangular projection along with various views of the scene are presented. In the naive approach, note the vertical seam in the middle of the back view as well as the distortion in the top and bottom views. In the original SelectionConv results, note the artifacts in the top and bottom views along the seam connections (making an x shape). Those artifacts are removed with our new method. Public domain image courtesy of polyhaven.com.
    }
\label{fig:spherestyle3}
\end{figure*}

\begin{figure*}
\begin{center}

\includegraphics[width=.20\linewidth]{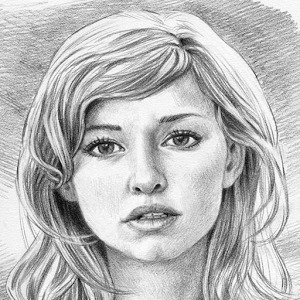}
\vspace{.5in}

\begin{tabular}{rcccc}
\raisebox{0.075\linewidth}{\rotatebox[origin=c]{90}{Spherical Image}} &
\includegraphics[width=.34\linewidth]{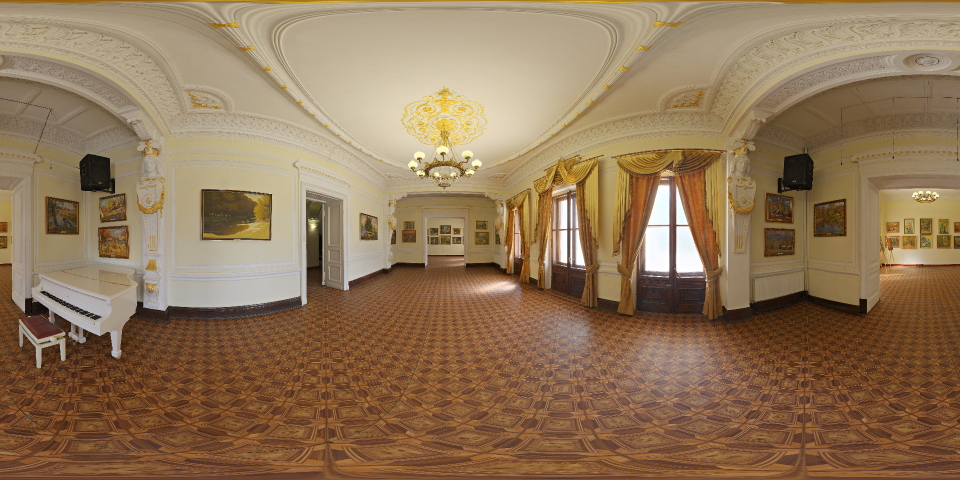} &
\includegraphics[width=.17\linewidth]{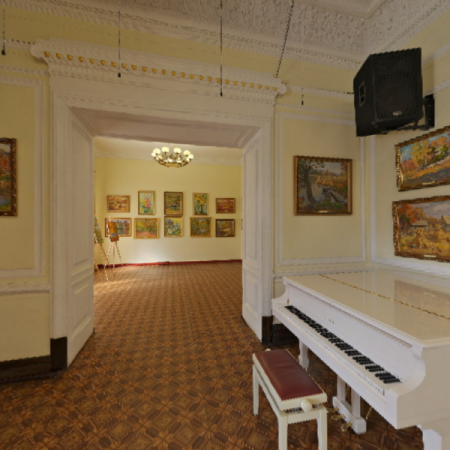} &
\includegraphics[width=.17\linewidth]{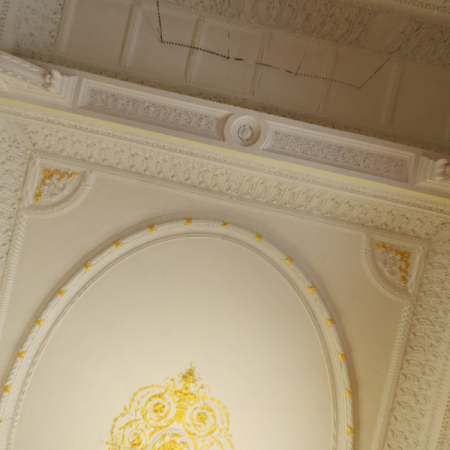} &
\includegraphics[width=.17\linewidth]{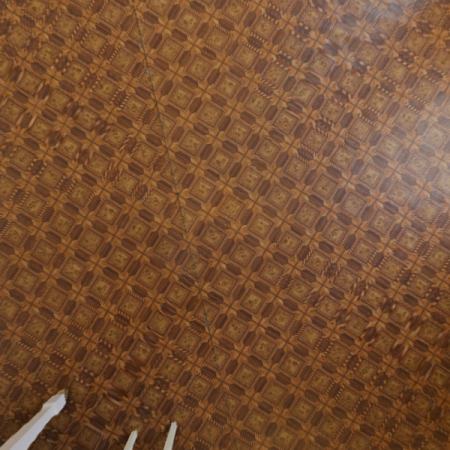}
\\
\raisebox{0.075\linewidth}{\rotatebox[origin=c]{90}{Naive}} &
\includegraphics[width=.34\linewidth]{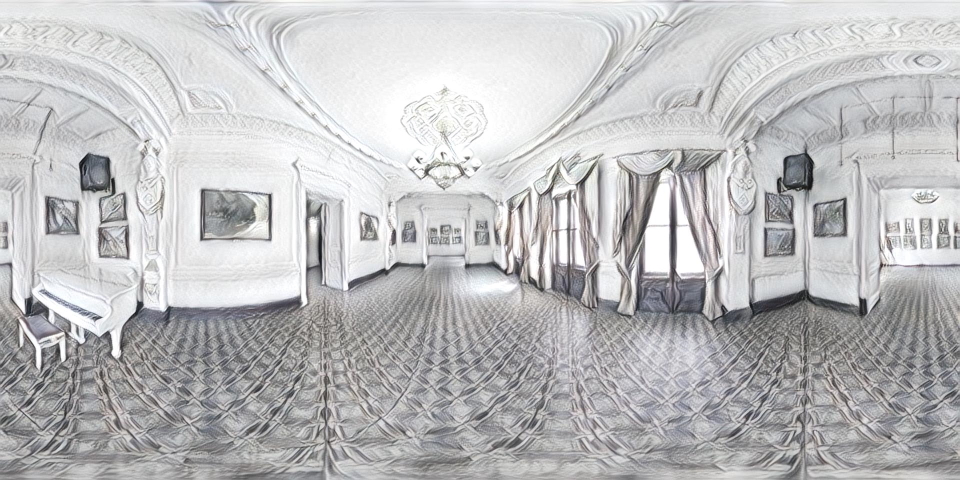} &
\includegraphics[width=.17\linewidth]{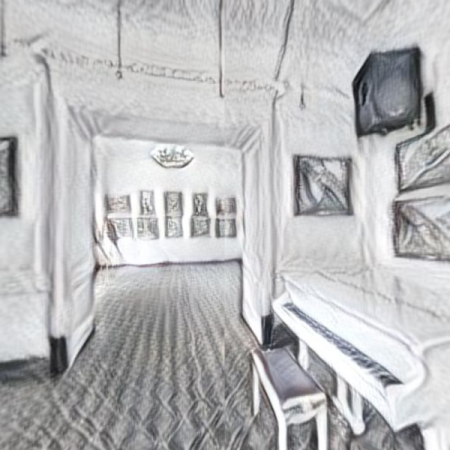} &
\includegraphics[width=.17\linewidth]{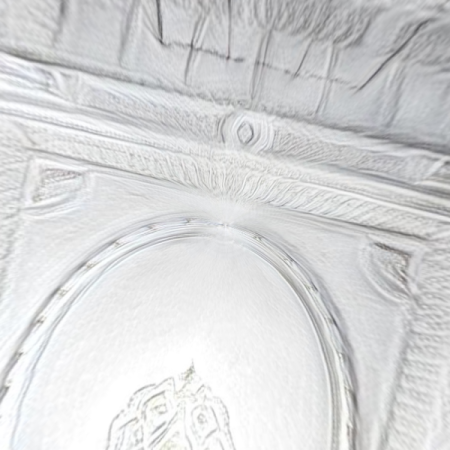} &
\includegraphics[width=.17\linewidth]{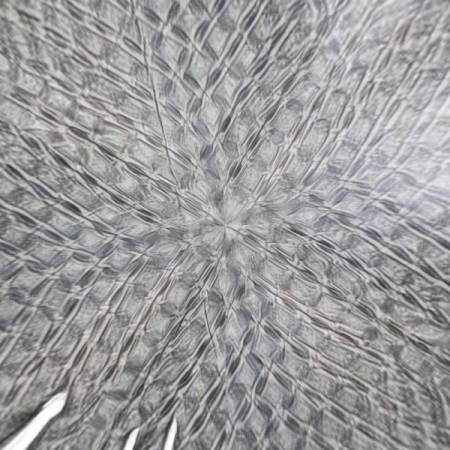} 
\\ 
\raisebox{0.075\linewidth}{\rotatebox[origin=c]{90}{SelectionConv}} &
\includegraphics[width=.34\linewidth]{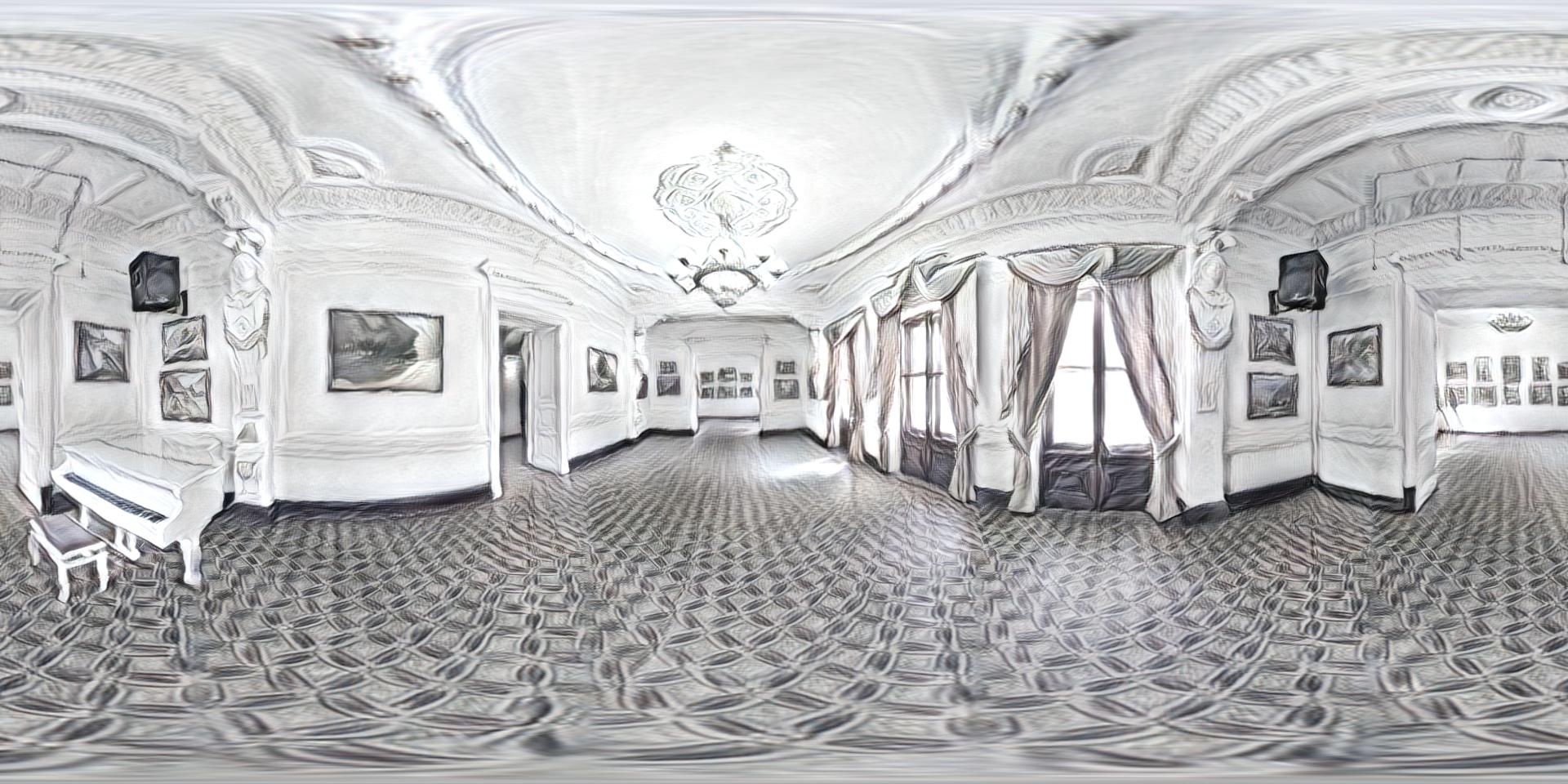} &
\includegraphics[width=.17\linewidth]{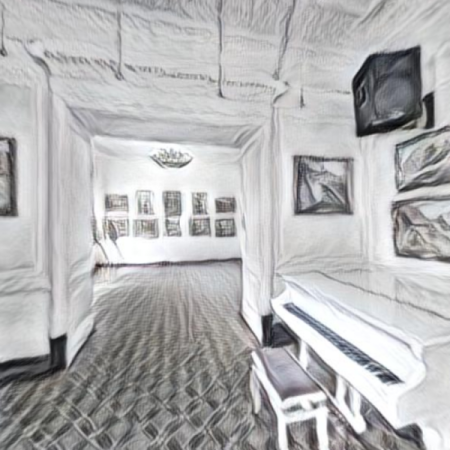} &
\includegraphics[width=.17\linewidth]{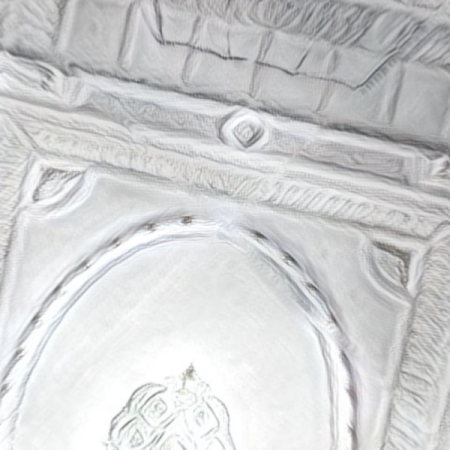} &
\includegraphics[width=.17\linewidth]{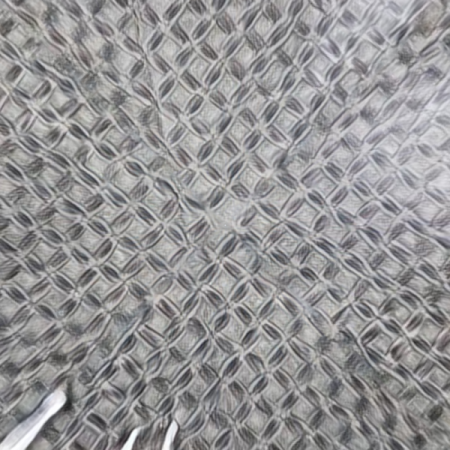} 
\\ 
\raisebox{0.075\linewidth}{\rotatebox[origin=c]{90}{Ours}} &
\includegraphics[width=.34\linewidth]{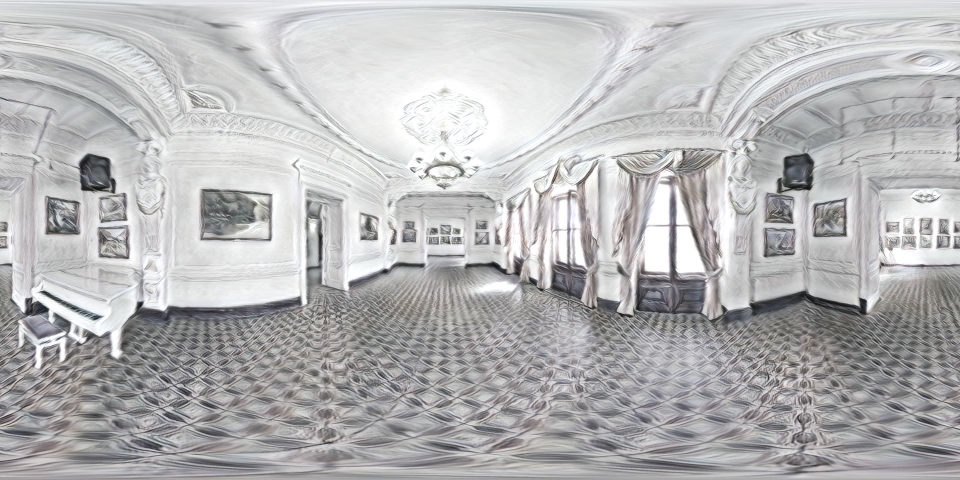} &
\includegraphics[width=.17\linewidth]{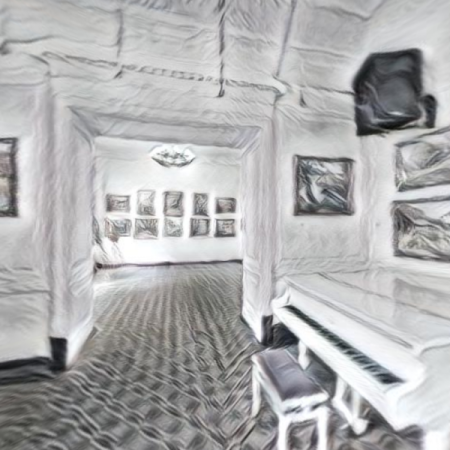} &
\includegraphics[width=.17\linewidth]{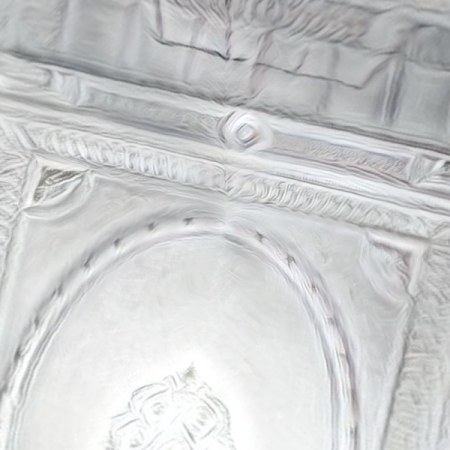} &
\includegraphics[width=.17\linewidth]{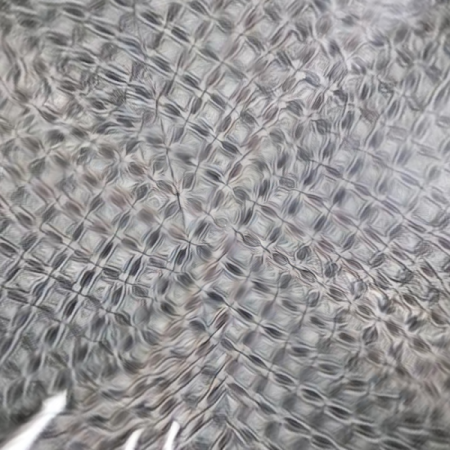} 
\\
&
Equirectangular &
Back &
Top &
Bottom
\end{tabular}
\end{center}
    \caption{A 
    360$^\circ$ image
    (1st row), its stylization when naively stylizing the equirectangular image (2nd row), using the cube-map graph setup from the original SelectionConv paper (3rd row), and compared to our interpolated spherical representation (4th row). The equirectangular projection along with various views of the scene are presented. In the naive approach, note the vertical seam in the middle of the back view as well as the distortion in the top and bottom views. In the original SelectionConv results, note the artifacts in the top and bottom views along the seam connections (making an x shape). Those artifacts are removed with our new method. Public domain image courtesy of polyhaven.com.
    }
\label{fig:spherestyle4}
\end{figure*}

We provide additional results similar to those shown in Figure 5 in the paper, where we compare our approach to previous spherical stylization methods. Those results are shown in Fig.~\ref{fig:spherestyle1} -~\ref{fig:spherestyle4}. 

\section{Mesh Stylization Results}

\begin{figure*}
\begin{center}
\begin{tabular}{ccccc}
\includegraphics[width=.17\linewidth]{Supp_Figures/style2.jpg} &
\includegraphics[width=.17\linewidth]{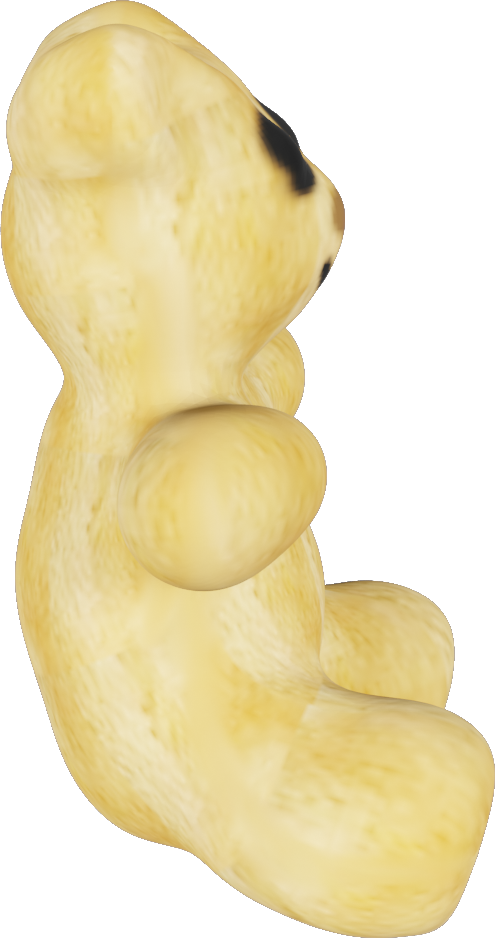} &
\includegraphics[width=.17\linewidth]{Figures/teddy_side_naive.png} &
\includegraphics[width=.17\linewidth]{Figures/teddy_side_sc.png} &
\includegraphics[width=.17\linewidth]{Figures/teddy_side_ours.png} 
\\
Style &
a) Mesh &
b) Naive &
c) SelectionConv &
d) Ours
\end{tabular}
\end{center}
   \caption{ 
   When the texture map of the original mesh (a) is stylized naively (b), many artifacts are present along the UV seams. Stylizing with SelectionConv (c) removed some of those artifacts, but inconsistencies remain. Interpolated SelectionConv (d) retains far greater consistency along seams.  }
\label{fig:teddystyle}
\end{figure*}

\begin{figure*}
\begin{center}
\begin{tabular}{cccc}
\raisebox{.2in}{\includegraphics[width=.12\linewidth]{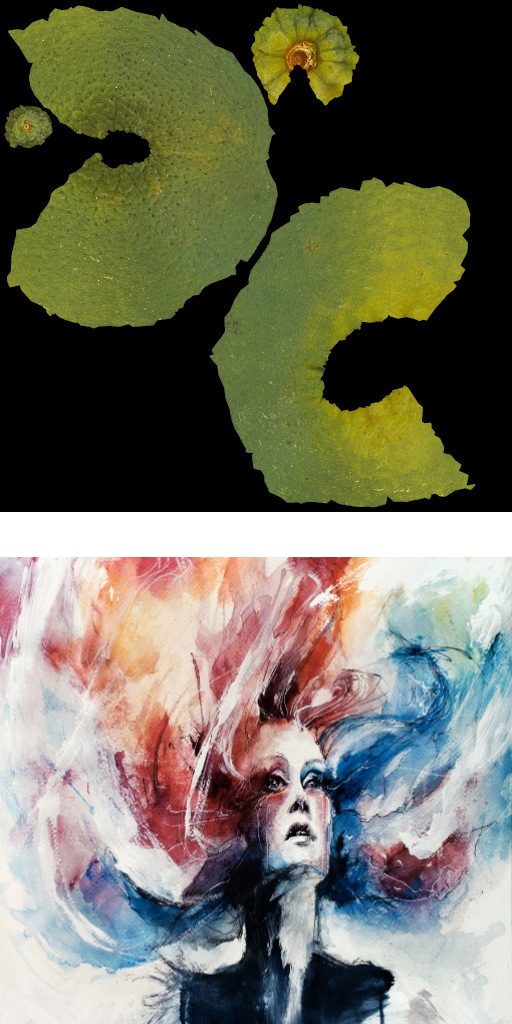}} &
\includegraphics[width=.25\linewidth]{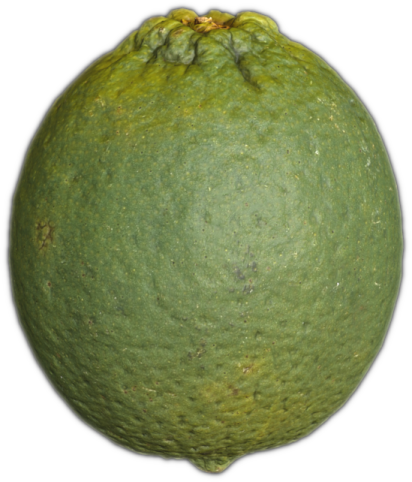} &
\includegraphics[width=.25\linewidth]{Figures/lime_naive.png} &
\includegraphics[width=.25\linewidth]{Figures/lime_ours.png}
\\
\raisebox{-.3in}{\includegraphics[width=.12\linewidth]{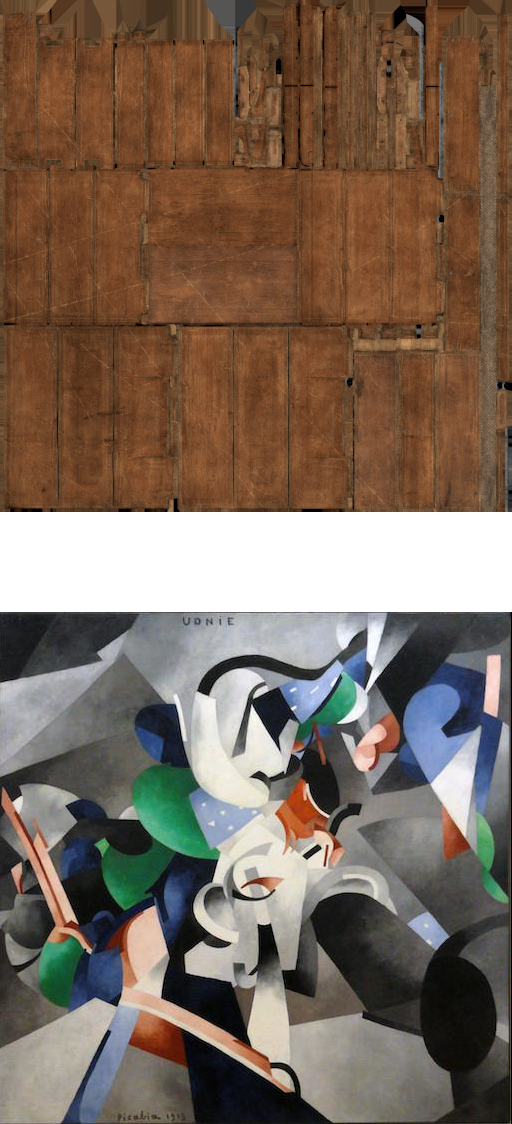}} &
\includegraphics[width=.25\linewidth]{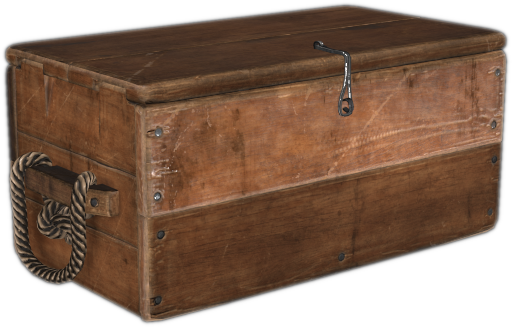} &
\includegraphics[width=.25\linewidth]{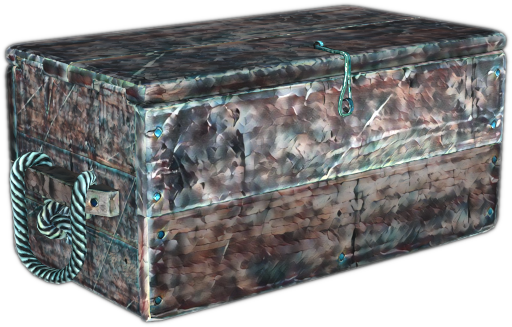} &
\includegraphics[width=.25\linewidth]{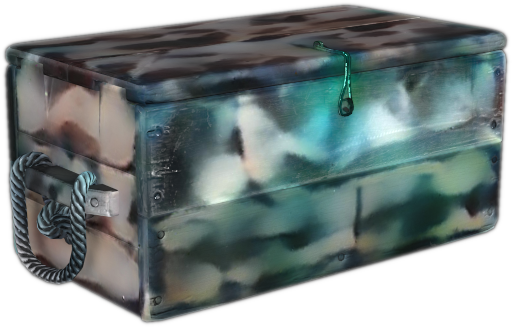} 
\\
&
a) Mesh &
b) Naive &
c) Ours
\end{tabular}
\end{center}
   \caption{ Example stylizations of high quality meshes with 4K textures (a). Naively stylizing the texture map (b) is slow, leaves artifacts on seams, and provides little control for the level of detail for the stylization. Our approach (c) removes seam artifacts while allowing the user to control the number of sampling points, which in turn determine the speed and detail of the stylization. Public domain meshes courtesy of polyhaven.com. }
\label{fig:meshstylization}
\end{figure*}

\begin{figure*}
\begin{center}
\begin{tabular}{cccccc}

\raisebox{.5in}{\includegraphics[width=.08\linewidth]{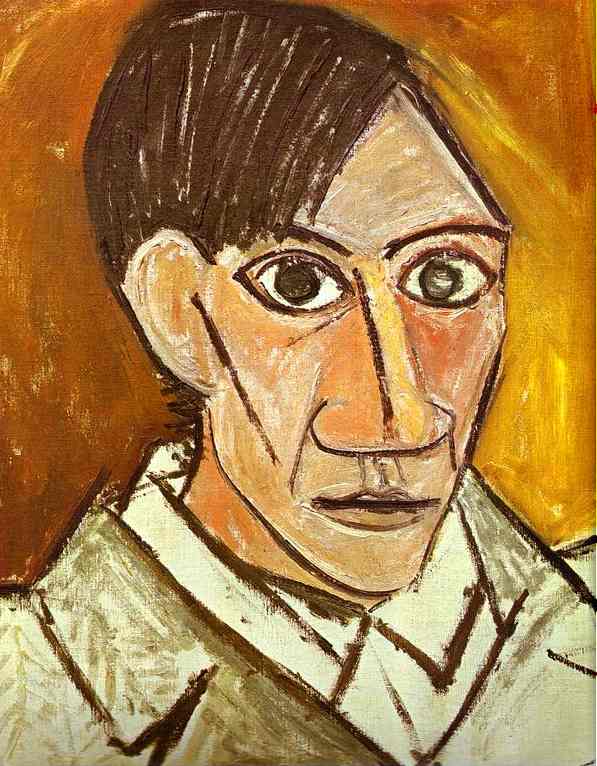}} &
\includegraphics[width=.15\linewidth]{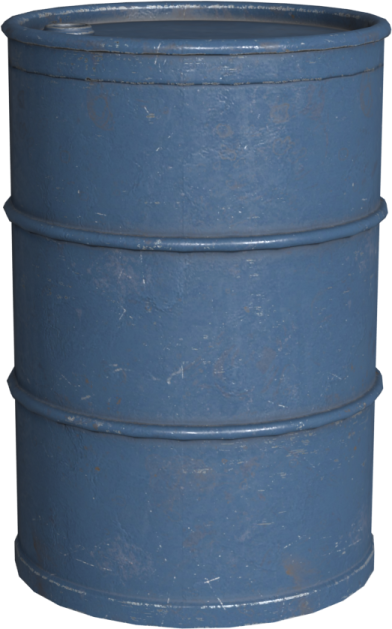} &
\includegraphics[width=.15\linewidth]{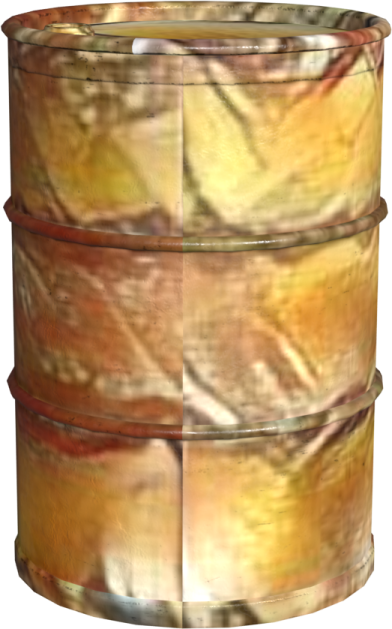} &
\includegraphics[width=.15\linewidth]{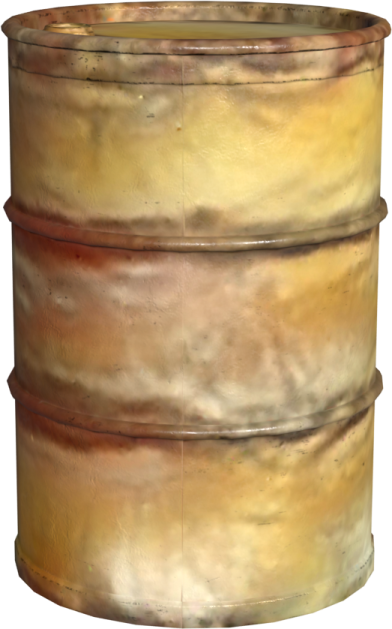} &
\includegraphics[width=.15\linewidth]{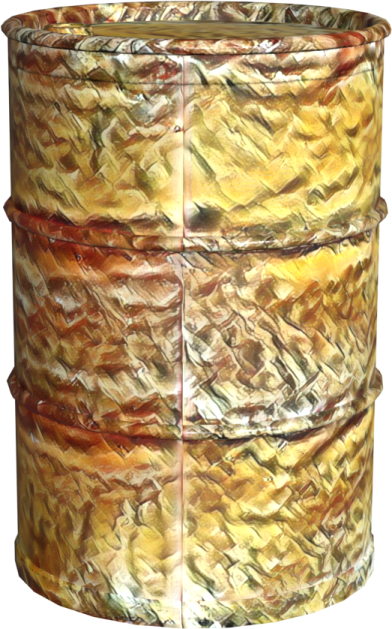} &
\includegraphics[width=.15\linewidth]{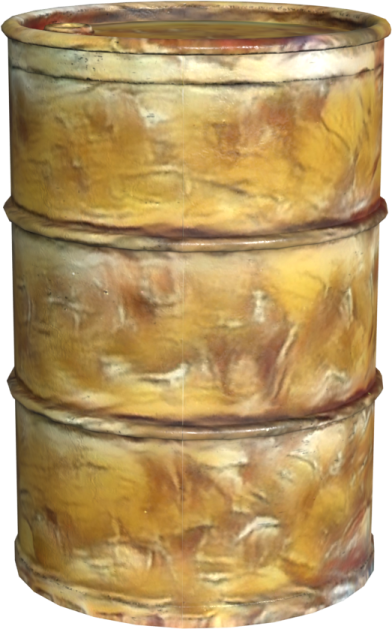}

\\
\raisebox{1.25in}{\includegraphics[width=.08\linewidth]{Supp_Figures/style0.jpg}} &
\includegraphics[width=.15\linewidth]{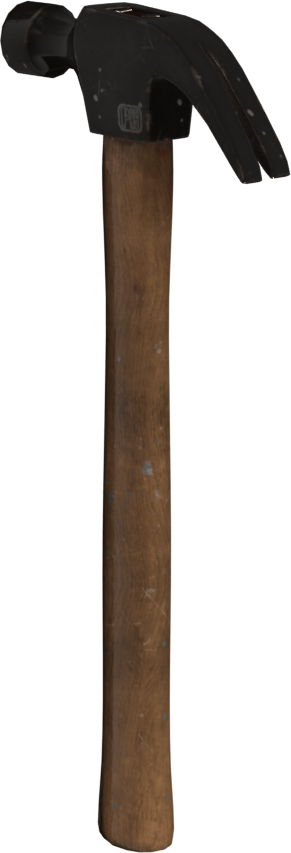} &
\includegraphics[width=.15\linewidth]{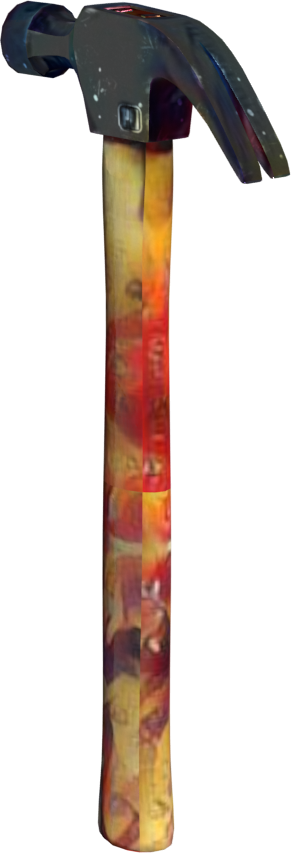} &
\includegraphics[width=.15\linewidth]{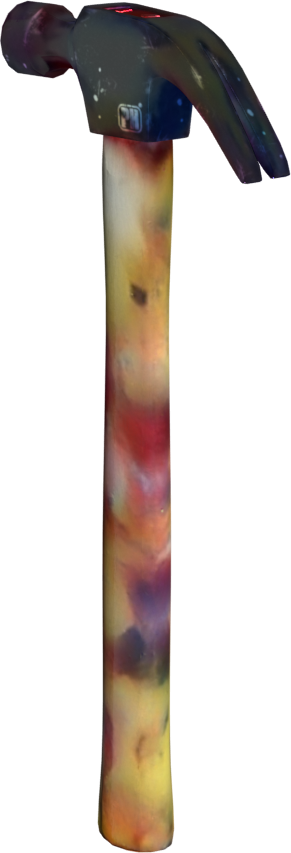} &
\includegraphics[width=.15\linewidth]{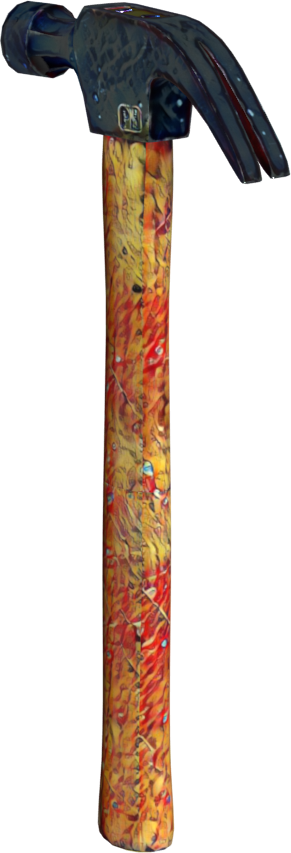} &
\includegraphics[width=.15\linewidth]{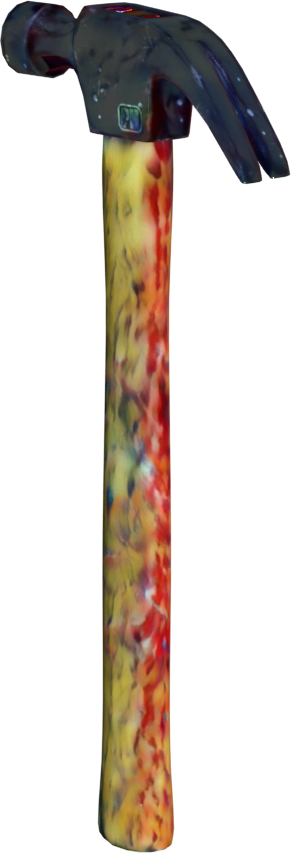}
\\

a) Style &
b) Mesh &
c) Naive (Low Res) &
d) Ours (Low Res) &
e) Naive (Full Res) &
f) Ours 
\end{tabular}
\end{center}
   \caption{ Given a style image (a) and 3D mesh (b), the naive method must style the texture map directly. It may do so on a downsampled version of the texture map (c) or at the original resolution (e), but both fail to properly handle UV seams or provide the user with an effective way to control the level of detail in the stylization. In comparison, our approach can control the number sampling points, with a low sampling giving a more smooth and larger-feature stylization (d) and a higher sampling giving a more varied stylization (f). Public domain meshes courtesy of polyhaven.com. }
\label{fig:meshstylization2}
\end{figure*}

\begin{figure*}
\begin{center}
\begin{tabular}{cc}

\includegraphics[width=.20\linewidth]{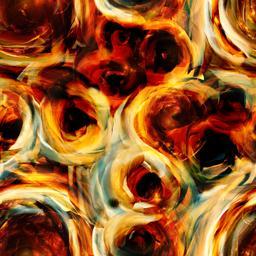} &
\includegraphics[width=.35\linewidth]{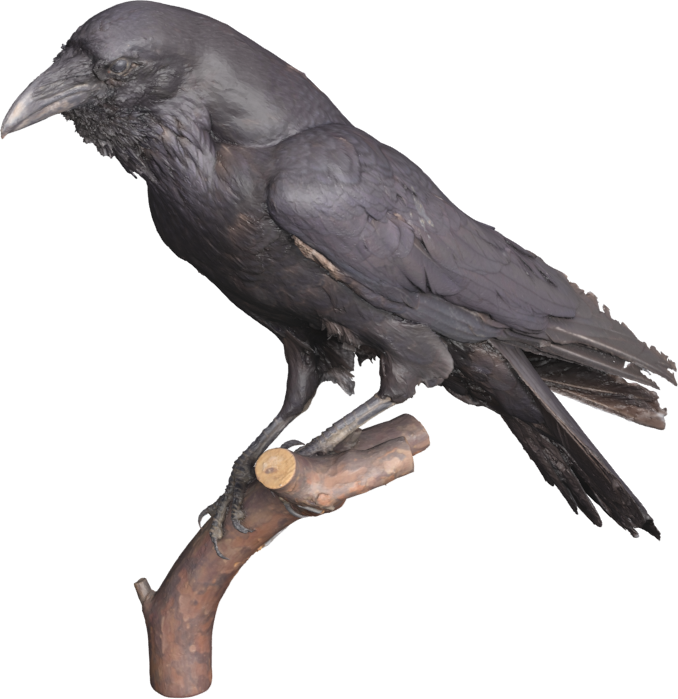} 
\\
a) Style &
b) Mesh
\\
\includegraphics[width=.35\linewidth]{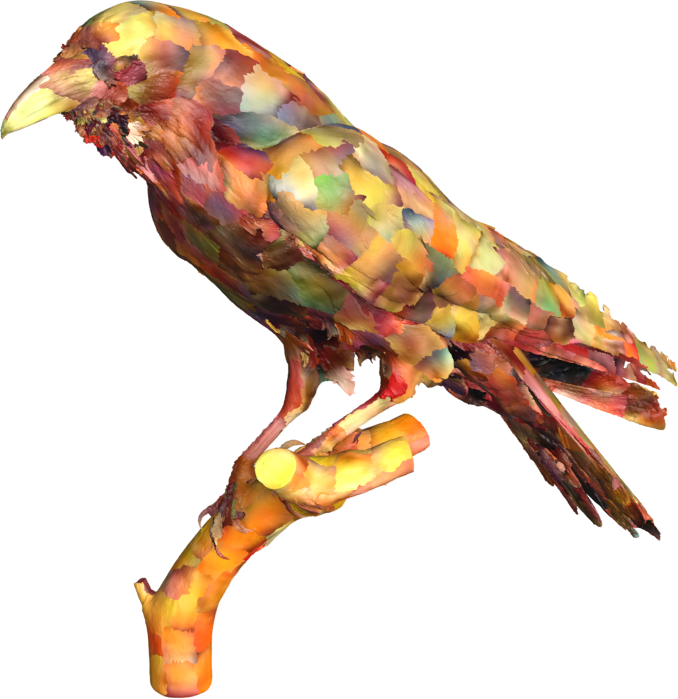} &
\includegraphics[width=.35\linewidth]{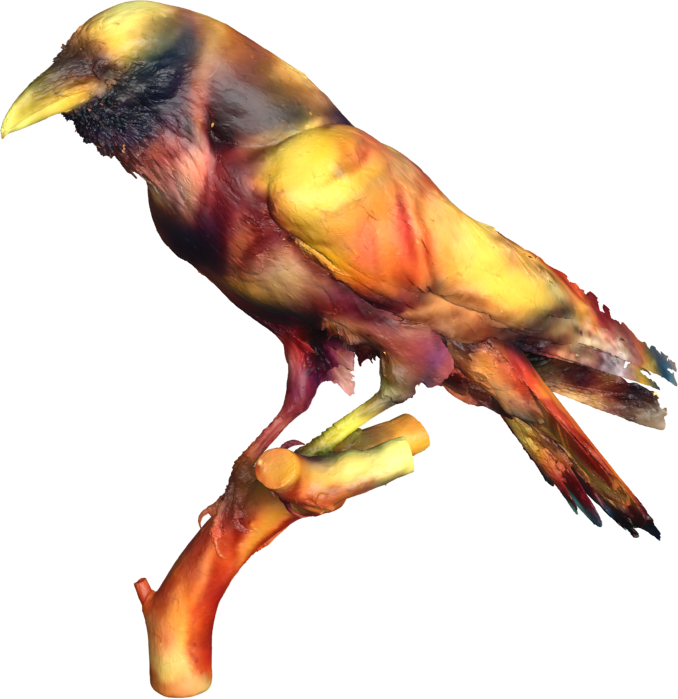} 
\\
c) Naive (Low Res) &
d) Ours (Low Res) 
\\
\includegraphics[width=.35\linewidth]{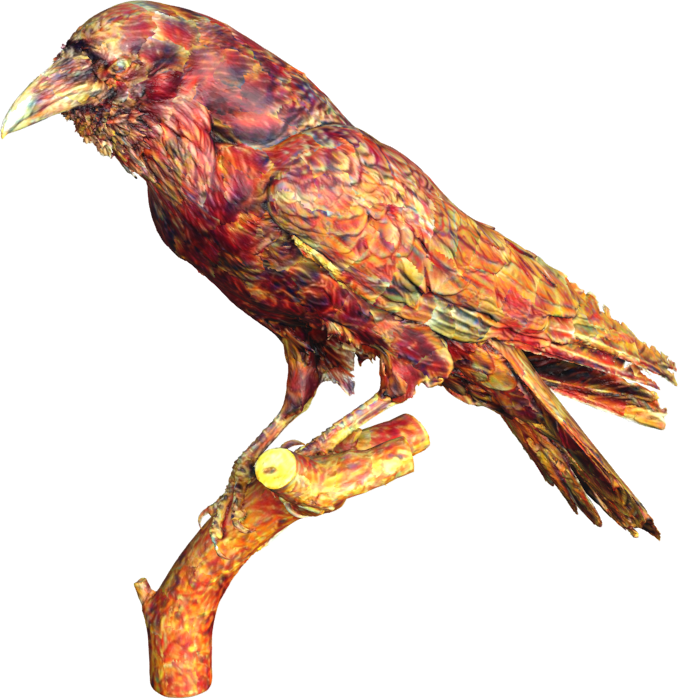} &
\includegraphics[width=.35\linewidth]{Supp_Figures/crow_ours.png}
\\
e) Naive (Full Res) &
f) Ours 
\end{tabular}
\end{center}
   \caption{ Given a style image (a) and 3D mesh (b), the naive method must style the texture map directly. It may do so on a downsampled version of the texture map (c) or at the original resolution (e), but both fail to properly handle UV seams or provide the user with an effective way to control the level of detail in the stylization. In comparison, our approach can control the number sampling points, with a low sampling giving a more smooth and larger-feature stylization (d) and a higher sampling giving a more varied stylization (f).}
\label{fig:meshstylization3}
\end{figure*}

We provide expanded versions of Figures 6 and 7 from the paper (Fig.~\ref{fig:teddystyle} and Fig.~\ref{fig:meshstylization}) as well as additional results in Fig.~\ref{fig:meshstylization2} and Fig.~\ref{fig:meshstylization3}.

